\documentclass[11pt,a4paper]{article}
\usepackage[pdftex]{graphicx}
\usepackage[cmex10]{amsmath}
\usepackage{amsfonts}
\usepackage{amssymb}
\usepackage{amsthm}
\usepackage{siunitx}

\usepackage{algorithm}
\usepackage{algpseudocode}
\usepackage[hidelinks=true]{hyperref}

\usepackage{xfrac}
\usepackage{enumerate}

\usepackage{subfigure}
\usepackage[font=footnotesize ]{caption}

\usepackage{multirow}
\usepackage{rotating}
\usepackage{makecell}
\usepackage{booktabs}
\usepackage{tabularx,ragged2e}
\usepackage{longtable}
\usepackage[semicolon,authoryear]{natbib}

\usepackage{anysize}
\marginsize{2.5cm}{2.5cm}{2.5cm}{2.5cm}

\usepackage{xcolor}
\definecolor{correct}{rgb}{0.77, 0.12, 0.23} % cardinal 
\definecolor{revise}{rgb}{0.0, 0.34, 0.25} % sacramentostategreen
\definecolor{insert}{rgb}{0.0, 0.2, 0.4} % darkmidnightblue

\usepackage{setspace}
\usepackage{parskip}
\usepackage[pagewise]{lineno}
\usepackage{lipsum}

\usepackage{authblk}
\title{\textbf{Backpropagation Neural Tree}}
\author[1]{Varun~Ojha\thanks{Corresponding Author: Varun~Ojha,~email: v.k.ojha@reading.ac.uk; \\Cite as: Ojha, Varun and Nicosia, Giuseppe (2022) \textit{Neural Networks}, Elsevier}}
\author[2,3]{Giuseppe Nicosia}

\affil[1]{Department of Computer Science, University of Reading, Reading, UK}
\affil[2]{Center of System Biology, University of Cambridge, Cambridge, UK}
\affil[3]{Department of Biomedical \& Biotechnological Sciences, University of Catania, Catania, Italy}

\date{}
\begin{document}
    
    %\setpagewiselinenumbers
    %\linenumbers
    \onehalfspacing
    \maketitle
    
    \begin{abstract}
        We propose a novel algorithm called \textit{Backpropagation Neural Tree} (BNeuralT), which is a \textit{stochastic computational dendritic tree}.
        BNeuralT takes random \textit{repeated inputs} through its leaves and imposes \textit{dendritic nonlinearities} through its internal connections like a biological \textit{dendritic tree} would do. 
        Considering the dendritic-tree like plausible biological properties, BNeuralT is a \textit{single neuron} neural tree model with its internal sub-trees resembling  dendritic nonlinearities.
        BNeuralT algorithm produces an ad hoc \textit{neural tree} which is trained using a stochastic gradient descent optimizer like gradient descent (GD), momentum GD, Nesterov accelerated GD, Adagrad, RMSprop, or Adam. 
        BNeuralT training has two phases, each computed in a depth-first search manner: the \textit{forward pass} computes neural tree's output in a \textit{post-order traversal}, while the error backpropagation during the \textit{backward pass} is performed recursively in a \textit{pre-order traversal}. 
        A BNeuralT model can be considered a \textit{minimal subset} of a neural network (NN), meaning it is a ``thinned'' NN whose complexity is lower than an ordinary NN. 
        Our algorithm produces high-performing and parsimonious models balancing the complexity with descriptive ability on a wide variety of machine learning problems: classification, regression, and pattern recognition. 
        ~
    \end{abstract}
    
    \section{Introduction}
    \label{sec:intro}
    %%%%%%%%%%%   INTRODUCTION    %%%%%%%%%%%%%%%
    Data-driven learning is a hypothesis (trained model) search from a hypothesis-space that fits input data to its target output as good as possible (a low error on test data). 
    A learning algorithm like neural networks (NNs) parameter optimization via backpropagation is the effort to find such a hypothesis~\citep{annBpRumelhart1986}. 
    We propose a new study of ad hoc \textit{neural trees} 
    generation and their optimization via our \textit{recursive backpropagation algorithm} to find such a hypothesis. 
    Hence, we propose a new algorithm called \textit{Backpropagation Neural Tree} (BNeuralT). 
    
   	A tree of BNeuralT is like a biological dendritic tree~\citep{travis2005regional,mel2016toward} that processes repeated inputs connected to a single neuron~\citep{beniaguev2020single,jones2021might}  through dendritic nonlinearities~\citep{london2005dendritic}.  
	Structurally, BNeuralT model is a stochastic \textit{computational dendritic tree} that takes random \textit{repeated inputs} through its leaves and imposes \textit{dendritic nonlinearities} through its internal nodes like a biological dendritic tree would do~\citep{travis2005regional,jones2021might}. 
    Hence, considering the plausible dendritic-tree-like  biological properties, BNeuralT is a \textit{single neuron} neural tree model with its internal nodes resembling  dendritic nonlinearities.

    Structurally, BNeuralT, being a tree, is a minimal subset of a  (highly sparse) NN whose complexity is comparatively low~\citep{poirazi2003pyramidal}. This means that a NN with a very high \textit{dropout} [a network regularization technique~\citep{srivastava14aJMLR}] prior to its training can be similar to BNeuralT, except BNeuralT has dedicated paths from input to output as opposed to sparse NN that has shared connections between nodes. Hence, we aim to gauge the performance of ad hoc \textit{neural trees} trained using \textit{stochastic gradient descent} (SGD) optimizers like gradient descent (GD), momentum gradient descent (MGD)~\citep{qian1999momentum}, Nesterov accelerated gradient descent (NAG)~\citep{bengio2013advances}, adaptive gradient (Adagrad)~\citep{dean2012large}, root-mean-square gradient propagation (RMSprop)~\citep{tieleman2012lecture}, and adaptive moment estimation (Adam)~\citep{diederik2015adam}. 
    
    Operationally, an expression-tree with its operator (node) being neural nodes (i.e., an operator is an activation function),  edges being neural weights, and leaves being inputs make a neural tree architecture, where the tree's architecture itself can be optimized~\citep{chen2005time,schmidt2009solving}. The tree's edges (parameters) optimization is straightforward using a gradient-free method~\citep{rios2013derivative,kennedy1995particle} where the tree is assumed a target function~\citep{ojha2017ensemble}. However, its gradient-based optimization is non-trivial, especially because the error-backpropagation through the tree data structure is recursive to traverse. Our proposed BNeuralT algorithm does a two-phase computation of a neural tree in  a depth-first search manner: the \textit{forward pass} computes neural tree's outputs in a \textit{post-order traversal}, while the error  backpropagation during the \textit{backward pass} is performed recursively in a \textit{pre-order traversal}. 
    
    We trained ad hoc neural trees in an \textit{online} (example-by-example) and a \textit{mini-batch} mode on a variety of learning problems:  classification, regression, and pattern recognition. For classification and pattern recognition problems, BNeuralT has its root node's children (nodes at tree depth one) strictly dedicated to each target class, and the root node decides the winner class on receiving input data. BNeuralT dedicates its root as the output node for a regression problem.
    
    We evaluated BNeuralT's convergence process on six SGD optimizers and analyzed BNeuralT's complexity against its convergence accuracy. Each training version was compared with a similar training version of a multi-layer perceptron (MLP) algorithm (i.e., an input-hidden-output NN architecture) and classification and regression algorithms such as decision tree (DT)~\citep{breiman1984classification}, random forest (RF)~\citep{breiman2001random}, single and multi-objective versions of a heterogeneous flexible neural tree (HFNT$^{\text{S}}$ and HFNT$^{\text{M}}$)~\citep{ojha2017ensemble}, multi-output neural tree (MONT)~\citep{ojha2020multi},  
    Gaussian process (GP)~\citep{rasmussen2006gaussian}, na\"ive Bayes classifier (NBC)~\citep{mitchell1997machine}, and support vector machine (SVM)~\citep{cortes1995support,chang2011libsvm,fan2008liblinear}. 
    %%
    %%%The BNeuralT on pattern recognition problem was trained using RMSprop optimizer. The results of BNeuralT on pattern recognition problem was compared with results from the literature. 
    %
    The results on all problems indicate the success of our BNeuralT algorithm that produces high-performing and parsimonious models balancing the complexity and descriptive ability with a minimal training hyperparameters setup. %We believe the success of BNeuralT invites exhaustive research on its further improvement and application.
    
    Our contribution is an innovative \textit{Recursive Backpropagation Neural Tree} algorithm that
    \begin{itemize}
    	\item takes inspiration from biological dendritic trees to solve a wide class of machine learning problems through  a single  neuron tree-like model performing dendritic nonlinearities through its internal nodes and resembling a highly sparse neural network. 
    	
    	\item generates low complexity and high accuracy models. Therefore, we have designed a learning system capable of producing minimal and sustainable neural trees that have fewer parameters to produce more compact and, therefore, sustainable neural models able to reduce CPU time and, consequently, CO$_2$ emissions for machine learning applications.
    	
    	\item shows that the sigmoidal dendritic nonlinearity of \textit{any} stochastic ad hoc neural tree structure can solve machine learning problems with high accuracy, and any such structure excels to genetically optimized neural tree structures, NNs, and other learning algorithms.
    \end{itemize}

    This paper presents relevant related work in Sec.~\ref{sec:related_work}. BNeuralT model's architecture and properties are described in Sec.~\ref{sec:bpnt}. Secs.~\ref{sec:hyperparamter} and \ref{sec:exp_version} outline the hyperparameter settings and experiment versions. The performance of BNeuralT on machine learning problems is summarized in Sec.~\ref{sec:res_dis} and discussed in Sec.~\ref{sec:discussion}, followed by conclusions in Sec.~\ref{sec:con}. Source code of BNeuralT algorithm and pre-trained models are available at \url{https://github.com/vojha-code/BNeuralT}.

    \section{Related works}
    \label{sec:related_work}
    %%%%%%%%%%%   RELATED WORK    %%%%%%%%%%%%%%%
    We review the works defining neural tree architectures and training processes. The early definition of neural trees appeared in~\citep{sakar1993growing,sirat1990neural}, where the tree's \textit{``root-to-leaf''} path is represented as a neural network (NN). Such a tree makes its decision through leaf nodes, and its internal nodes are NNs (or neural nodes). \cite{jordan1994hierarchical} proposed a hierarchical mixture expert model that performs construction of a binary tree structure where the model hierarchically combines the outputs of expert networks (feed-forward NNs at the terminal) though getting networks (feed-forward NNs at non-terminal) and propagates computation from \textit{``leaf-to-root''} and where each NN uses the whole input features set. 
    
    In contrast, our model is purely a single network (tree) structure representation, whereas a hierarchical mixture expert model is a hierarchical combination of several (preferably small) networks. Therefore, unlike hierarchical mixture expert model, our model is a subset of a NN where ``leaf-to-root'' has a specific information processing path. 
    In fact, considering plausible inspiration from biological computational dendritic tree~\citep{travis2005regional,mel2016toward,poirazi2003pyramidal}, our model behaves as a single neuron model~\citep{jones2021might}.
    
    Our proposed BNeuralT algorithm generates an \textit{$m$-ary} tree structure stochastically and assigns edge weights randomly. BNeuralT's each leaf node (terminal node) takes a single input variable from a set of all available variables (data features). Therefore, in a generated tree, some features could remain unused by the model leading to \textit{only} select features responsible for the prediction. Moreover, tree's each neural node (non-terminal node) takes a weighted summation of its child's output. Hence, a BNeuralT model potentially performs an \textit{input dimension reduction} and propagates the computation from \textit{leaf to root}. 

    A recent work of~\cite{tanno2019adaptive} demonstrates neural tree as an arrangement of convolution layers and linear classifier as a learning model resembling a decision tree-like classifier where the incoming inputs at the nodes are inferred through the so-called router, processed through tree edges (transformers), and classified through leaf (solver) nodes. In contrast, our model takes image pixels as its inputs.
    A \textit{leaf-to-root} as a neural tree definition appeared in~\citep{zhang1997evolutionary,chen2005time}, where the tree's leaf nodes are designated inputs, internal nodes are neural nodes, and edges are weights. Such types of neural trees have been subjected to structure optimization~\citep{chen2005time,ojha2017ensemble} and parameter optimization via gradient-free optimization techniques like particle swarm optimization~\citep{chen2007flexible} and differential evolution~\citep{ojha2017ensemble}. 
    
    \cite{zhang1997evolutionary} demonstrated that a neural tree could be evolved as a subset of an MLP. Their effort was to evolve a neural tree using genetic programming and optimize parameters using a genetic algorithm. \cite{lee2016generalizing} focused on implementing pooling layers within a convolutional NN as a tree structure. However, our approach is to generate and train ad hoc neural trees using  our proposed recursive backpropagation algorithm. To the best of our knowledge and review, this is the first and novel attempt to generate and train \textit{ad hoc} neural trees using our \textit{recursive error-backpropagation} algorithm. Our motivation is to avoid any prior assumptions on network architecture and complicated hyperparameter settings.
    
    \cite{srivastava14aJMLR} proposed dropout technique that suggests randomly dropping neurons from a large NN. This creates ``thinned'' NN instances during training and prevents a NN from overfitting. Our proposed BNeuralT randomly generates a tree architecture, which can be considered a sparse NN in a similar sense with rather a higher dropout. Also, the branching and pruning of the tree branches in BNeuralT are performed at the tree generation stage, where a branch is probabilistically  pruned by generating a leaf node at a depth lower than terminals.

    \section{Backpropagation neural tree}
    \label{sec:bpnt}
    %%%%%%%%%%%   NEURAL TREE MAIN ALGORITHM    %%%%%%%%%%%%%%%
    \subsection{Problem statement} 
    \label{sec:problem_stmt}
    Let $ \mathcal{X} \in \mathbb{R}^d$ be an instance-space and $ \mathcal{Y} = \{c_1, \ldots,c_r \} $ be a set of $ r $ labels such that a label $ y \in \mathcal{Y} $ is assigned to an instance $ \textbf{x} \in \mathcal{X} $. Therefore, for a training set of $N$ instance-label pairs $ \mathcal{S} = \left(\textbf{x}_i, y_i \right)_{i=1}^{N} $, we induce a classifier $ \mathcal{G}(\mathcal{X}, \mathbf{w}) $ that reduces classification cost 
    $\mathcal{L}_{\mbox{Error}}(\mathcal{G}) =  \sfrac{1}{N}\sum_{i=1}^{N} (\hat{y}_i \ne y_i)$, 
    where $ \hat{y}_i $ is a predicted output class on an input instance $\textbf{x}_i = \left\langle x^i_1, x^i_2, \ldots, x^i_d \right\rangle$ labeled with the target class $ y_i \in \{c_1, \ldots,c_r \}$. Additionally, when an instance $ \textbf{x} \in \mathcal{X} $ is associated with a continuous variable $ y \in \mathbb{R} $ rather than  a set of  discrete class labels, then $ \mathcal{G}(\mathcal{X}, \mathbf{w}) $ is a predictor that for a training set of instance-output pairs $ \mathcal{S} $ reduces a prediction cost like mean squared error (MSE) $\mathcal{L}_{\mbox{MSE}}(\mathcal{G}) =  \sfrac{1}{N}\sum_{i=1}^{N} (\hat{y} - y_i)^2.$
    
    \subsection{Backpropagation neural tree algorithm}
\begin{figure}
        \centering
        \subfigure[{Classification neural tree}]
        {\includegraphics[width=0.58\linewidth]{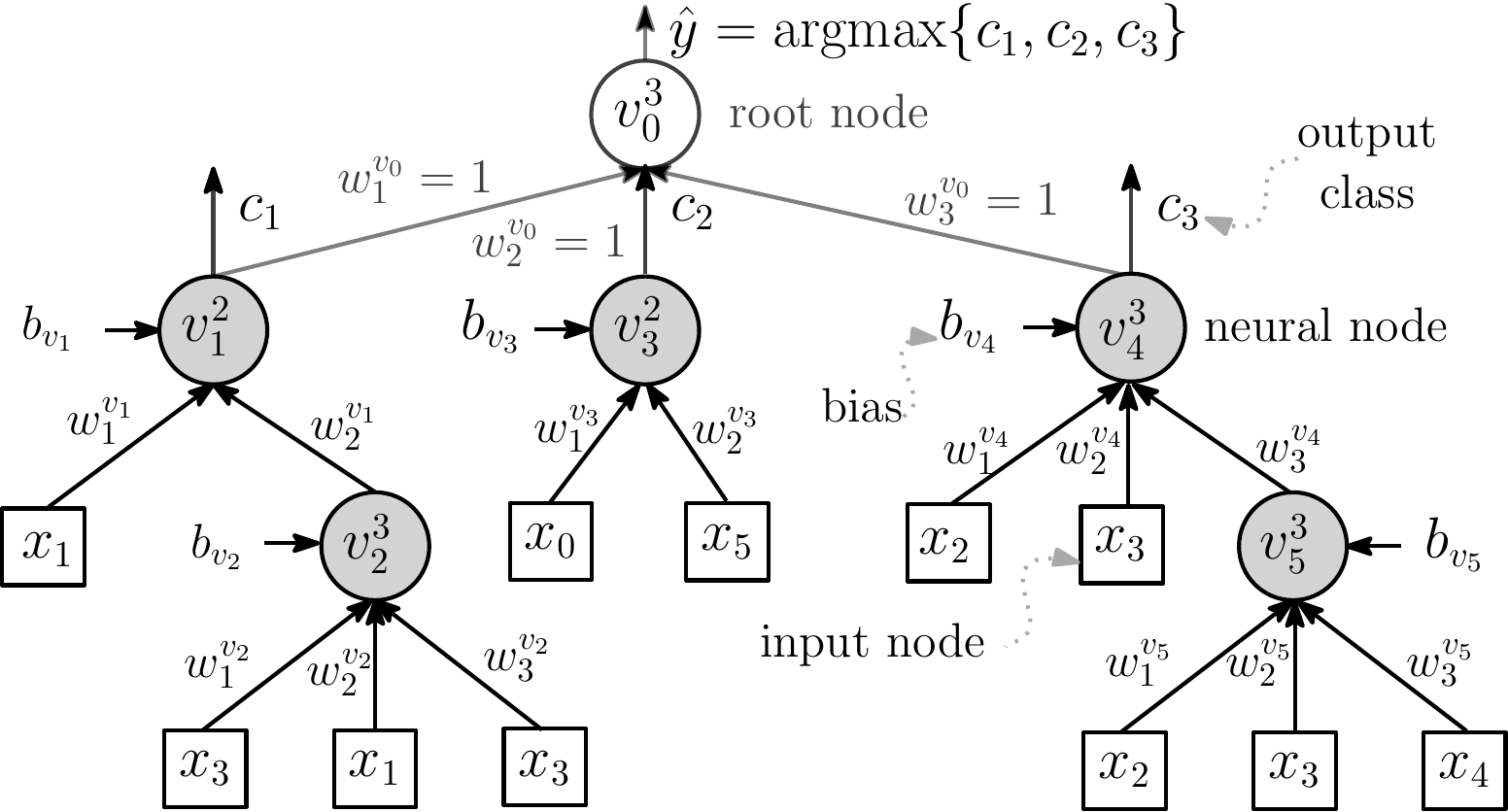}}
        \subfigure[{Regression neural tree}]
        {\includegraphics[width=0.38\linewidth]{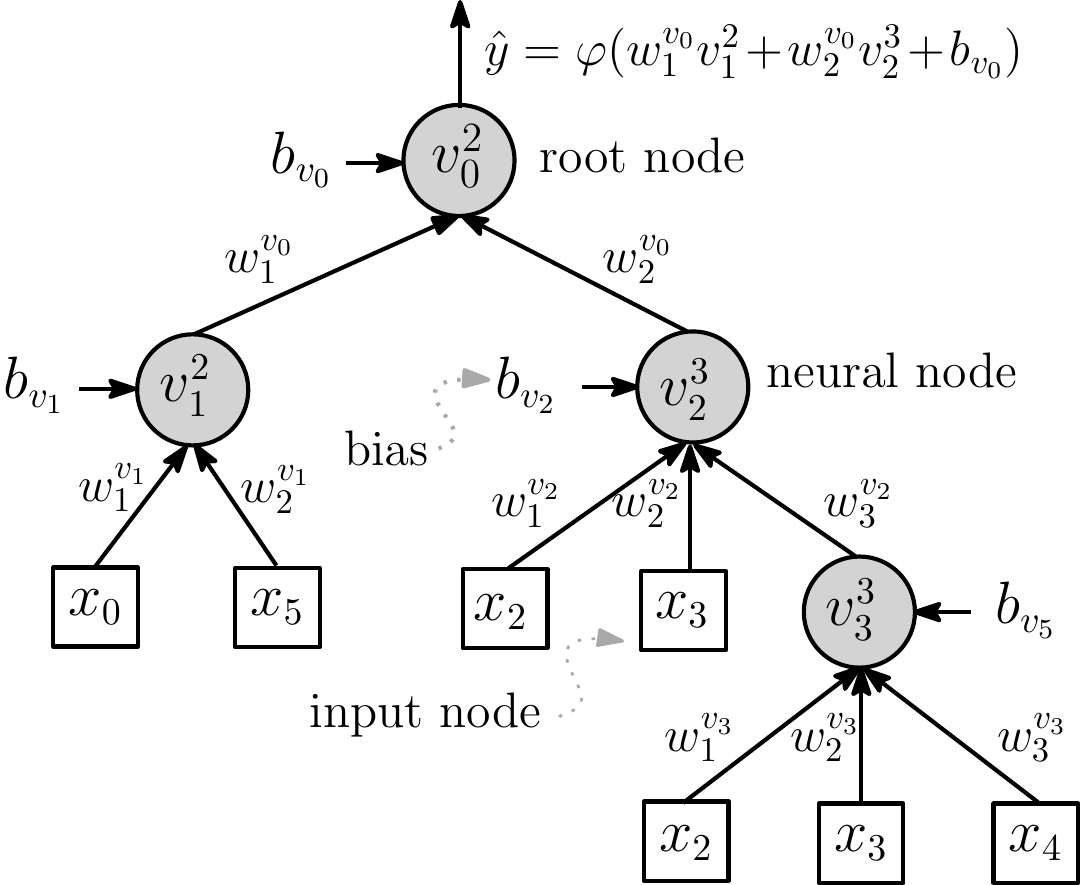}}
        \caption{Neural Trees: (a) A neural tree example of a three-class classification learning problem. The root node $ v^3_0 $ takes three immediate children: $ v_1, v_3, $ and $ v_4 $, each respectively designated to a class $ c_1, c_2, $ and $ c_3 $. The internal nodes (shaded in gray) are neural nodes and take an activation function $ \varphi(\cdot) $ and leaf nodes are inputs. Each designated output class has its subtree. This tree takes its input from the set \{$x_1, x_2, \ldots, x_5$\}. The link $ w_i^{v_j} $ between nodes are neural weights. (b) A neural tree example for a regression  problem has one output node $ v_0 $.}
        \label{fig:neural_tree}
    \end{figure}

    Backpropagation neural tree (BNeuralT) takes a tree-like architecture whose root node is a decision node, and leaf nodes are inputs. For classification learning problems, BNeuralT has strictly dedicated nodes at level-1 (child nodes of the root) of the tree to represent classes. Fig.~\ref{fig:neural_tree}(a) is an example of a \textit{classification neural tree} where root's each immediate child is a sub-tree dedicated to a class, and the root node only decides a winner class $ \hat{y} = \text{argmax}\{c_1, \ldots,c_r \} $ for an instance-label pair $ (\textbf{x}, y)$. For regression learning problems, BNeuralT is a \textit{regression neural tree} whose root node decides the tree's predicted output $ \hat{y} = \varphi(\sum_{i=1}^{child} w_i v_i + b_{v_0}) $, where $ \varphi(\cdot) $ is an activation function yielding a value in $[0, 1]$, $w_i$ is a edge weight, $v_i$ is the activation of $i$-th child, and $b_{v_0}$ is the root's bias (cf. Fig.~\ref{fig:neural_tree}(b)).
    
    BNeuralT, denoted as $ \mathcal{G} $ is an \textit{$ m $--ary} rooted tree with its one node designated as the root node, and each node takes at least $ m \ge 2$ child nodes except for a leaf node that takes no child node. Hence, for a tree depth $ p $, BNeuralT takes $ n \le [(m^{p+1}-1)/(m-1)]$ nodes (including the number of internal nodes $ |V| $ and the leaf nodes $|T|$). Thus, BNeuralT can be defined as a union of internal and leaf nodes    
    %\begin{equation*}
    %\label{eq_mont}
    $
    \mathcal{G} = V \cup T = \left\lbrace v^j_1,v^j_2,\ldots,v^j_K \right\rbrace  \cup \left\lbrace t_1, t_2,\ldots, t_L \right\rbrace 
    $
    %\end{equation*}
    where $ k $-th node $ v^j_k \in V$ is an internal node and receives $ 2 \le j \le m $ inputs from its child nodes. The $ k $-th leaf node $ t_k \in T$ has no child, and it has a designated input $ x_i \in  \{x_1,x_2,\ldots,x_d\}$. 
    
    Fig.~\ref{fig:neural_tree} is an example of classification (left) and regression (right) trees. All internal nodes (shaded in gray) of the tree are neural nodes and may behave like the nodes of a NN. That is, a neural node computes a weighted summation $z$ of inputs and squashes that using an activation function $ \varphi(z)$, e.g., \textit{sigmoid}: $ \varphi(z) = \sfrac{1}{(1 + e^{z})} $ or \textit{ReLU}: $ \varphi(z) = \max(0,z) $. We installed sigmoid or ReLU functions as BNeuralT's neural nodes, which can be any other activation function like \textit{tanh}. The trainable parameters $ \textbf{w} $ are the edges and the bias weights of the nodes. The number of nodes $ n $ in a tree grows as per $ \mathcal{O}(m^p) $. The number of edges $(n-1) $ is proportional to the growth of $ n $, so is the number of tree's \textit{trainable parameters} $ \textbf{w} $.
    
    \textbf{Complexity of BNeuralT.} A BNeuralT model resembles an expression tree, and its computation is a depth-first-search post-order or pre-order traversal where each node needs to be visited at least once. Hence, the worst-case \textit{time complexity} of BNeuralT is $ \mathcal{O}(n) $, $n$ being the number of nodes. In a BNeuralT model, each internal node has a bias, each leaf has an input, and each edge has a weight. Therefore, the space requirement of a BNeuralT model is $2|V|+|T|$, i.e., two times internal nodes plus leaf nodes, which will grow  proportional to the growth of tree's total nodes $n = |V|+|T|$. Thus, BNeuralT's worst-case \textit{space complexity} is $ \mathcal{O}(n) $.

\begin{figure}
        \centering
        \subfigure[{\scriptsize \cite{travis2005regional}}]{
            % Trivis et al. 2005
            \includegraphics[width=0.2\linewidth]{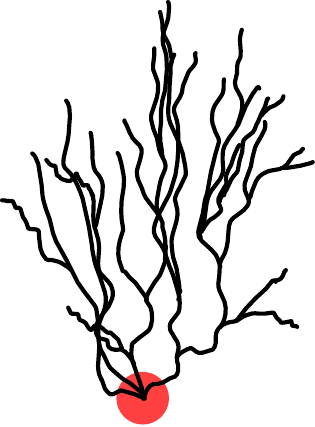}\label{sfig:bio_tree_a}}
        \qquad~\quad
        \subfigure[{\scriptsize \cite{jones2021might}}]{ 
            %Ilenna et al. 2021
            \includegraphics[width=0.25\linewidth]{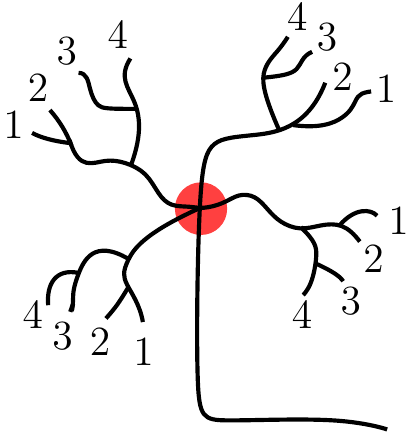}\label{sfig:bio_tree_b}}
        \qquad~\quad
        \subfigure[{\scriptsize BNeuralT}]{
            \includegraphics[width=0.3\linewidth]{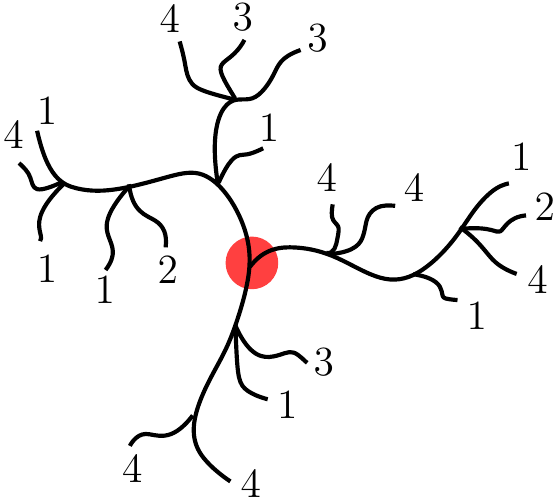}\label{sfig:bio_tree_c}}
        \caption{Biologically plausible neural computation using \textit{dendritic trees}. Red circle represents a neuron (\textit{soma}), black lines are dendrites, and the numbers indicate inputs.
            \label{fig:bio_trees}}
    \end{figure}

    \textbf{Biologically plausible neural computation of BNeuralT.} A typical NN uses \cite{mcculloch1943logical} neurons. Such a neuron operates on a weighted sum of inputs and processes the sum via a nonlinear threshold function. Such neural computation considers that the dendrites (synaptic inputs) of a neuron are summed at ``soma,'' thereby exciting a neuron, i.e., providing it a firing strength~\citep{mcculloch1943logical,hodgkin1952quantitative,poirazi2003arithmetic}. However, the biological behavior of dendrites shows that dendrites themselves impose nonlinearity on their synaptic inputs before summing at ``soma''~\citep{london2005dendritic,hay2011models}. This dendritic nonlinearity is possibly a sigmoidal nonlinearity~\citep{poirazi2003pyramidal}. Additionally, the synaptic connections in a fully connected NN are symmetric, whereas biological dendritic connections are asymmetric~\citep{mel2016toward,travis2005regional,farhoodi2018sampling} [cf. Fig.~\ref{fig:bio_trees}(a)].% shows asymmetric dendritic connection to a single neuron. 
        
    \cite{poirazi2003pyramidal} using a sparse two-layer NN analogous to a binary tree-like  dendritic NN has shown the possibility of modeling a single neuron as a NN. The work of  \cite{beniaguev2020single} demonstrated a proof of concept single neuron model as a synaptic \textit{integration and fire} model capable of performing classification of two types of classes with a high degree of temporal accuracy. \cite{jones2021might} considered the biologically asymmetric morphology of ``\textit{dendritic tree}'' and its repeated synaptic inputs to a neuron to show the computational capability of a \textit{single neuron} for solving machine learning problems. 
        
    Fig.~\ref{fig:bio_trees}(b) shows \cite{jones2021might}'s a single neuron computational model with repeated inputs as a binary tree structure.  Unlike the work of \cite{poirazi2003pyramidal}, BNeuralT has asymmetric \textit{dendritic connections} to a \textit{``single'' neuron} \citep{beniaguev2020single,jones2021might}. \cite{jones2021might}'s dendritic tree has a systematic and regular binary-tree-like structure and solves a binary classification problem. Whereas BNeuralT's neuron is like the neuron of \cite{travis2005regional} (cf. Fig.~\ref{fig:bio_trees}(a)), and it has a stochastic $m$-ary rooted tree-like structure (cf. Fig.~\ref{fig:bio_trees}(c)). Thus, through BNeuralT we investigate the ability of a single neuron with sigmoidal nonlinearity (and linear when using ReLU) in its dendritic connections on three machine-learning problems: multi-class classification, regression, and pattern recognition. %% end insert
    
    \textbf{Stochastic gradient descent (SGD) training.}
    BNeuralT's trainable parameters $ \textbf{w} $ are iteratively optimized by a stochastic gradient descent (SGD) method (cf. Algorithm~\ref{algo:sgd}) that at an iteration $ j $ requires gradient $\nabla \mathbf{w}_j \leftarrow $ \textsc{Gradient}$ \nabla_\mathbf{w} \mathcal{L}(\mathbf{x}_j, \mathcal{G}_{\mathbf{w}_j})$ computation (cf. Algorithm~\ref{algo:gradient_compute}) and weight update as per $ \mathbf{w}_j \leftarrow \mathbf{w}_{j-1} + \eta \nabla \mathbf{w}_j$. The weight update $ \mathbf{w}_j \leftarrow \mathbf{w}_{j-1} + \eta \nabla \mathbf{w}_j$ in line number 7 of Algorithm~\ref{algo:sgd} is a simple GD method, where other similar optimizers like MGD, NAG, Adagrad, RMSprop, or Adam can also be used. Table~\ref{tab:sgds_mehods}  details the expressions of weight updates for these optimizers.  
    \begin{algorithm}[h!]
        \caption{Stochastic gradient descent (SGD)}
        \label{algo:sgd}
        \begin{algorithmic}[1] 
            \Procedure{SGD}{$ \mathcal{G}_\mathbf{w}$, $\mathcal{S}$} \Comment{SGD($\cdot,\cdot$) takes a neural tree $ \mathcal{G}_\mathbf{w}$ and training data $ \mathcal{S} $}
            \State $ \mathbf{w}_0 \leftarrow \mathcal{G}_\mathbf{w}$ \Comment{initial weights}
            \For {epoch$_i $ $< $  epoch$_{\text{max}}$}
            \State $\mathcal{S} \leftarrow$ \textsc{Shuffle}$(\mathcal{S})$ \Comment{function \textsc{Shuffle} returns a randomly shuffle dataset}
            \For{an instance $ \mathbf{x}_j \in \mathcal{S}$}
            \State $ \Delta \mathbf{w}_j \leftarrow $ \textsc{Gradient} $ \nabla_\mathbf{w} \mathcal{L}(\mathbf{x}_j, \mathcal{G}_{\mathbf{w}_j})$ \Comment{computed as per Algorithm~\ref{algo:gradient_compute}} 
            \State $ \mathbf{w}_j \leftarrow \mathbf{w}_{j-1} + \eta\Delta \mathbf{w}_j$ \Comment{GD  weights update on an input instance $ \mathbf{x}_j $}
            \EndFor
            \EndFor    
            \EndProcedure  
        \end{algorithmic}
    \end{algorithm}
    
    \begin{table}
        \centering
        \setlength{\tabcolsep}{10pt}
        {\caption{Gradient descent versions to replace line number 7 in Algorithm~\ref{algo:sgd}. Symbols $ \eta, \gamma, \beta, \beta_1$, and $\beta_2 $ are constants (hyperparameter) of respective algorithms. Symbol $ v_j$ and $w_j$ show previous momentum and weights, respectively, and $ \nabla_\mathbf{w} \mathcal{L}(\mathbf{x}_j, \mathcal{G}_{\mathbf{w}_j}) $ shows gradient of loss $ \mathcal{L} $ over input $ \mathbf{x}_j $ and $ \textbf{w} $ of tree $ \mathcal{G} $.}
            \label{tab:sgds_mehods}}
        \begin{tabular}{ll}
            \toprule
            Algorithm & Expression\\ 
            \midrule
            MGD~\citep{qian1999momentum}  &  
            %\parbox{7cm}{
            %\begin{eqnarray}
            %\label{eq:Momentum_gd}
            $v_j \leftarrow \gamma v_{j - 1} + \eta \nabla_\mathbf{w} \mathcal{L}(\mathbf{x}_j, \mathcal{G}_{\mathbf{w}_j})$ \\
            & $\mathbf{w}_j \leftarrow \mathbf{w}_{j-1} - v_j $%~\\
            %\end{eqnarray} 
            \\[7pt]
            %\midrule
            Nesterov accelerated  GD~\citep{bengio2013advances}  & 
            %\parbox{7cm}{
            %\begin{eqnarray}
            %\label{eq:nag}
            $v_j \leftarrow \gamma v_{j - 1} + \eta \nabla_\mathbf{w} \mathcal{L}(\mathbf{x}_j, \mathcal{G}_{\mathbf{w}_j - \mathbf{w}_{j-1}})$ \\
            & $\mathbf{w}_j \leftarrow \mathbf{w}_{j-1} - v_j$%~\\
            %\end{eqnarray}
            %}
            \\[7pt]
            %\midrule
            Adagrad~\citep{dean2012large} & 
            %\parbox{7cm}{
            %\begin{eqnarray}
            %\label{eq:Adagrad}
            $v_j \leftarrow v_{j - 1} + \nabla_\mathbf{w} \mathcal{L}(\mathbf{x}_j, \mathcal{G}_{\mathbf{w}_j})^2 $ \\ 
            & $\mathbf{w}_j \leftarrow \mathbf{w}_{j-1} - \sfrac{\eta}{\sqrt{v_j + \epsilon}}  \nabla_\mathbf{w} \mathcal{L}(\mathbf{x}_j, \mathcal{G}_{\mathbf{w}_j})$%~\\
            %\end{eqnarray}
            %}
            \\[7pt]
            %\midrule
            RMSprop~\citep{tieleman2012lecture} & 
            %\parbox{7cm}{
            %\begin{eqnarray}
            %\label{eq:RMSprop}
            $v_j \leftarrow (1-\gamma)v_{j - 1} + \gamma  \nabla_\mathbf{w} \mathcal{L}(\mathbf{x}_j, \mathcal{G}_{\mathbf{w}_j})^2 $ \\
            & $\mathbf{w}_j \leftarrow \mathbf{w}_{j-1} - \left(\sfrac{\eta}{\sqrt{v_j + \epsilon}}\right)  \nabla_\mathbf{w} \mathcal{L}(\mathbf{x}_j, \mathcal{G}_{\mathbf{w}_j})$%~\\
            %\end{eqnarray}
            %}
            \\[7pt]
            %\midrule
            Adam~\citep{diederik2015adam} & 
            %\parbox{7cm}{
            %\begin{eqnarray}
            %\label{eq:Adam}
            $m_j \leftarrow \sfrac{\beta_1 m_{j - 1} + (1 - \beta_1)  \nabla_\mathbf{w} \mathcal{L}(\mathbf{x}_j, \mathcal{G}_{\mathbf{w}_j})}{\left(1 - \beta^j_1\right)} $ \\
            & $v_j \leftarrow \sfrac{ \beta_2 v_{j - 1} + (1 - \beta_2)   \nabla_\mathbf{w} \mathcal{L}(\mathbf{x}_j, \mathcal{G}_{\mathbf{w}_j})^2}{\left(1 - \beta^j_2\right)} $ \\
            & $\mathbf{w}_j \leftarrow \mathbf{w}_{j-1} - \sfrac{\eta}{\left(\sqrt{v_j + \epsilon}\right)} m_j$
            %\end{eqnarray}
            %}
            \\
            \bottomrule
        \end{tabular} 
    \end{table}
    
\begin{figure}
    	\centering
    	\includegraphics[width=0.9\linewidth]{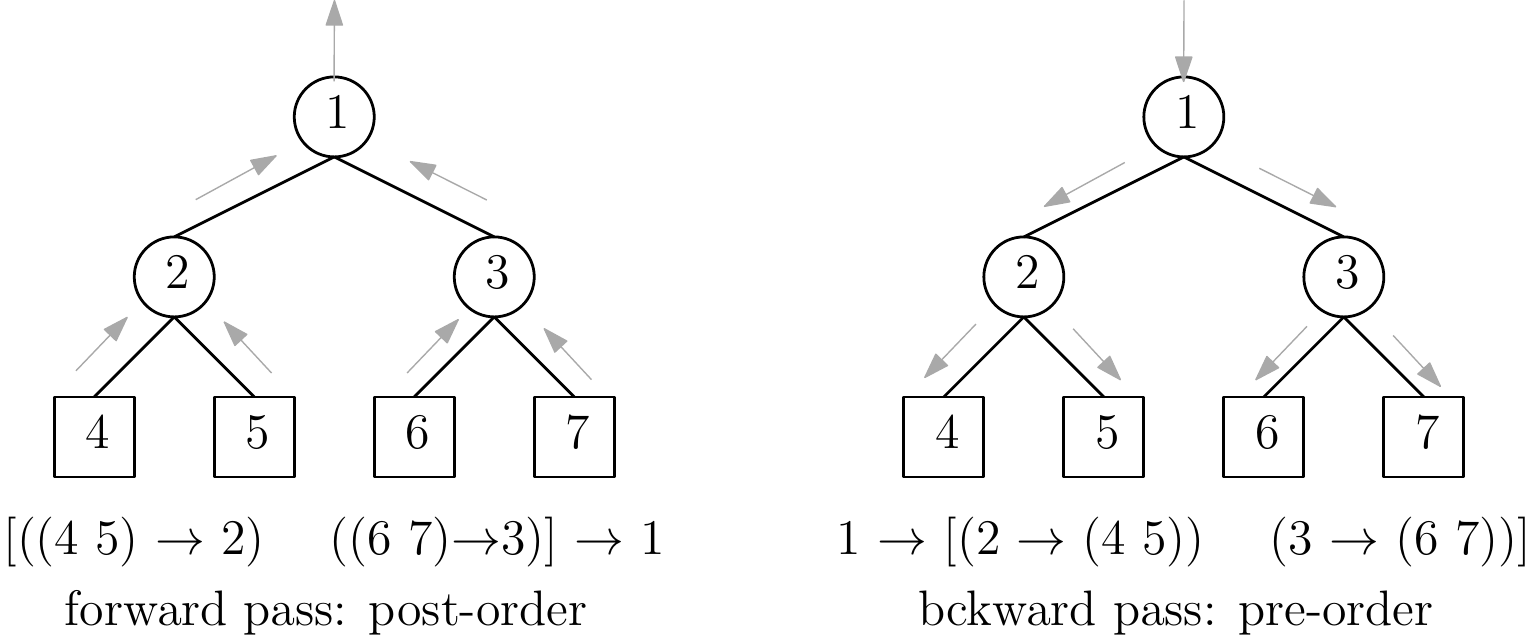}
    	\caption{Forward pass (left) and backward pass (right) computation. The arrows show the direction of computation.
    		\label{fig:tree_traversal}}
    \end{figure}

    \textbf{Error-backpropagation in BNeuralT.}
    Our proposed \textit{recursive error-backpropagation} in BNeuralT algorithm has two computation phases: \textit{forward pass} and \textit{backward pass} (cf. Fig.~\ref{fig:tree_traversal}). Both work in a \textit{depth-first search} manner. Since a tree data structure is algorithmically recursive to traverse through, both forward pass and backward (error-backpropagation) pass take place in a recursive manner. The forward pass computation produces the output for a tree in a \textit{post-order traversal} manner (cf. Fig.~\ref{fig:tree_traversal}(left)). That is, each leaf node propagates its input through dendrite (edge) to its parent node, and subsequently, each internal node, after computing received inputs from its child nodes, propagates activation to their respective parent node. Finally, the root node computes the tree's output.
    
\begin{figure}
        \centering
        \includegraphics[width=0.49\linewidth]{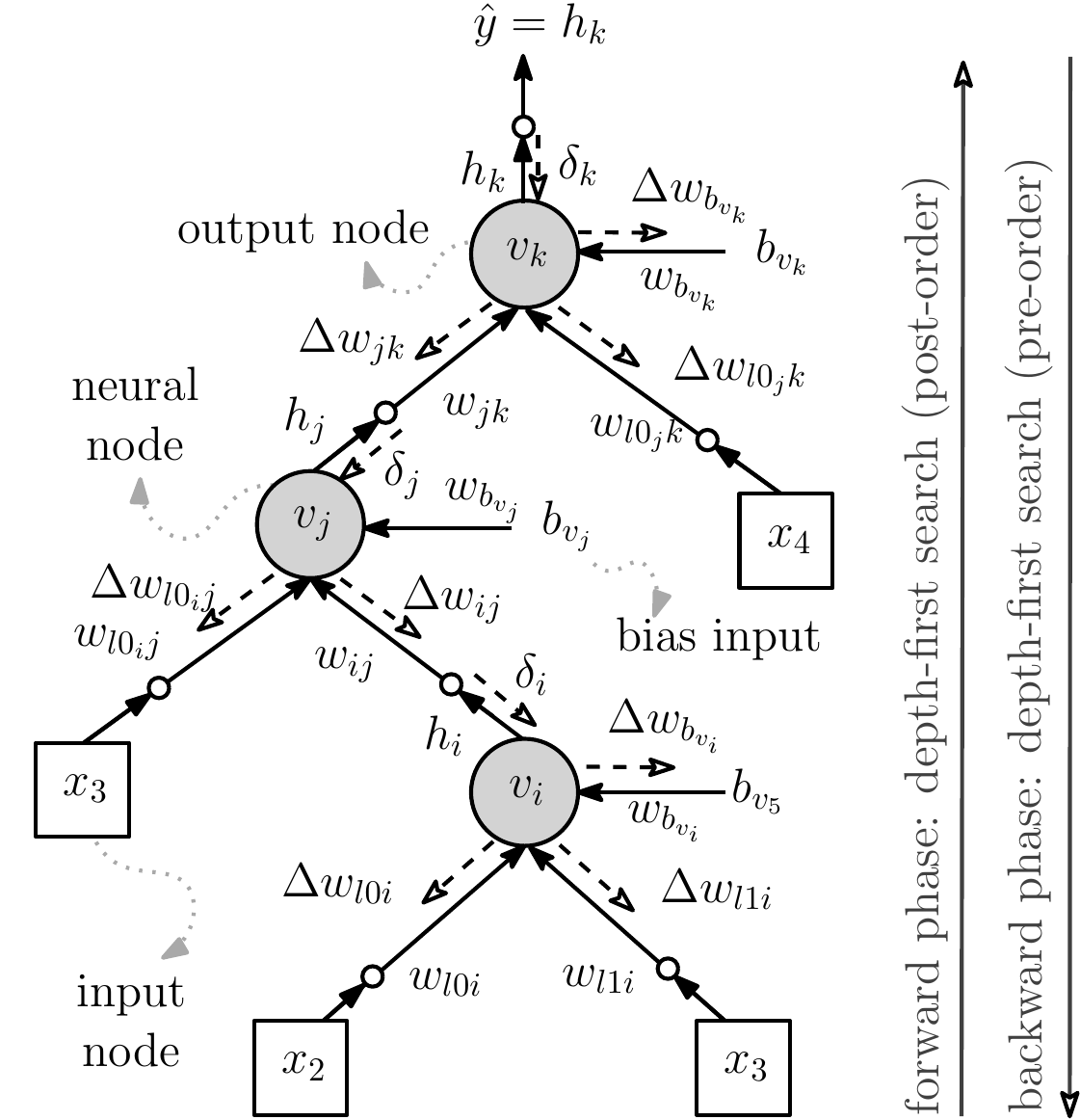}	\includegraphics[width=0.49\linewidth]{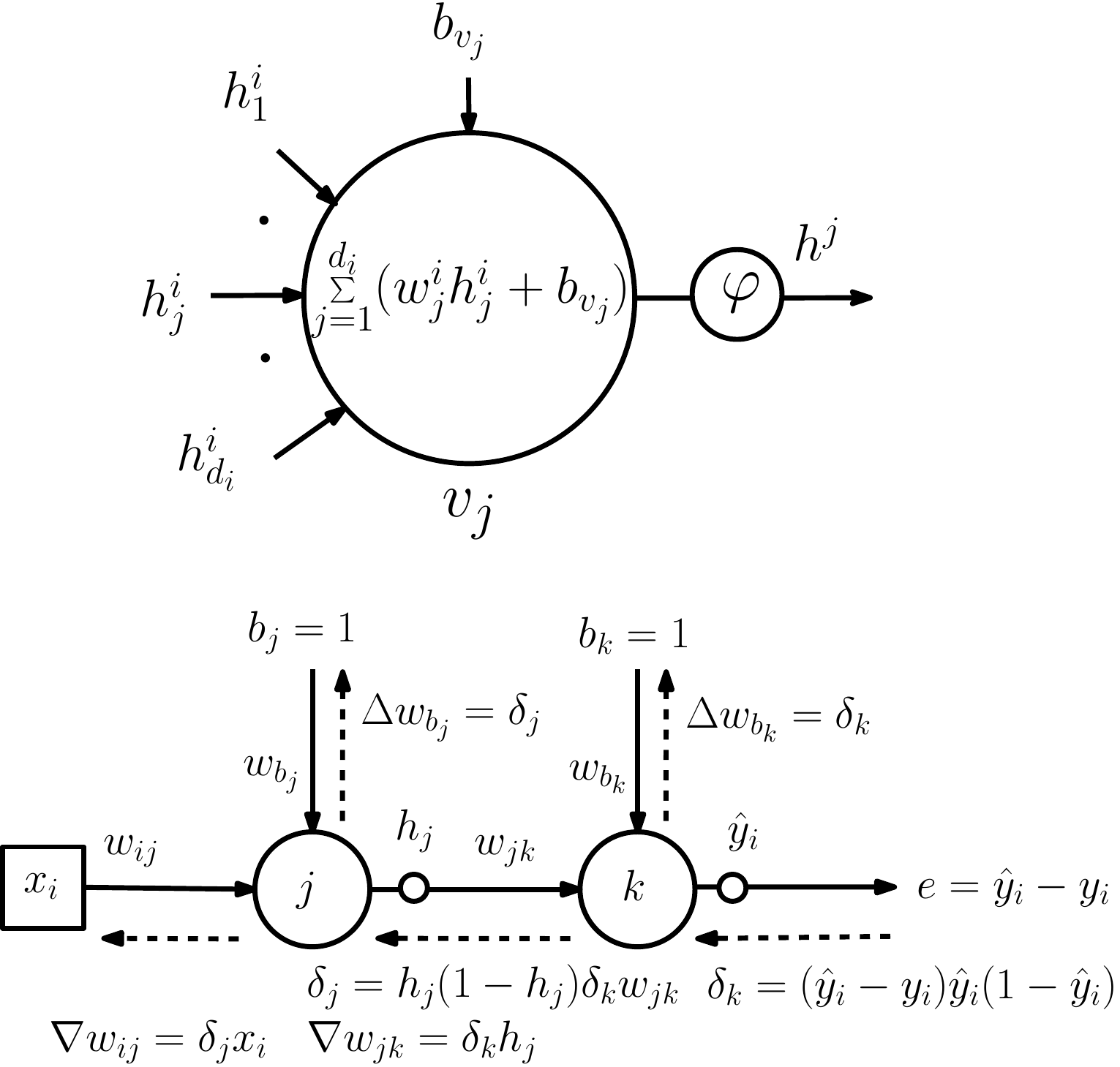}
        %%%
        \caption{\textit{Left.} Backpropagation Neural Tree.  Output node $ v_k $ yields output $ y $ using forward pass upon receiving inputs $ x_i $ from leaf nodes. Each node is linked with an edge weight $ w_{ij} $. The backward pass propagates error $ e = (y - \hat{y}) $ back to input nodes to compute weight change $ \Delta w$. \textit{Right.} Backpropagation of error from an output node, $ k $; to a hidden node, $ j $; to an input node, $ i $; and to bias inputs $ b_k $ and $ b_j $. Dashed lines represent error backpropagation  and computation $ \delta $ and gradient $ \nabla w $ (cf. Algorithm~\ref{algo:gradient_compute}) to find weight change $ \Delta w $ that help stochastic gradient descent (cf. Algorithm~\ref{algo:sgd}).
        \label{fig:BNeuralT}}
        %%%%%%%
        %This example is an error backpropagation for sigmoid activation function.} %; and (c) An internal (neural) node of a neural tree.
        %\end{figure}	
        %\begin{figure}{R}{0.5\textwidth}	
        %	\centering
        %	\includegraphics[width=0.5\textwidth]{fig/nodes_gradient_bold}
        %%\caption{Backpropagation of error from an output node $ k $ to a hidden node $ j $ to an input node $ i $ and to the bias inputs $ b_k $ and $ b_j $. Dashed lines represent error backpropagation  and computation $ \delta $ and gradient $ \nabla w $ to find weight change $ \Delta w $ (Algorithm~\ref{algo:gradient_compute}). This example is an error backpropagation for sigmoid activation function.}
        %%	\label{fig:back_prop_ex}
    \end{figure}

    The backward pass computes the gradient of the error with respect to edge weights. The backward pass computes gradient $ \delta $ for each internal node and propagates it back to each edge depth-by-depth. Hence, the backward pass is a \textit{pre-order traversal} of the tree (cf. Fig.~\ref{fig:tree_traversal}(right)). That is gradient $ \delta $ computed at the root node flow backward to its child node until it reaches leaf nodes.
    Fig.~\ref{fig:BNeuralT} (left) shows the forward pass and backward pass computation labeled with variables of neural tree computation. Fig.~\ref{fig:BNeuralT} (right) shows the backpropagation of gradient from an output node to the inputs (leaf nodes). Algorithm~\ref{algo:gradient_compute} is a summary of error-backpropagation and Algorithm~\ref{algo:delta_compute} shows recursive gradient computation for BNeuralT that facilitates error-backpropagation.
    
    %{\color{red}
    %\lipsum[15-15]
    %
    %\lipsum[5-5]
    %}
    \begin{algorithm}[h!]
        \caption{Backpropagation computation of neural tree $ \mathcal{G} $}
        \label{algo:gradient_compute}
        \begin{algorithmic}[1]            
            \Procedure{Gradient $ \nabla \mathbf{w} \mathcal{L}$}{$ \mathbf{x}, \mathcal{G}$}\Comment{$ \mathbf{x} $ and $ \mathbf{w} $ are the inputs and trainable parameters}
            \State $ \mathcal{G}_{\delta} \leftarrow $\Call{Compute $ \delta$ }{$ y $, $  \mathcal{G}(\mathbf{x} )$, $N_0 \leftarrow \mathcal{G} $, $ \mathcal{G} $} \Comment{Compute $ \delta $ using Algorithm~\ref{algo:delta_compute} for target $ y $ and prediction $ \mathcal{G}(\mathbf{x} )$ at tree's root $ N_0 $}
            \For {all nodes $ N $ in $ \mathcal{G}_{\delta} $ }
            \If{$N\rightarrow Type \rightarrow$ Output node}  \Comment{output node's bias weight update}
            \State {$N\rightarrow \nabla w_{b_k} = N\rightarrow \delta_k $}  \Comment{bias weight is set to current node's gradient $ \delta $}
            \ElsIf{$N\rightarrow Type \rightarrow$ Internal (hidden) node}
            \State {$h_j = N\rightarrow h_j$}  \Comment{current node's activation is the incoming signal for weight $ w_{jk} $}
            \State {$\delta_k = N\rightarrow N_{Parent}\rightarrow \delta$}\Comment{gradient back-propagated from the parent node}
            \State {$N\rightarrow \nabla w_{jk} = \delta_k h_j$}\Comment{$ w_{jk} $ is the weight between node $ N_j $ and its parent node $ N_k $}
            \State {$N\rightarrow \nabla w_{b_j} = N\rightarrow \delta_j$}\Comment{bias weight is set to current node's gradient $ \delta $}
            \ElsIf{$N\rightarrow Type \rightarrow$ Leaf node} 
            \State {$x_i = N\rightarrow x$} \Comment{$ N\rightarrow x \in  \{x_1,x_2, \ldots, x_d\} $ is an input attribute}
            \State {$\delta_j = N\rightarrow N_{Parent}\rightarrow \delta$}\Comment{gradient back-propagated from parent to child}
            \State {$N\rightarrow \nabla w_{ij} = \delta_j x_i$}\Comment{$ w_{ij} $ is the weight between child $ N_i $ and parent $ N_j $}
            \EndIf
            \EndFor~\\\Return $ \nabla \mathbf{w} $
            \EndProcedure  
            %\item[]
        \end{algorithmic}
    \end{algorithm}
    
    \begin{algorithm}[h!]
        \caption{Computation of $ \delta $ for each neural node}
        \label{algo:delta_compute}
        \begin{algorithmic}[1]            		
            \Procedure{Compute $ \delta$ }{$ y $, $ \hat{y} $,  $ N \leftarrow \mathcal{G} $, $ \mathcal{G} $}\Comment{$ y $ is a target, $ \hat{y} $ is a prediction, and $ N $ is the current (or entry) node of tree $ \mathcal{G} $}
            \If{$N\rightarrow Type \rightarrow$ Output node}  \Comment{compute gradient $\delta_k$ for output node $ N $ in $ \mathcal{G} $}
            %\If{ $G.N_j.Function$ = sigmoid}
            \State {$N\rightarrow \delta_k = (\hat{y} - y) \hat{y} (1-\hat{y}) $}  \Comment{gradient at sigmoid output node}
            %\ElsIf{ $\mathcal{G}.N_j.Function$ = tanh}
            %\State {$\mathcal{G}.N_j.\delta_j = -1.0 (\hat{y}-d) \hat{y}^2 $}
            %\EndIf
            \Else  \Comment{compute gradient at an internal (hidden) neural node} 
            \State {$h_j = N\rightarrow h_j$}  \Comment{ activation (value) of an internal (neural) node}
            \State {$\delta_k = N\rightarrow N_{Parent}\rightarrow \delta_k$}  \Comment{retrieve parent's gradient $ \delta $ of the current node $ N $}
            \State {$w_{jk} = N\rightarrow w_{jk}$} \Comment{edge (weight) between current node $ N_j $ and parent node $ N_k $}
            %\If{ $G.N_j.Function$ = sigmoid}
            \State {$N\rightarrow \delta_j = h_j(1 - h_j)\delta_k w_{jk}$}  \Comment{gradient at an internal sigmoid node }
            %\ElsIf{ $\mathcal{G}.N_j.Function$ = tanh}
            %\State {$\mathcal{G}.N_j.\delta_j = -1.0 h_j^2 \delta_k w_{jk} $}
            \EndIf
            \For {all child (neural) node $ N_c $ of $ N $ in $ \mathcal{G} $}
            \State $ \mathcal{G}_{\delta} \leftarrow $\Call{Compute $ \delta$ }{$ \emptyset $, $ \emptyset $, $N_c \leftarrow \mathcal{G} $, $ \mathcal{G} $} \Comment{call \textsc{Compute} $\delta$, $\emptyset $ indicates unused argument } 
            \EndFor
            \EndProcedure       
        \end{algorithmic}
    \end{algorithm}

    \section{Experiments}
    \label{sec:exp_setup}
    %%%%%%%%%%%   EXPERIMENT    %%%%%%%%%%%%%%%
    %Experiment setup follows the descriptions of the dataset, BNeuralT and MLP algorithm hyper parameter setup, SGDs training parameter setup, training loss function and evaluation metric setup, algorithm implementation and computational environment setup.  
    \subsection{Hyperparameters settings}
    \label{sec:hyperparamter}
    \textbf{Datasets.} We select a set of nine \textit{classification problems}: Australia (Aus), Heart (Hrt), Ionosphere (Ion), Pima (Pma), Wisconsin (Wis), Iris (Irs), Wine (Win), Vehicle (Vhl), Glass (Gls), which respectively have 14, 13, 33, 8, 30, 4, 13, 18, and 9 input attributes; 2, 2, 2, 2, 2, 3, 3, 4, and 7 target classes; and 690, 270, 351, 768, 569, 150, 178, 846, and 214 examples. For \textit{regression problems}, we select Baseball (Bas), Daily Electricity Energy (Dee), Diabetes (Dia), Friedman (Frd), and Miles Per Gallon (Mpg), which respectively have 16, 6, 10, 5, and 6 input attributes and 337, 365, 442, 1200, and 392 examples. Each regression dataset has one target. These datasets are available at~\citep{lichman2013uci,keelDataSet}. These problems are significantly different not only in terms of the number of classes and examples but also in terms of their attribute types and range. This differing nature of these problems poses significant variations in difficulty for one algorithm to excel on all problems~\citep{wolpert1996lack}. 
    
    Both classification and regression learning datasets were \textit{normalized} using min-max normalization between $0$ and $1.$ Each dataset was randomly shuffled and partitioned into training ($80\%$) and test ($20\%$) sets for each instance of the experiment. 
    For a \textit{pattern recognition} problem, we select the MNIST dataset~\citep{mnistDataSet}, which has $60,000$ training examples and $10,000$ test examples labeled with a set of $10$ handwritten characters, and this dataset was normalized by dividing gray-scale pixel value by $255.$

    \textbf{BNeuralT hyperparameters.} We repeated experiments 30 times (independently) for each classification and regression problem. In each run, we generated ad hoc BNeuralTs (stochastically generated tree structures) for each dataset with a maximum tree depth $ p = 5$; max child per node $ m = 5$, and branch pruning factor $ P[\text{leaf}_{p} < p]  \in \{0.4,0.5\}$ which is a probability of a leaf node being generated  at a depth lower than the tree height $p$. A higher leaf generation probability (e.g., $0.5$) at internal nodes means that tree height terminates earlier than its predefined depth, which means a tree will be generated with fewer parameters. A lower leaf generation probability (e.g., $0.4$) means a deeper tree structure with more parameters.

    For classification problems, BNeuralT's output neural node was \textit{argmax} node (i.e., winner takes all node), whereas, for regression problems, it was \textit{sigmoid} activation function. Its internal nodes were \textit{sigmoid} activation (or \textit{ReLU} for some trial experiment versions). 
    For pattern recognition (MNIST dataset), BNeuralT models were generated on an ad hoc basis via setting maximum tree depth $ p \in \{5, 10\}$, and max child per node $ m \in \{10, 15\}$ with  $ P[\text{leaf}_{p} < p]  \in \{0.4, 0.5\}$ and a tree size threshold $ |\mathcal{G}|_{\ge \text{MIN}} \in \{1\text{K}, 20\text{K} \}$, where ``K'' is 1000. 
    
    \textbf{Other algorithms hyperparameter.}
    MLP architecture was a fixed three-layer architecture [inputs--hidden (100 nodes)--targets] for each dataset of classification and regression problems. A \textit{SoftMax} layer acted as the MLP classifier's output nodes, and an MLP regression had a \textit{sigmoid} activation as its output node. The internal neural nodes in an MLP were \textit{sigmoid} (or \textit{ReLU}) functions. Other algorithms HFNT$^{\text{S}}$, HFNT$^{\text{M}}$, MONT$_3$, DT, RF, GP, NBC, SVM, and CARTs had their default setups as they are in their libraries~\citep{pedregosa2011scikit} or in the literature~\citep{ojha2017ensemble,ojha2020multi,zharmagambetov2019experimental}. A detailed list of hyperparameters of all algorithms is provided in Supplementary Table~\ref{tab:parameters}.

    \textbf{SGD hyperparameters.} BNeuralT and MLP algorithms take \textit{optimizers} like GD, MGD, NAG,  Adagrad, RMSprop, or Adam. The training parameters were learning rate $ \eta = 0.1$, momentum rate $ \gamma = 0.9$, $ \beta_1 = 0.9 $, $ \beta_2 = 0.9 $, $\epsilon = 1e^{-8}$, training mode was \textit{stochastic} (online), and training epochs were $500.$ Since the gradient descent computation was stochastic, both BNeuralT and MLP do the same number of forward-pass (function) evaluations, i.e., number training examples $\times$ epochs. All six optimizers were used for training BNeuralT and MLP with an \textit{early-stopping} restore-best strategy (or without an \textit{early-stopping} for some trail experiments), whereas other algorithms take their own default optimizer~\citep{pedregosa2011scikit}. While BNeuralT and MLP were trained in \textit{online} mode (example-by-example training), other algorithms take only \textit{offline} mode (epoch-by-epoch) training.

    For the pattern recognition problem (MNIST), we set a  \textit{mini-batch} training with a batch size of 128 examples. RMSprop was used as an \textit{optimizer}, and BNeuralT was trained by varying learning rate $ \eta \in \{0.1, 0.01\}$ and the number of epochs $\in \{10, 25, 50, 70\}$. The results of other algorithms on MNIST were collected from literature to compare performances.
    
    \textbf{Loss functions.}
    The loss function for BNeuralT training for classification and pattern recognition problems was a miss-classification rate $\mathcal{L}_{\text{Error}}(\mathcal{G})$. MLP training on classification problems was best with categorical cross-entropy loss~\citep{bishop2006pattern}. The training of other algorithms had default setups recommended in their libraries~\citep{pedregosa2011scikit}. For regression problems, all algorithms were trained by reducing  $\mathcal{L}_{\text{MSE}}(\mathcal{\cdot})$. 
    The test metric for classification problems for all algorithms was a miss-classification rate $\mathcal{L}_{\text{Error}}(\mathcal{\cdot})$ and for regression problems, it was a regression fit $\mathcal{L}_{r^2}(\mathcal{\cdot}) = 1 - \sfrac{\sum_{i=1}^{N} (y_i - \hat{y}_i)^2}{\sum_{i=1}^{N} (y_i - \bar{y})^2}$ (Nash–Sutcliffe model efficiency coefficient) which gives a value between [$- \infty, 1$], where $\bar{y} $ is the mean of target $y$.
    
    \textbf{Forward pass computation time, $\tau$.} BNeuralT was implemented in Java 11, and MLP was implemented using TensorFlow and Keras libraries~\citep{kerasSequential}. Other algorithms were implemented using scikit-learn library~\citep{pedregosa2011scikit} in Python 3.5. The forward pass computation time, $ \tau $ is a wall-clock time on Windows 10 operating system with configuration x64 Intel i5-2400 CPU, 3.1GHz, 3101Mhz, 4 Cores, and 16GB physical memory.
    
    \subsection{BNeuralT and MLP experiment versions}
    \label{sec:exp_version}
    We experimented with multiple versions of BNeuralT and MLP settings to bring out the best of both. We tried \textit{sigmoid} and \textit{ReLU} as the internal activation functions. We tried BNeuralT's branch pruning factor $ P[\text{leaf}_{p} < p]$ with $0.5$ and $0.4$. 
    
    The learning rates of the optimizers had two sets: (i) A flat learning rate $\eta = 0.1$ for all optimizers. (ii) The learning rate recommended in the Keras library for respective optimizers, i.e., RMSprop, Adam, and Adagrad had $\eta=0.001$, and MGD, NAG, and GD had  $\eta=0.01$. We call the library's recommended $\eta$ value a \textit{default} learning rate. In addition, the SGD learning was tried ``with'' and ``without'' \textit{early-stopping} (ES) strategies.
    
    These variations produced five BNeuralT settings: (i) ES training of BNeuralT having \textit{sigmoid} nodes, $0.1$ learning rate, and $0.5$ leaf generation rate; (ii) ES training of BNeuralT having \textit{sigmoid} nodes, \textit{default} learning rate, and $0.5$ leaf generation rate; (iii) ES training of BNeuralT having \textit{ReLU} nodes, $0.1$ learning rate, and $0.5$ leaf generation rate; (iv) ES training of BNeuralT having \textit{sigmoid} nodes, $0.1$ learning rate, and $0.4$ leaf generation rate; and (v) without ES training of BNeuralT having \textit{sigmoid} nodes, $0.1$ learning rate, and $0.5$ leaf generation rate. 
    
    Multiple MLP settings were tried. Out of which, some best performing settings were: (i) ES training of MLP having \textit{sigmoid} nodes and $0.1$ learning rate; (ii) ES training of MLP having \textit{sigmoid} nodes and \textit{default} learning rate; (iii) ES training of MLP having \textit{sigmoid} nodes, $0.1$ learning rate, and L2-norm regularization; (iv) ES training of MLP having \textit{sigmoid} nodes, \textit{default} learning rate, and L2-norm regularization; and (v) experiment setting same as (i) but without ES; and (vi) experiment setting same as (ii) but without ES. Other trials were using dropout with and without early stopping. 
    
    For each algorithm (BNeuralT, MLP, HFNT$^{\text{S}}$, HFNT$^{\text{M}}$, MONT$_3$, DT, RF GP, NBC, and SVM), each optimizer (GD, MGD, NAG, Adagrad, RMSprop, and Adam), and each variation of hyperparameter setting, there were 110 experiments (cf. Tables~\ref{tab:All_experiments_A} and ~\ref{tab:All_experiments_B} in Supplementary). We repeated each experiment for each dataset for 30 independent runs, and their average performance on test sets was evaluated.
    
    \section{BNeuralT performance}
    \label{sec:res_dis}
    %%%%%%%%%%%   RESULTS    %%%%%%%%%%%%%%%
    \subsection{Selection of the best performing setting} 
    We selected the best performing setting based on the average test accuracy computed over 30 independent runs of BNeuralT, MLP, HFNT$^{\text{S}}$, HFNT$^{\text{M}}$, MONT$_3$, DT, RF, GP, NBC, and SVM  to report them in detail in this section. The best performing BNeuralT setting was the ``ES training of BNeuralT having \textit{sigmoid} nodes, $0.1$ learning rate, and $0.4$ leaf generation rate.'' The best MLP setting was ``ES training of MLP having \textit{sigmoid} nodes and \textit{default} learning rate.'' We found that HFNT$^{\text{S}}$, HFNT$^{\text{M}}$, DT, RF, GP, NBC, and SVM worked best with their recommended setting.
    
    We found that collectively on all classification and regression datasets, BNeuralT with \textit{sigmoid} nodes, $0.1$ learning rate, and $0.4$ leaf generation rate trained using RMSprop  performed the best among all experiment versions of all algorithms. This setting produced an average accuracy of $83.2\%$ across all datasets with an average of $222$ trainable parameters. This same setting also performed the best across all classification datasets among all algorithms, i.e., it produced an average accuracy of $89.1\%$ with an average of $261$ trainable parameters.  In fact, the top six best results over classification datasets were from BNeuralT settings. GP algorithm came 7th with an average classification accuracy of $86.79\%$. MLP with \textit{sigmoid} node and ES  training using MGD optimizer with \textit{default} learning performed 8th with an average accuracy of $86.78\%$ with an average $1970$ trainable parameters.
    
    MLP, however, performed slightly better on regression problems than the other algorithms. MLP with \textit{sigmoid} node and ES training using NAG optimizer with \textit{default} learning rate produced an average regression fit value of $0.775$. Whereas BNeuralT with \textit{sigmoid} nodes, $0.1$ learning rate, and $0.4$ leaf generation rate trained using RMSprop optimizer produced an average regression fit value of $0.727$. It is important to note that this performance of BNeuralT comes with a much lower average trainable parameter. BNeuralT used only $152$ trainable parameters compared to MLP that used $1041$ parameters. This means BNeuralT's performance comes with an order magnitude less parameter than MLP on both classification and regression tasks.
    
    Table~\ref{tab:BNeuralT_all_results} reports the details of each algorithm's best-performing settings. However, exhaustive lists of $110$ experiments, from which we selected these best performing settings, are provided in Supplementary Tables~\ref{tab:All_experiments_A} and \ref{tab:All_experiments_B}. 
    
    %% REMOVED A list of all BNeuralT's results from its all setting is separately provided as a supplementary in Table~\ref{tab:BNeuralT_All_experiments}.
    
    \subsection{BNeuralT models summary}
\begin{figure}
        \centering
        \subfigure[{\scriptsize australian (92\%, 63)}]{
            \includegraphics[width=0.32\linewidth]{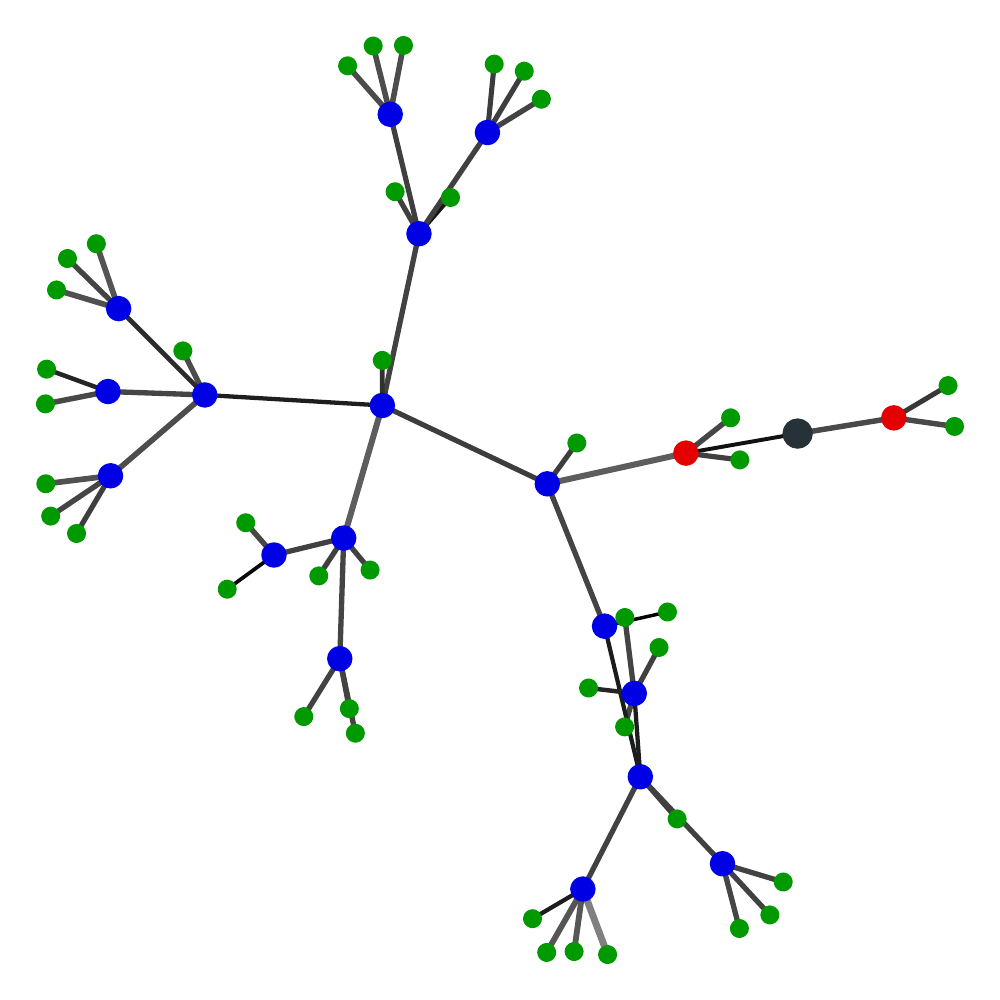}\label{sfig:img_aus}}
        \subfigure[{\scriptsize heart (96\%, 173)}]{
            \includegraphics[width=0.32\linewidth]{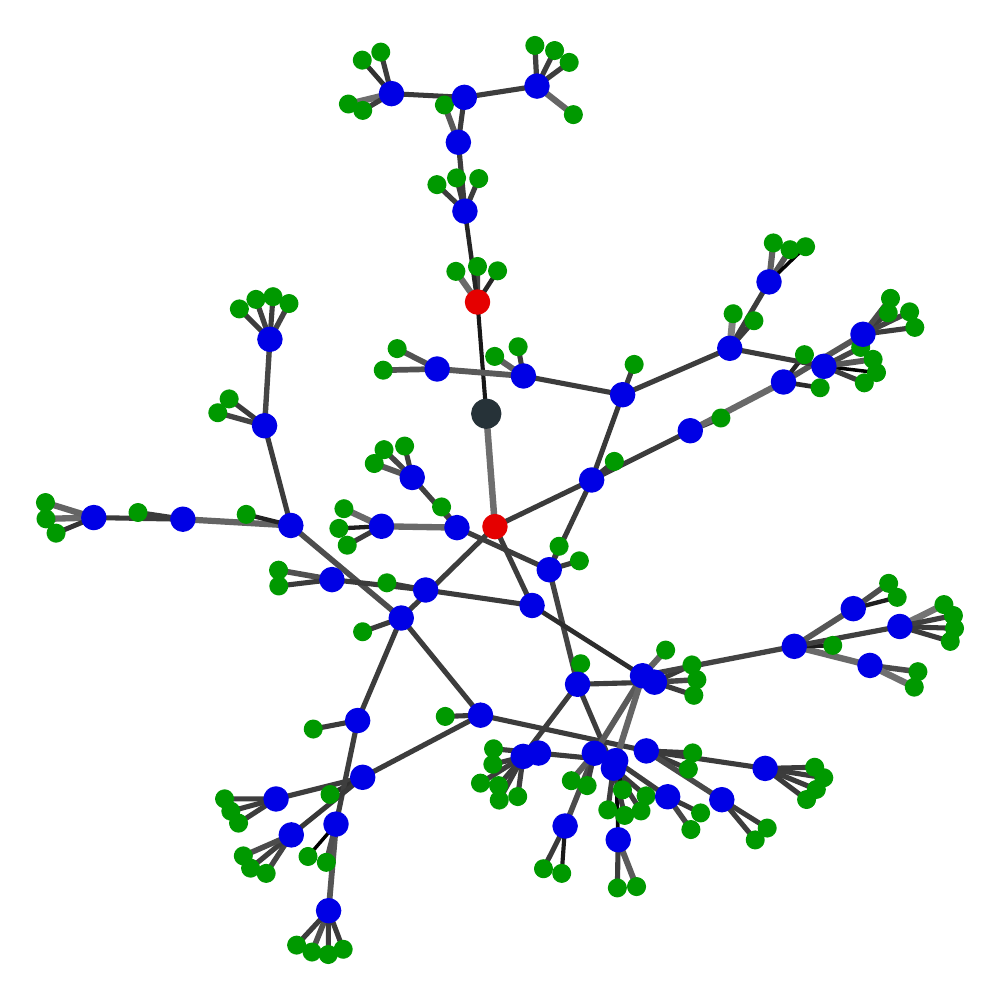}\label{sfig:img_hrt}}
        \subfigure[{\scriptsize ionosphere (99\%, 60)}]{
            \includegraphics[width=0.32\linewidth]{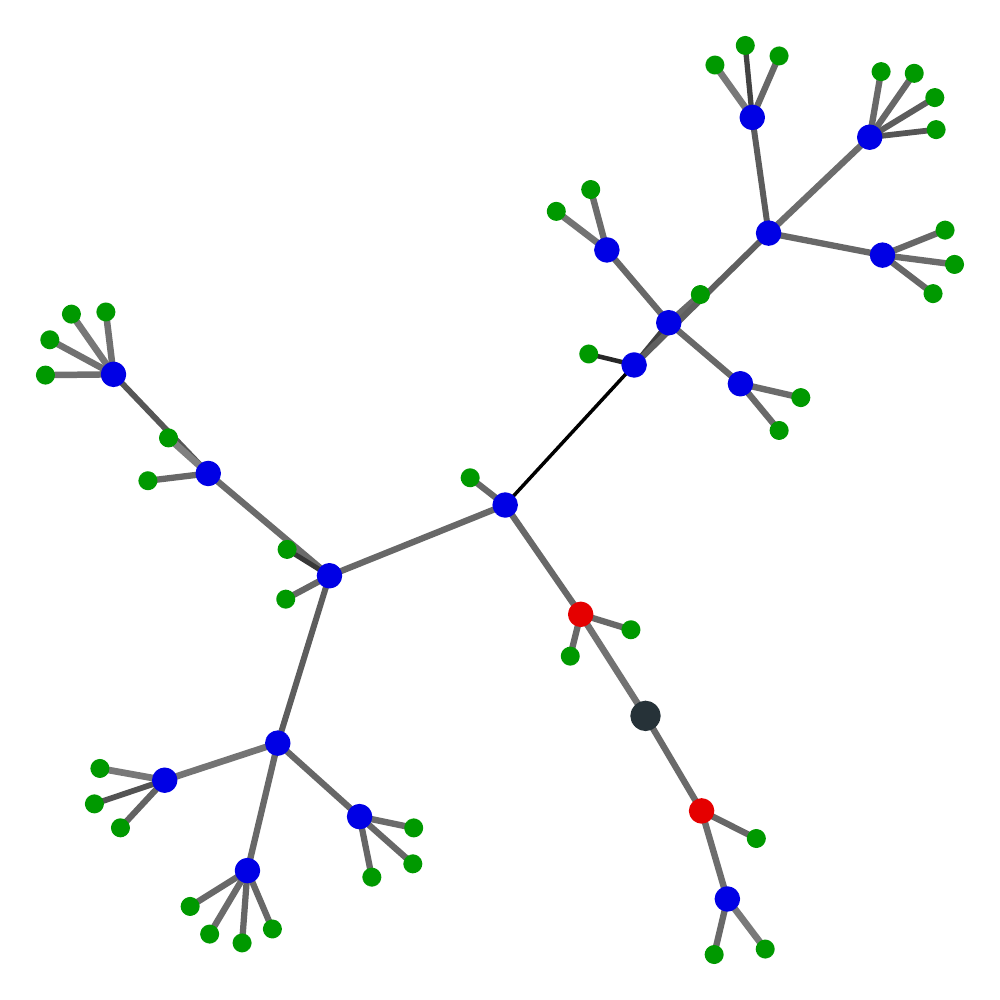}\label{sfig:img_ion}}

        \subfigure[{\scriptsize pima (87\%, 125)}]{
            \includegraphics[width=0.32\linewidth]{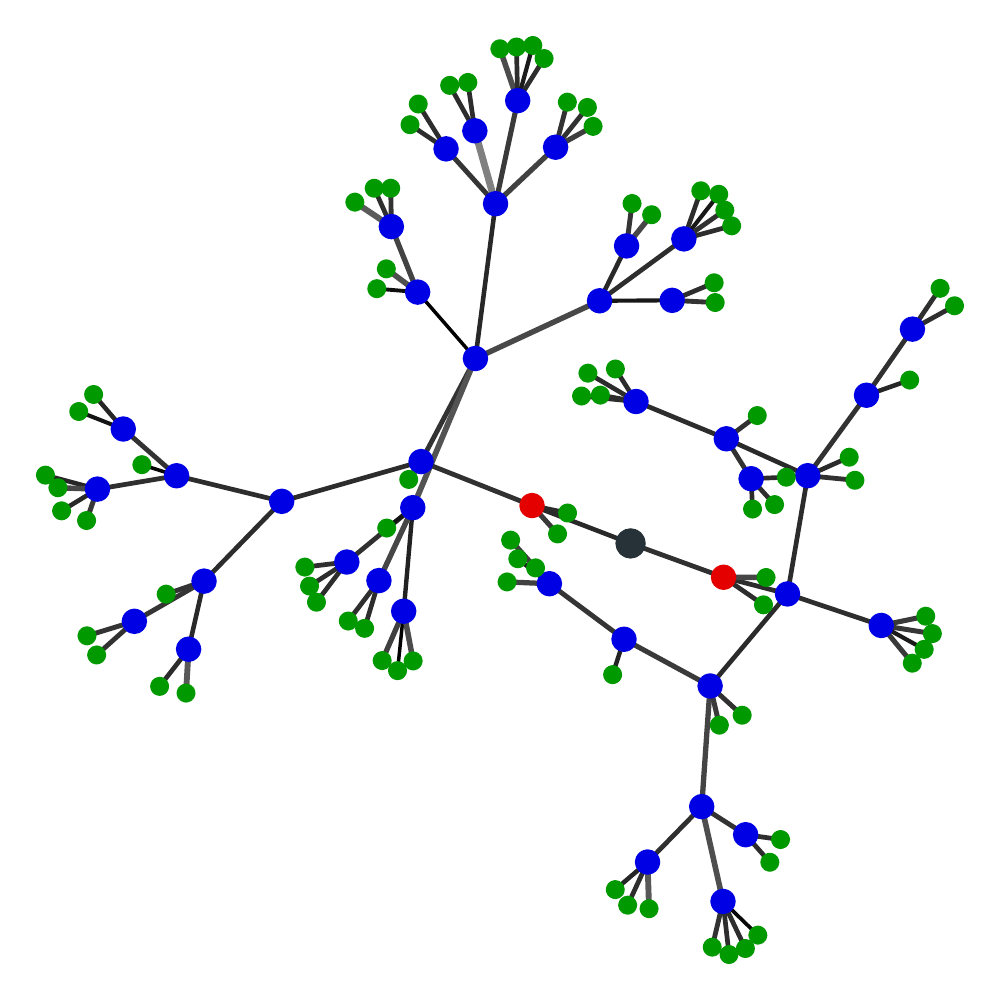}\label{sfig:img_pim}}
        \subfigure[{\scriptsize wiscosin (100\%, 85)}]{
            \includegraphics[width=0.32\linewidth]{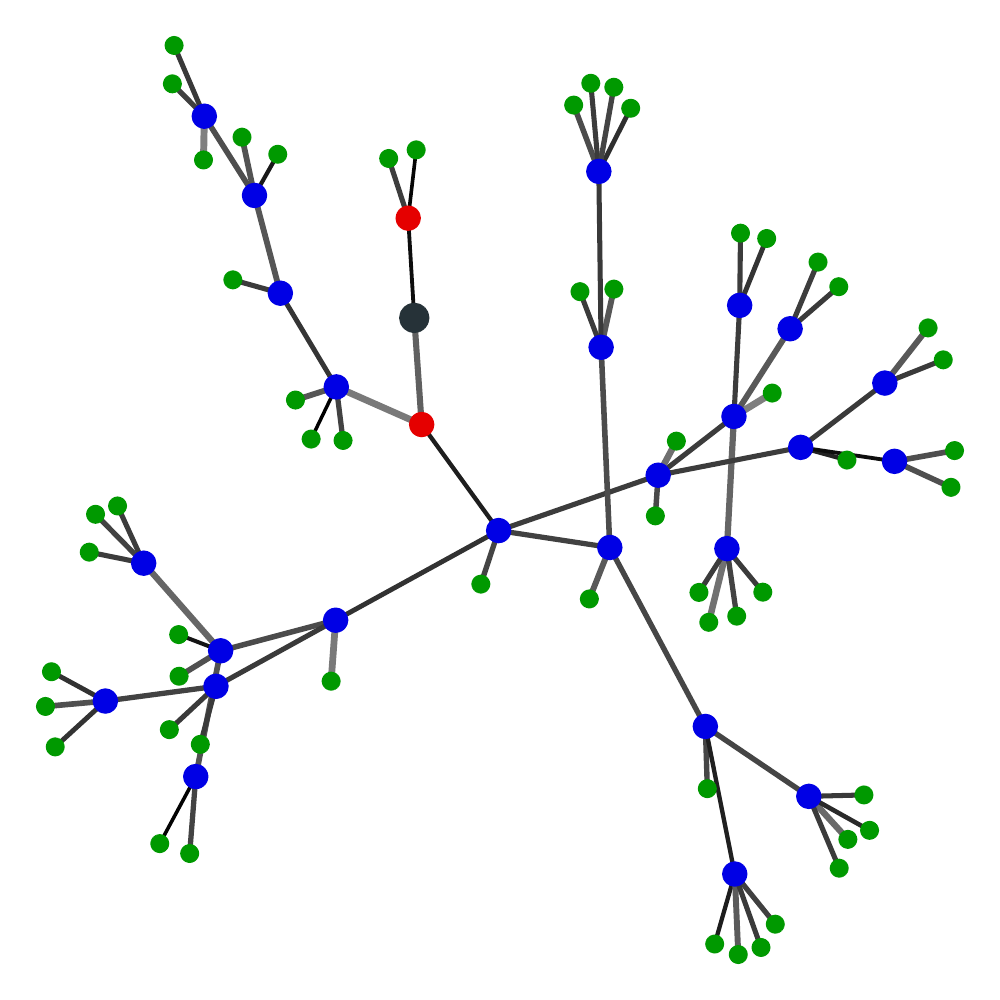}\label{sfig:img_wis}}
        \subfigure[{\scriptsize iris (100\%, 86)}]{
            \includegraphics[width=0.32\linewidth]{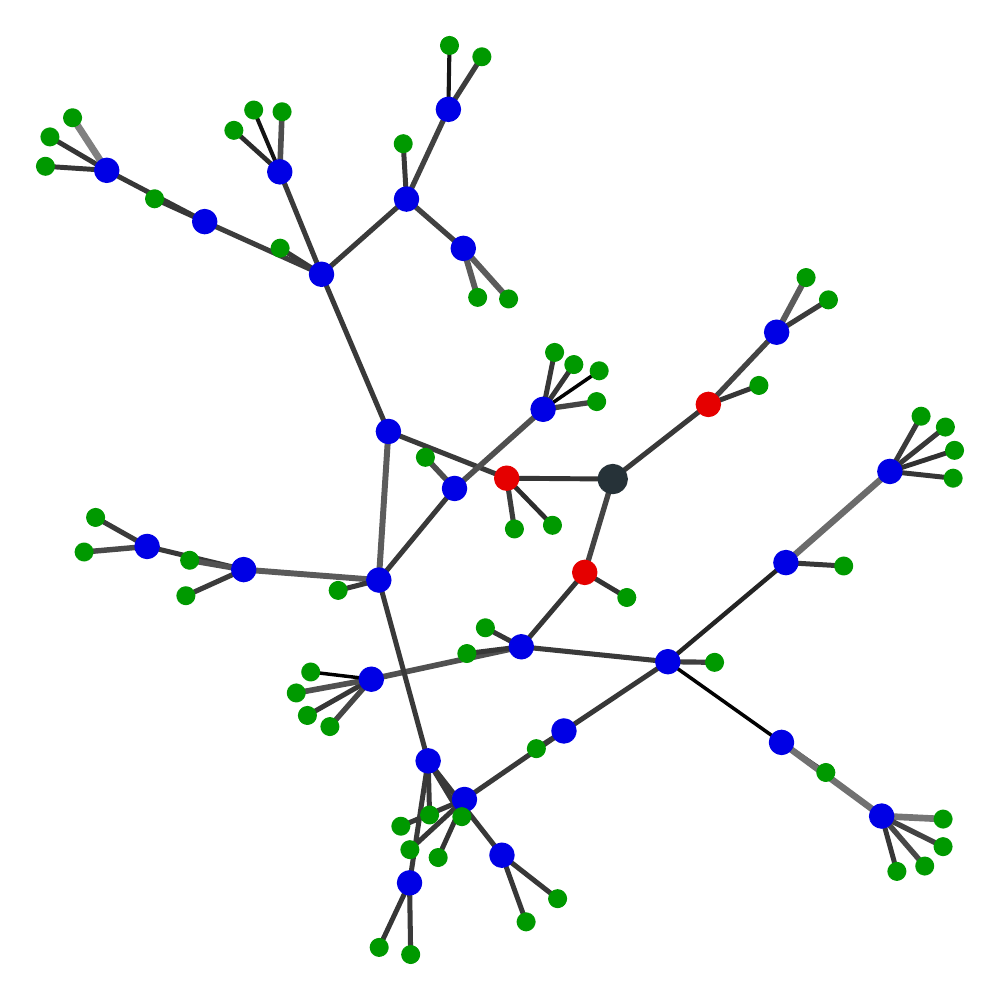}\label{sfig:img_irs}}
        
        \subfigure[{\scriptsize wine (100\%, 28)}]{
            \includegraphics[width=0.32\linewidth]{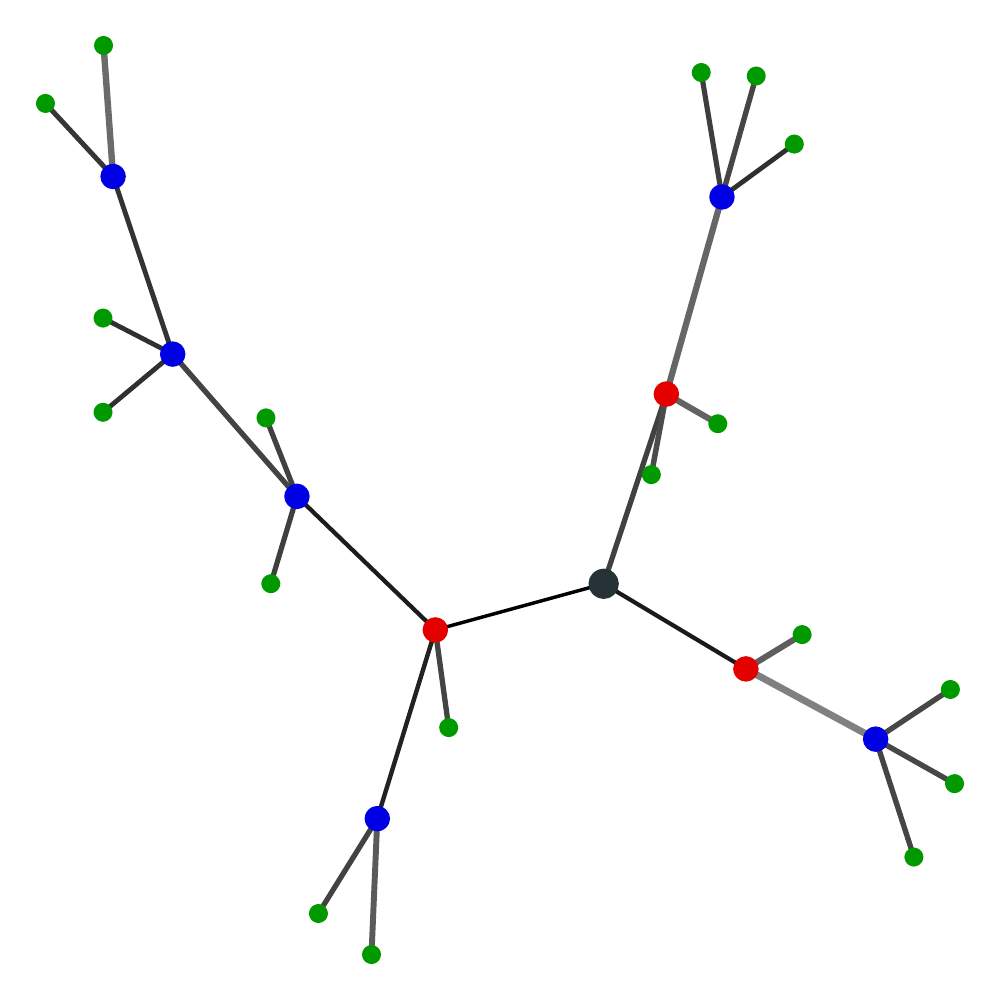}\label{sfig:img_win}}
        \subfigure[{\scriptsize vehicle (82\%, 155)}]{
            \includegraphics[width=0.32\linewidth]{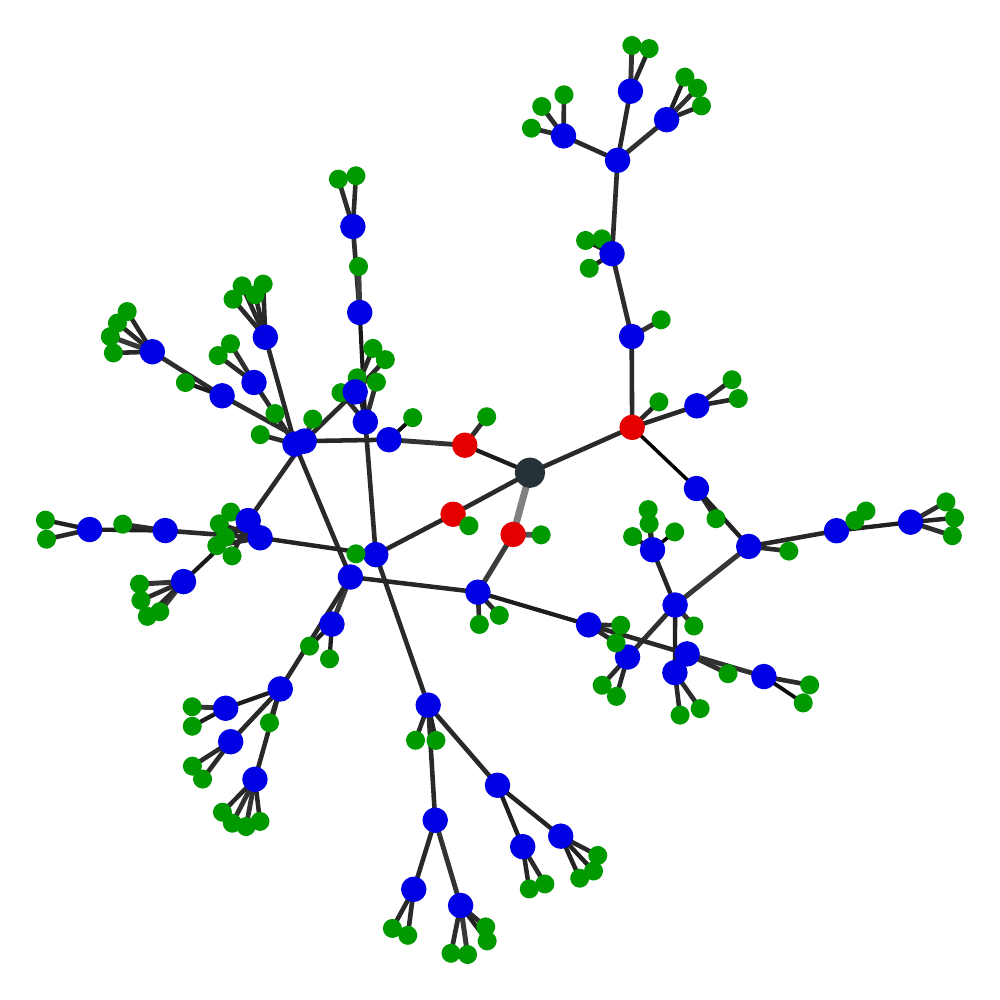}\label{sfig:img_vhl}}
        \subfigure[{\scriptsize glass (88\%, 466)}]{
            \includegraphics[width=0.32\linewidth]{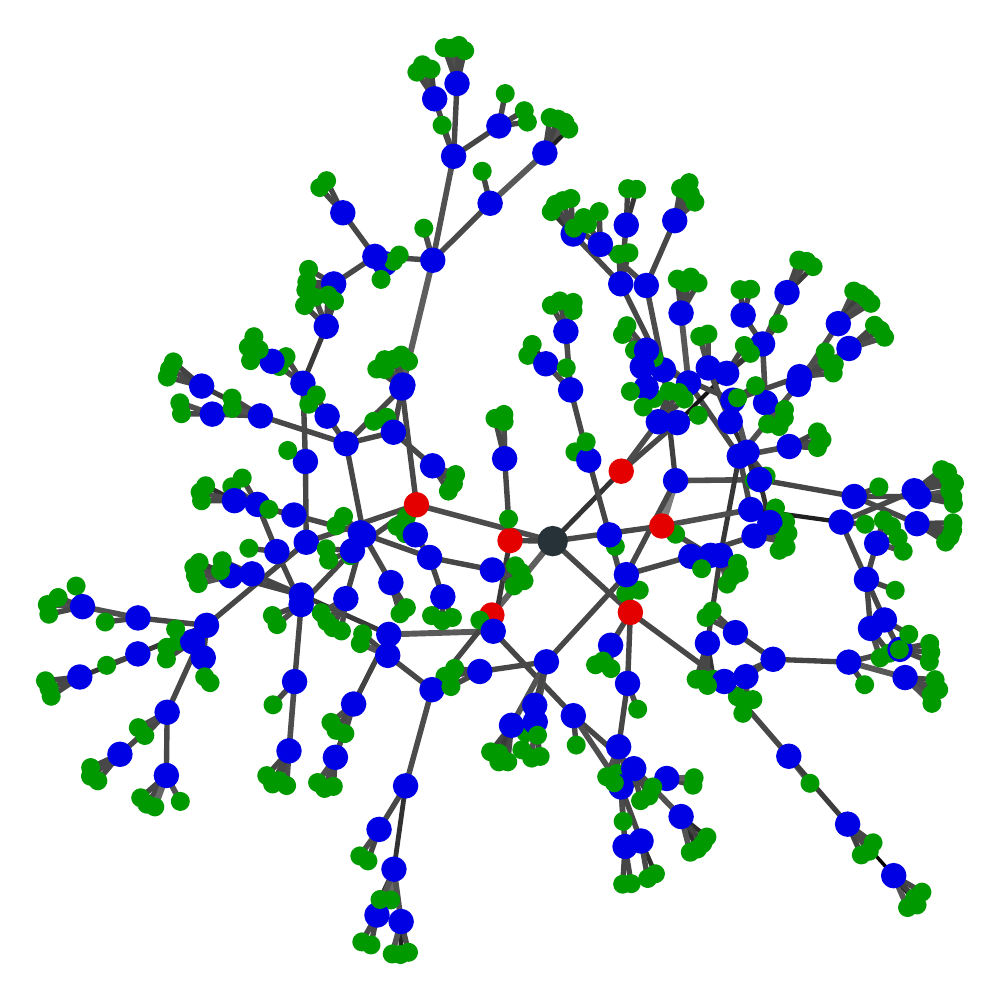}\label{sfig:img_gls}}
        \caption{Classification Trees. (a) - (i) show test accuracy and tree size of select high performing trees of datasets. The black node in a tree is its root node, class output nodes are in red, function nodes are in blue,  and leaf nodes are in green. The link connecting nodes are neural weights.
            \label{fig:tree_image_class}}
    \end{figure}

    \textbf{BNeuralT classification models summary.}  
    Table~\ref{tab:BNeuralT_all_results} suggests that BNeuralT performance on both classification and regression problems is highly competitive with MLP and other algorithms. For example, the average performance of BNeuralT's RMSprop on all classification problems is 2.65\% (average accuracy: 89.1\%) higher than the nearest best performing non-BNeuralT algorithm. The best MLP model offered an average accuracy of 86.8\%, and MLP with a $0.4$ dropout rate using Adam produced an $85.9\%$ accuracy. Other algorithms were as follows: HFNT$^{\text{S}}$, $78.9\%$; HFNT$^{\text{M}}$, $72.4\%$; MONT$_3$, $83.1\%$; DT, $81.3\%$; RF, $86.4\%$; GP, $86.8\%$; NBC, $78.2\%$; and SVM, $84.1\%$. For this performance, BNeuralT uses \textit{only} 13.25\% ($\mathbf{w} = 261$) trainable parameters than MLP's $1969$ parameters. The structures of some select best performing BNeuralT classification models are shown in Fig.~\ref{fig:tree_image_class}, where black edges indicate dendrites; and  green, blue, red, and black nodes, respectively indicate inputs, dendritic nonlinearities, root, and class nodes.
    
    The average tree size of BNeuralT with $ P[\text{leaf}_{p} < p] = 0.5$ and RMSprop was $119$ ($89\%$ accuracy). The average tree size of HFNT$^{\text{M}}$, MONT$_3$, and HFNT$^{\text{S}}$ algorithms were $29$ ($72.4\%$ accuracy), $36$ ($83.1\%$ accuracy), and $92$ ($78.9\%$ accuracy), respectively. Since tree construction and forward pass computation are similar for BNeuralT, HFNT, and MONT algorithms, there is a trade-off between the model's compactness and accuracy. In fact, this produces a set of trade-off solutions (between accuracy and complexity). Along with this set of Pareto solutions, one can choose which is the best candidate solution for the given machine learning problem under examination: more accurate but less sustainable or a little less accurate but more robust and sustainable.
    
    The forward pass computation time on a single example (in multiple of $10^{-6}$ seconds) of BNeuralT was $11.2$ seconds, whereas MLP took $1288$; DT, $3.1$; RF, $485.4$; GP, $455.6$; NBC, $31.9$; and SVM, $16.1$ seconds. DT was the fastest, and BNeuralT was the second-fastest. However, DT has a much lower accuracy ($81.3\%$) than BNeuralT ($89.1\%$). The time computation is difficult to compare as the algorithms were implemented in different programming languages~\citep{pereira2017energy}. BNeuralT was implemented in Java 11, and all other algorithms were implemented in Python $3.5.$ However, BNeuralT's performance on classification problems was clearly better among all algorithms. This is further evident from BNeuralT's collective average accuracy of all optimizers on all classification datasets was $86.1\%.$ Whereas on all optimizers, MLP's average accuracy was $83.8\%,$ tree algorithms had $80.4\%,$ and other algorithms had $83.6\%$ accuracy.
    
    We selected BNeuralT's best optimizer RMSprop for the statistical significance test. This test was designed to examine whether the performance of BNeuralT's RMSprop is statistically significant than that of the other algorithms.  Table~\ref{tab:BNeuralT_RMSprop_VS_All_KS_Test} presents Kolmogorov–Smirnov (KS) test results on two samples to examine the \textit{null hypothesis} that there is no difference between the performance distributions of BNeuralT's RMSprop and other algorithms. The results show that for most datasets and most algorithms, the classification results of BNeuralT's RMSprop show statistical significance over other algorithms' performance as the \textit{null hypothesis} of no difference is rejected in most cases. This was the case, despite using a restrictive Bonferroni correction to adjust p-values. Wilcoxon signed-rank test and Independent T-test in Supplementary Table~\ref{tab:BNeuralT_RMSprop_VS_All_W_Rank_Test} and Table~\ref{tab:BNeuralT_RMSprop_VS_All_T_Test} also favor BNeuralT's RMSprop.
    
\begin{figure}
        \centering
        \subfigure[{\scriptsize baseball (.85, 48)}]{
            \includegraphics[width=0.18\linewidth]{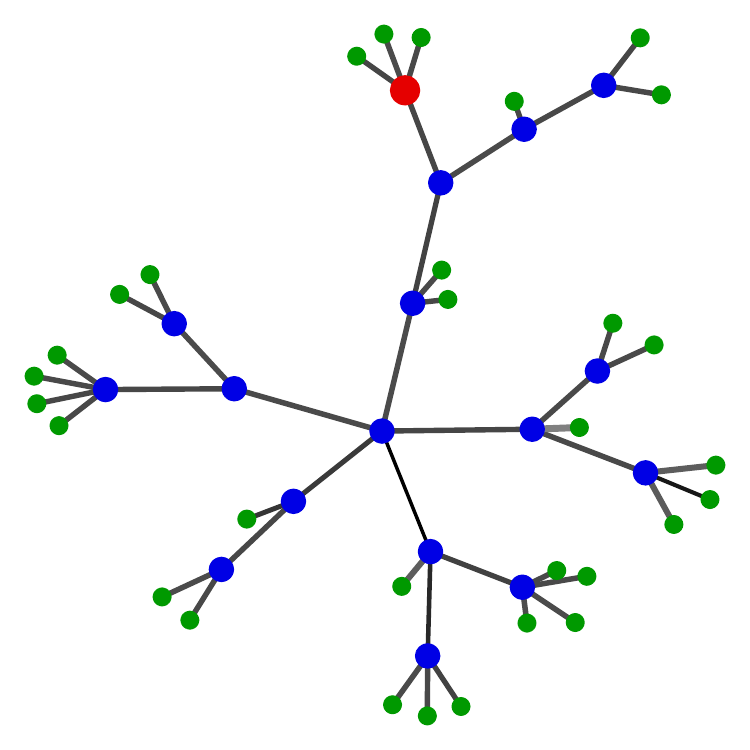}\label{sfig:img_bas}}
        \subfigure[{\scriptsize dee (.89, 89)}]{
            \includegraphics[width=0.18\linewidth]{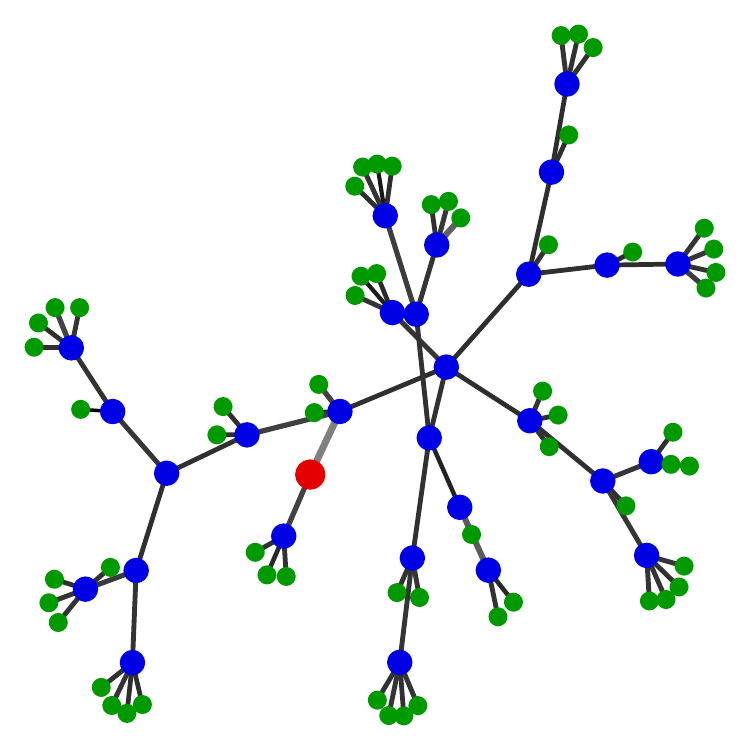}\label{sfig:img_dee}}
        \subfigure[{\scriptsize diabetese (.63, 67)}]{
            \includegraphics[width=0.18\linewidth]{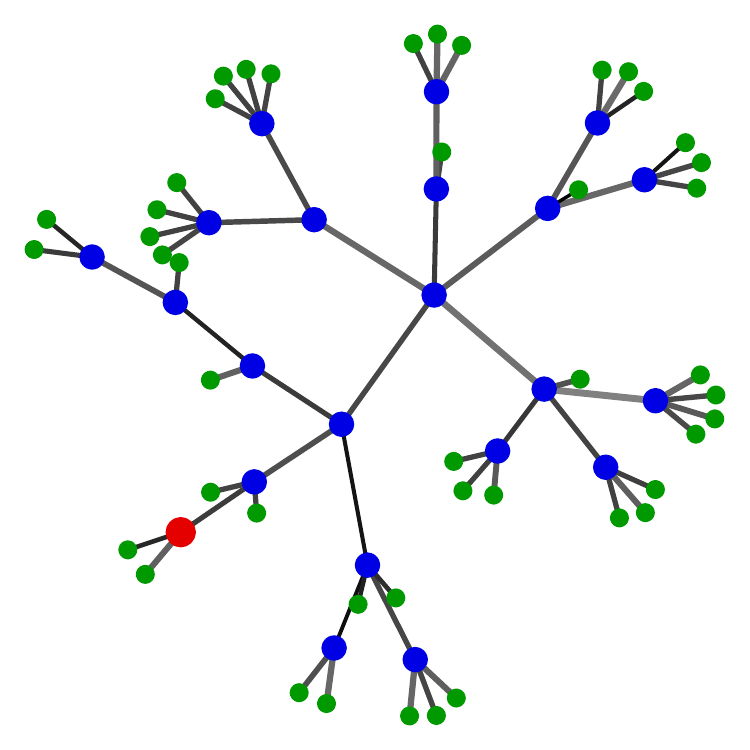}\label{sfig:img_dia}}
        \subfigure[{\scriptsize friedman (.95, 116)}]{
            \includegraphics[width=0.18\linewidth]{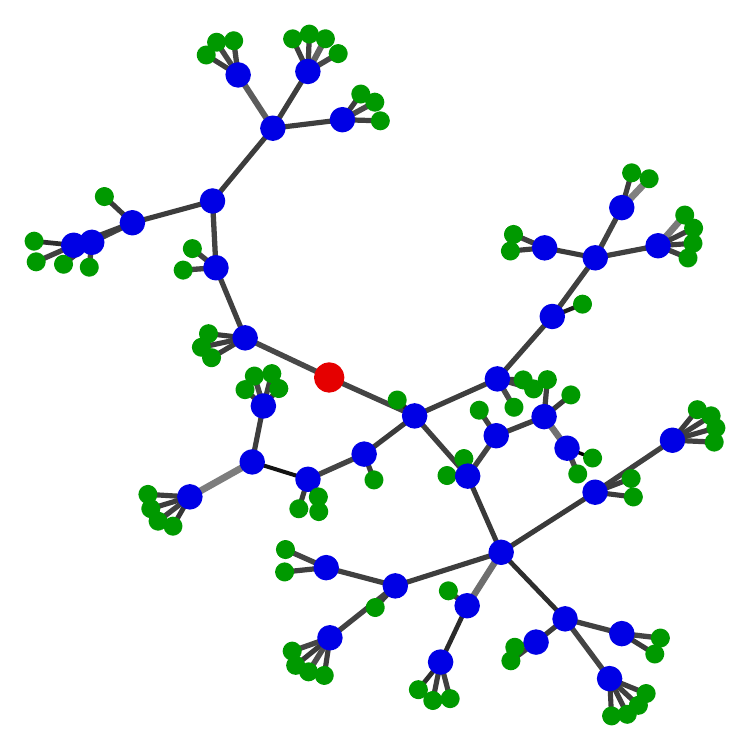}\label{sfig:img_frd}}
        \subfigure[{\scriptsize mpg6 (.9, 82)}]{
            \includegraphics[width=0.18\linewidth]{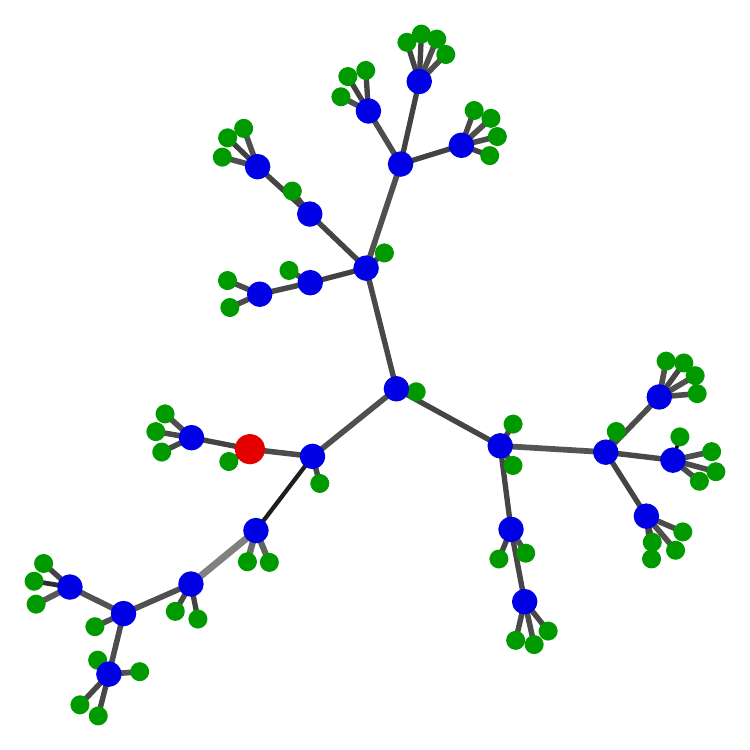}\label{sfig:img_mpg}}
        \caption{Regression Trees. (a) - (e) show best performing tree structure of the respective dataset, their accuracy and tree size shown in brackets. The red node in a tree is the root node (output node), function nodes are blue,  and leaf nodes are green. The link connecting nodes are neural weights.
            \label{fig:tree_image_reg}}
    \end{figure}

    \textbf{BNeuralT regression models summary.}  
    MLP's Adam performed best for regression problems. MLP's Adam produced an average regression fit of $0.772$ without dropout and $0.754$ with dropout on all datasets. BNeuralT's RMSprop offered an average regression fit of $0.727$, which differs only by $5.8\%$ with the best MLP result. This performance of  BNeuralT comes with the use of only $14.6\%$ trainable parameters than the parameters used by the MLP ($\textbf{w} = 1014$). (Note that an MLP dropout model during its test phase uses all weights since dropout only regularizes weights by averaging gradient over epochs during the training phase~\citep{srivastava14aJMLR}.) This suggests that BNeuralT is highly capable of learning data with very low complexity with a faster forward pass computation time. The structure of some select best performing BNeuralT regression models is shown in Fig.~\ref{fig:tree_image_reg}.
    %%%
    %%%%%In Fig.~\ref{fig:tree_image_reg}. black edges indicate dendrites, blue circles indicate nonlinearities, green nodes indicate inputs, and red nodes indicate output neurons.}
    
    The average tree size of BNeuralT with $ P[\text{leaf}_{p} < p] = 0.5$ and RMSprop for regression problems was $64$ ($\mathcal{L}_{r^2} = 0.675$). The average tree size of HFNT$^{\text{S}}$ and HFNT$^{\text{M}}$ algorithms were $127$ ($\mathcal{L}_{r^2} = 0.562$) and $90$ ($\mathcal{L}_{r^2} = 0.567$) respectively. Here, BNeuralT was able to perform accurately compared to genetically optimized HFNT algorithms with less complex models.
    
    The statistical tests in  Table~\ref{tab:BNeuralT_RMSprop_VS_All_KS_Test} also suggest that BNeuralT's RMSprop performance distribution on regression problems compared with MLP's Adam is statistically insignificant \textit{only} on the Friedman dataset. On all other regression datasets, BNeuralT's RMSprop performance is equally significant as other algorithms. 

    \begin{table} %[b]
        \footnotesize
        \centering
        \renewcommand{\arraystretch}{1.1}
        \setlength{\tabcolsep}{1.5pt}
        \caption{BNeuralT and other algorithms performance as per average (avg.) accuracy ($1 - \mathcal{L}_{\text{Error}}(\cdot) $) and avg. regression fit ($\mathcal{L}_{r^2}(\cdot)$) on the test sets of 30 independent runs for nine classification and five regression learning problems. Both accuracy and regression fit take $1.0$ as the best value. Trainable parameters ($ \mathbf{w}$) of BNeuralT and MLP are neural weights. The best accuracy among all algorithms is marked in bold. Both BNeuralT and MLP are trained on an online mode, whereas other algorithms take offline mode training. Average forward pass (single example prediction time) wall-clack time is $\tau \times \text{e}^{-6}$ seconds (a lower value is better), where $\tau^J$ and $\tau^P$ respectively indicate time for Java $11$ and Python $3.7$. Symbol $\mathcal{M}_{0.4}$ indicates MLP with a dropout rate of $0.4$, $(Avg.)$ indicates the average performance of an optimizer over datasets, and $[Avg.]$ indicates the average performance of optimizers on a dataset. The classifier results of models MONT$_3$ and HFNT$^{\text{M}}$ are from~\cite{ojha2020multi}. RF$^{\dagger} $ indicates random forest which is an ensemble model.
        %%%
        \label{tab:BNeuralT_all_results}}% caption ends
        %%%%
        %BNeuralT training is highly competitive with MLP despite having a very low number of parameters compared with MLPs. % The optimizers (algo.) for which best model $ \mathcal{G}^* $ and $ \text{M}^* $ obtained is indicated with a later and where all optimizer performed equally marked $ \star $.
        \begin{tabular}[t]{llrrrrrrrrrrrrrrrrr}
            \toprule
            & & \multicolumn{9}{c}{Avg. classification accuracy $[1 - \mathcal{L}_{\text{Error}}(\cdot)]$} & & & & \multicolumn{5}{c}{Avg. regression fit $\mathcal{L}_{r^2}(\cdot)$} \\
            \cmidrule{3-12}\cmidrule{14-19}
            & Algorithm & Aus & Hrt & Ion & Pma & Wis & Irs & Win & Vhl & Gls & $(Avg.)$ & ~ & Bas & Dee & Dia & Frd & Mpg & $(Avg.)$ \\
            %\cline{3-11}\cline{13-17}
            \midrule
            \parbox[t]{3mm}{\multirow{7}{*}{\rotatebox[origin=c]{90}{BNeuralT $(\mathcal{G})$}}}
            & GD & .862 & .848 & .874 & .789 & .968 & .947 & .953 & .671 & .580 & .832 &  & .485 & .748 & .323 & .653 & .803 & .602 \\
            & MGD & .886 & .879 & .935 & .806 & .980 & .988 & .980 & .726 & .687 & .874 &  & .585 & .804 & .434 & .763 & .849 & .687 \\
            & NAG & .886 & .878 & .938 & .808 & .980 & .987 & .978 & .731 & .688 & .875 &  & .585 & .804 & .434 & .757 & .851 & .686 \\
            & Adagrad & .872 & .852 & .907 & .780 & .974 & .966 & .981 & .697 & .638 & .852 &  & .621 & .819 & .432 & .820 & .851 & .708 \\
            & RMSprop & \textbf{.895} & \textbf{.897} & \textbf{.952} & \textbf{.822} & \textbf{.986} & \textbf{.992} & \textbf{.991} & .750 & \textbf{.732} & \textbf{.891} &  & .665 & \textbf{.837} & \textbf{.492} & .776 & \textbf{.867} & .727 \\
            & Adam & .875 & .866 & .870 & .791 & .982 & .978 & .974 & .599 & .657 & .843 &  & .579 & .765 & .360 & .587 & .825 & .623 \\[3pt]
            & $[Avg.]$ & .880 & .870 & .913 & .799 & .978 & .976 & .976 & .696 & .663 & .861 &  & .587 & .796 & .412 & .726 & .841 & .672 \\[5pt]
            \cmidrule{1-2}
            \parbox[t]{3mm}{\multirow{8}{*}{\rotatebox[origin=c]{90}{MLP $(\mathcal{M})$}}}
            & GD & .870 & .831 & .880 & .763 & .979 & .968 & .973 & .806 & .614 & .854 &  & .697 & .821 & .481 & .736 & .844 & .716 \\
            & MGD & .874 & .831 & .907 & .774 & .981 & .977 & .985 & \textbf{.853} & .629 & .868 &  & .718 & .826 & .486 & .804 & .852 & .737 \\
            & NAG & .873 & .828 & .902 & .772 & .980 & .976 & .984 & .852 & .640 & .868 &  & .718 & .826 & .485 & .786 & .851 & .733 \\
            & Adagrad & .870 & .827 & .723 & .670 & .944 & .736 & .877 & .462 & .362 & .719 &  & .407 & .584 & .221 & .571 & .630 & .482 \\
            & RMSprop & .874 & .832 & .872 & .758 & .980 & .969 & .991 & .804 & .605 & .854 &  & .718 & .826 & .486 & .866 & .866 & .752 \\
            & Adam & .876 & .833 & .882 & .774 & .984 & .972 & \textbf{.991} & .826 & .635 & .863 &  & \textbf{.721} & .829 & .490 & \textbf{.943} & \textbf{.874} & \textbf{.772} \\[3pt]
            & $[Avg.]$ & .873 & .830 & .861 & .752 & .975 & .933 & .967 & .767 & .581 & .838 &  & .663 & .785 & .442 & .784 & .820 & .699 \\[3pt]
            &$\mathcal{M}_{0.4}$Adam & .871 & .815 & .912 & .774 & .971 & .960 & .973 & .806 & .645 & .859  &  &  .707  & .828  & .491  & .884  & .862  & .754 \\
            \cmidrule{1-2}
            \parbox[t]{3mm}{\multirow{6}{*}{\rotatebox[origin=c]{90}{Trees}}}
            & HFNT$^{\text{S}}$ & .824 & .825 & .871 & .754 & .973 & .912 & .918 & .481 & .544 & .789  &  & .576 & .794 & -0.062 & .728 & .775 & .562  \\
            & HFNT$^{\text{M}}$ & .826 & .77 & .822 & .716 & .935 & .811 & .824 & .409 & .399 & .724  &  & .608 & .803 & -0.13 & .715 & .838 & .567 \\
            & MONT$_3$ & .889 & .809 & .898 & .799 & .962 & .989 & .952 & .55 & .629 & .831  &  &  &  &  &  &  &  \\
            &  DT & .801 & .744 & .890 & .705 & .932 & .943 & .897 & .708 & .698 & .813 &  & .431 & .656 & -0.12 & .689 & .756 & .484 \\
            & RF$^{\dagger} $ & .868 & .809 & .930 & .752 & .962 & .959 & .981 & .745 & \textbf{.768} & .864 & & .663  & .829  & .442  & .871  & .871  & .735 \\[3pt]
            & $[Avg.]$ & .842  & .791  & .882  & .745  & .953  & .923  & .914  & .579  & .608  & .804 &   & .458  & .605  & .081  & .601  & .633  & .476 \\[5pt]
            \cmidrule{1-2}
            \parbox[t]{3mm}{\multirow{4}{*}{\rotatebox[origin=c]{90}{Others}}}
            & GP & .861 & .820 & .916 & .769 & .970 & .960 & .983 & .843 & .689 & .868 &  & .648 & .820 & .484 & .724 & .801 & .695 \\
            & NBC & .797 & .833 & .882 & .758 & .930 & .953 & .969 & .455 & .457 & .782 &  &  &  &  &  &  &  \\
            & SVM & .861 & .841 & .880 & .769 & .977 & .926 & .978 & .762 & .575 & .841 &  & .647 & .838 & .405 & .912 & .861 & .733 \\[3pt]
            &[$Avg.$] & .840  & .831  & .893  & .765  & .959  & .946  & .977  & .687  & .574  & .830 &   & .648  & .829  & .445  & .818  & .831  & .714 \\[5pt]
            \cmidrule{1-2}
            \parbox[t]{3mm}{\multirow{9}{*}{\rotatebox[origin=c]{90}{Parameter and Time}}}
            & $\textbf{w}_\mathcal{G}$ & 204 & 560 & 185 & 157 & 268 & 180 & 358 & 249 & 190 & 261 &  & 140 & 163 & 140 & 141 & 178 & 152 \\
            & $\textbf{w}_\mathcal{M} \vert \textbf{w}_{\mathcal{M}_{0.4}}$ & 1702 & 1606 & 1602 & 3602 & 803 & 1102 & 2304 & 1703 & 3302 & 1969 &  & 1801 & 801 & 1201 & 701 & 701 & 1041 \\[3pt]
            %& ${n}_\mathcal{G}$ & & & & & & & & & & & & & & & & &  \\
            %& ${n}_{\text{HFNT}^S}$ & & & & & & & & & & & & & & & & &  \\
            %& ${n}_{\text{HFNT}^M}$ & & & & & & & & & & & & & & & & &  \\
            %& ${n}_\text{MONT}$ & & & & & & & & & & & & & & & & &  \\
            & $\tau^{J}_{\mathcal{G}}$  & 8.5 & 8.5 & 6.7 & 9.6 & 7.2 & 15.4 & 9.3 & 11.3 & 24.2 & 11.2 &  & 5.1 & 5.4 & 5.2 & 4.7 & 6.5 & 5.4 \\
           	& $\tau^{P}_{\mathcal{M}}$ & 1173 & 802 & 628 & 860 & 410 & 1452 & 1206 & 412 & 4648 & 1288 &  & 1931 & 2031 & 1409 & 163 & 1074 & 1322\\
           	%& \color{insert} $\tau^{J}_{\text{HFNT}^S}$ & & & & & & & & & & & & & & & & & \\
           	%& \color{insert} $\tau^{J}_{\text{HFNT}^M}$ & & & & & & & & & & & & & & & & & \\
           	& $\tau^{P}_{\text{DT}}$ & 1.0 & 4.2 & 1.9 & 1.1 & 1.5 & 6.4 & 5.6 & 1.2 & 4.5 & 3.1 &  & 2.0 & 2.6 & 1.8 & 0.6 & 2.0 & 1.8 \\
           	& $\tau^{P}_{\text{RF}}$ & 278.8 & 547.8 & 425.6 & 253.5 & 277.5 & 890.6 & 743.5 & 257.1 & 694.1 & 485.4  & & 195.5 & 178.0 & 155.7 & 91.7 & 175.0 & 159.2 \\
            & $\tau^{P}_{\text{GP}}$ & 187.2 & 58.6 & 108.7 & 195.8 & 197.6 & 226.1 & 253.7 & 2180 & 692.1 & 455.6 &  & 12.3 & 13.2 & 21.7 & 73.8 & 18.8 & 28.0\\
            & $\tau^{P}_{\text{NBC}}$  & 8.7 & 42.6 & 21.1 & 6.7 & 9.9 & 38.9 & 75.9 & 11.2 & 72.1 & 31.9 &  &  &  &  &  &  & \\
            & $\tau^{P}_{\text{SVM}}$ & 5.8 & 20.4 & 15.5 & 4.5 & 5.0 & 36.7 & 25.9 & 5.3 & 25.6 & 16.1 &  & 17.6 & 15.5 & 30.7 & 9.2 & 12.2 & 17.1\\
            \bottomrule
            %\multicolumn{18}{p{15cm}}{\textit{Note:} } %$^{\ddagger} $MLP architecture remain fixed for all 30 instances. Hence avg. $ \mathbf{w} $ and best $ \mathbf{w} $ remain same.}	%parallel loop and Python Keras library
        \end{tabular}
    \end{table}

    \begin{table} %[b]
        \small
        \centering
        \renewcommand{\arraystretch}{1.05}
        \setlength{\tabcolsep}{3pt}
        \caption{Kolmogorov–Smirnov (KS) test on two samples: BNeuralT's RMSprop against all other algorithms for each data. The stat, pval, and post respectively indicate KS statistic, two-tailed $p$-value, and Bonferroni correction post-hoc adjusted $p$-value. The values are marked in bold where the \textit{null hypothesis} that BNeuralT's RMSprop and other algorithms come from the same distribution is rejected as per Bonferroni correction.}
        \label{tab:BNeuralT_RMSprop_VS_All_KS_Test}
        \begin{tabular}[t]{lrrrrrrrrrrrrrrrrr}
            \toprule
            \multicolumn{3}{c}{BNeuralT's} & \multicolumn{9}{c}{Classification} & ~~ & \multicolumn{5}{c}{Regression} \\
            \cline{4-12}\cline{14-18}
            \multicolumn{3}{c}{RMSprop vs.}  & Aus & Hrt & Ion & Pma & Wis & Irs & Win & Vhl & Gls & ~ & Bas & Dee & Dia & Frd & Mpg \\
            \midrule
            \parbox[t]{3mm}{\multirow{21}{*}{\rotatebox[origin=c]{90}{MLP}}}
            & GD & stat & .43 & .63 & .73 & .70 & .47 & .67 & .60 & .63 & .70 &  & .27 & .30 & .23 & .73 & .43 \\
            & & pval & .01 & 0 & 0 & 0 & 0 & 0 & 0 & 0 & 0 &  & .24 & .14 & .39 & 0 & .01 \\
            & & post & .07 & \textbf{0} & \textbf{0} & \textbf{0} & \textbf{.03} & \textbf{0} & \textbf{0} & \textbf{0} & \textbf{0} &  & 1 & 1 & 1 & \textbf{0} & .06 \\
            & MGD & stat & .40 & .63 & .53 & .60 & .43 & .47 & .30 & .87 & .60 &  & .30 & .20 & .20 & .20 & .37 \\
            & & pval & .02 & 0 & 0 & 0 & .01 & 0 & .14 & 0 & 0 &  & .14 & .59 & .59 & .59 & .03 \\
            & & post & .16 & \textbf{0} & \textbf{0} & \textbf{0} & .07 & .03 & 1 & \textbf{0} & \textbf{0} &  & 1 & 1 & 1 & 1 & .31 \\
            & NAG & stat & .43 & .63 & .63 & .57 & .43 & .47 & .33 & .87 & .53 &  & .30 & .20 & .17 & .33 & .33 \\
            & & pval & .01 & 0 & 0 & 0 & .01 & 0 & .07 & 0 & 0 &  & .14 & .59 & .81 & .07 & .07 \\
            & & post & .07 & \textbf{0} & \textbf{0} & \textbf{0} & .07 & \textbf{.03} & .71 & \textbf{0} & \textbf{0} &  & 1 & 1 & 1 & .64 & .64 \\
            & Adagrad & stat & .43 & .67 & 1 & 1 & .87 & 1 & .83 & 1 & 1 &  & .83 & 1 & .93 & .93 & 1 \\
            & & pval & .01 & 0 & 0 & 0 & 0 & 0 & 0 & 0 & 0 &  & 0 & 0 & 0 & 0 & 0 \\
            & & post & .07 & \textbf{0} & \textbf{0} & \textbf{0} & \textbf{0} & \textbf{0} & \textbf{0} & \textbf{0} & \textbf{0} &  & \textbf{0} & \textbf{0} & \textbf{0} & \textbf{0} & \textbf{0} \\
            & RMSprop & stat & .33 & .60 & .77 & .67 & .40 & .50 & .23 & .60 & .73 &  & .30 & .23 & .20 & .57 & .20 \\
            & & pval & .07 & 0 & 0 & 0 & .02 & 0 & .39 & 0 & 0 &  & .14 & .39 & .59 & 0 & .59 \\
            & & post & .71 & \textbf{0} & \textbf{0} & \textbf{0} & .16 & \textbf{.01} & 1 & \textbf{0} & \textbf{0} &  & 1 & 1 & 1 & \textbf{0} & 1 \\
            & Adam & stat & .30 & .63 & .67 & .63 & .47 & .50 & .17 & .70 & .63 &  & .30 & .17 & .17 & 1 & .20 \\
            & & pval & .14 & 0 & 0 & 0 & 0 & 0 & .81 & 0 & 0 &  & .14 & .81 & .81 & 0 & .59 \\
            & & post & 1 & \textbf{0} & \textbf{0} & \textbf{0} & \textbf{.03} & \textbf{.01} & 1 & \textbf{0} & \textbf{0} &  & 1 & 1 & 1 & \textbf{0} & 1 \\
            %%& \color{revise} %% NO use$\mathcal{M}_{0.4}$NAG & stat & .40  & .50  & .40  & .60  & .53  & .57  & .17  & .77  & .73 &   & .27  & .53  & .33  & .73  & .43\\
            %% No use& & pval & .02  & 0  & .02  & 0  & 0  & 0  & .81  & 0  & 0 &   & .24  & 0  & .07  & 0  & .01\\
            %% NO use& & post & .20  & \textbf{.01}  & .20  & \textbf{0}  & \textbf{0}  & \textbf{0}  & 1  & \textbf{0}  & \textbf{0} &   & 1  & \textbf{0}  & .85  & \textbf{0}  & .08\\
            & $\mathcal{M}_{0.4}$Adam & stat & .37 & .50 & .53 & .57 & .40 & .77 & .30 & .63 & .70 &  & .30 & .40 & .17 & .50 & .17  \\ 
             & & pval & .03 & 0 & 0 & 0 & .02 & 0 & .14 & 0 & 0 &  & .14 & .02 & .81 & 0 & .81  \\ 
             & & post & .45 & \textbf{.01} & \textbf{0} & \textbf{0} & .20 & \textbf{0} & 1 & \textbf{0} & \textbf{0} &  & 1 & .19 & 1 & \textbf{.01} & 1  \\ 
            \cmidrule{1-2}
            \parbox[t]{3mm}{\multirow{12}{*}{\rotatebox[origin=c]{90}{Trees}}}	
            & HFNT$^{\text{S}}$ & stat & .67 & .70 & .87 & .70 & .53 & .53 & .73 & .93 & .73 &  & .47 & .37 &  & .40 & .40 \\
            &  & pval & 0 & 0 & 0 & 0 & 0 & 0 & 0 & 0 & 0 &  & 0 & .03 &  & .02 & .02 \\
            &  & post & \textbf{0} & \textbf{0} & \textbf{0} & \textbf{0} & \textbf{0} & \textbf{0} & \textbf{0} & \textbf{0} & \textbf{0} &  & \textbf{0} & \textbf{0} & \textbf{0} & \textbf{0} & \textbf{0} \\
            & HFNT$^{\text{M}}$ & stat & .67 & .87 & .83 & .87 & .63 & .53 & .77 & 1 & .87 &  & .43 & .40 &  & .63 & .43 \\
            &  & pval & 0 & 0 & 0 & 0 & 0 & 0 & 0 & 0 & 0 &  & .01 & .02 &  & 0 & .01 \\
            &  & post & \textbf{0} & \textbf{0} & \textbf{0} & \textbf{0} & \textbf{0} & \textbf{0} & \textbf{0} & \textbf{0} & \textbf{0} &  & 1 & 1 & 1 & .85 & .85 \\
            & DT & stat & .90 & .87 & .70 & .93 & .87 & .67 & .90 & .53 & .33 &  & .67 & .90 & .93 & .77 & .87 \\
            & & pval & 0 & 0 & 0 & 0 & 0 & 0 & 0 & 0 & .07 &  & 0 & 0 & 0 & 0 & 0 \\
            & & post & \textbf{0} & \textbf{0} & \textbf{0} & \textbf{0} & \textbf{0} & \textbf{0} & \textbf{0} & \textbf{0} & .71 &  & \textbf{0} & \textbf{0} & \textbf{0} & \textbf{0} & \textbf{0} \\
            & RF & stat & .43 & .73 & .40 & .90 & .80 & .57 & .47 & .30 & .33 &  & .30 & .20 & .37 & .77 & .20 \\
            & & pval & .01 & 0 & .02 & 0 & 0 & 0 & 0 & .14 & .07 &  & .14 & .59 & .03 & 0 & .59 \\
            & & post & .09 & \textbf{0} & .20 & \textbf{0} & \textbf{0} & \textbf{0} & \textbf{.03} & 1 & .92 &  & 1 & 1 & .41 & \textbf{0} & 1 \\
            \cmidrule{1-2}
            \parbox[t]{3mm}{\multirow{9}{*}{\rotatebox[origin=c]{90}{Other}}}
            & GP & stat & .43 & .63 & .57 & .67 & .43 & .50 & .33 & .83 & .30 &  & .40 & .37 & .17 & .73 & .90 \\
            & & pval & .01 & 0 & 0 & 0 & .01 & 0 & .07 & 0 & .14 &  & .02 & .03 & .81 & 0 & 0 \\
            & & post & .07 & \textbf{0} & \textbf{0} & \textbf{0} & .07 & \textbf{.01} & .71 & \textbf{0} & 1 &  & .14 & .31 & 1 & \textbf{0} & \textbf{0} \\
            & NBC & stat & .93 & .57 & .77 & .77 & .93 & .57 & .50 & 1 & .97 &  &  &  &  &  &  \\
            & & pval & 0 & 0 & 0 & 0 & 0 & 0 & 0 & 0 & 0 &  &  &  &  &  &  \\
            & & post & \textbf{0} & \textbf{0} & \textbf{0} & \textbf{0} & \textbf{0} & \textbf{0} & \textbf{.01} & \textbf{0} & \textbf{0} &  &  &  &  &  &  \\
            & SVM & stat & .50 & .53 & .70 & .67 & .30 & .70 & .47 & .20 & .77 &  & .37 & .13 & .50 & .90 & .20 \\
            & & pval & 0 & 0 & 0 & 0 & .14 & 0 & 0 & .59 & 0 &  & .03 & .96 & 0 & 0 & .59 \\
            & & post & \textbf{.01} & \textbf{0} & \textbf{0} & \textbf{0} & 1 & \textbf{0} & \textbf{.03} & 1 & \textbf{0} &  & .31 & 1 & \textbf{.01} & \textbf{0} & 1 \\
            \bottomrule
        \end{tabular}
    \end{table}
    
    \textbf{BNeuralT pattern recognition (MNIST) models summary.}   RMSprop optimizer was found robust and converging fastest for classification models 
    (cf. Sec.~\ref{sec:BNeuralT_covergence}).
    %(cf. Fig.~\ref{fig:BneuralT_class_covg} and~\ref{fig:BneuralT_tree_size_vs_acc}). 
    %%%
    Hence, we train BNeuralT on the MNIST character classification dataset~\citep{mnistDataSet} using RMSprop. We fed BNeuralT with pixels of MNIST character images since we do not use convolution in BNeuralT. 
    We aimed at generating varied BNeuralT models with varied trainable parameters length by varying tree size. We hoped that a low complexity (few parameters) BNeuralT model would perform competitively with a few reported state-of-the-art. Therefore, we compare BNeuralT performance to gauge its robustness not only on learning small-scale problems reported in Table~\ref{tab:BNeuralT_all_results} but on large-scale learning problems like MNIST. Table~\ref{tab:BNeuralT_MNSIT} summarizes BNeuralT models compared with the performances of tree-based state-of-the-art classification algorithms. 
    	
    Table~\ref{tab:BNeuralT_MNSIT} presents MNIST (pixels) results compared with classification trees. BNeuralT performs the best among the reported trees that work on MNIST (pixels) for character classification. However, convolution of images has been proven efficient for image classification problems. For example, CapsNet~\citep{sabour2017dynamic}, a state-of-the-art algorithm on MNIST (convolution), has an error rate of $0.25$, but it uses $8$ \textit{million parameters}. Compared to that, a BNeuralT on MNIST (pixels) used $23,835$ \textit{trainable parameters} for an error rate of $6.08$, and another model used $241,999$ \textit{trainable parameters} for an error rate of $5.19$. Obviously, there is a trade-off between the model's parameters size and accuracy. The performances of a varied range of other algorithms on MNIST dataset are available at~\citep{mnistDataSet}. Our goal is to use as much compact model as we can for high accuracy. 
    	
    In our few trials, BNeuralT does perform competitively with many state-of-the-art (cf. Table~\ref{tab:BNeuralT_MNSIT}). The performance of BNeuralT is better than
    tree-alternating optimization (TAO)~\citep{carreira2018alternating}, 
    CART~\citep{breiman1984classification},
    C5.0~\citep{quinlan2014c4}, 
    oblique classifier 1 (OC1)~\citep{murthy1993oc1}, and 
    generalized unbiased interaction detection and estimation (GUIDE)~\citep{loh2014fifty} 
    algorithms that worked on MNIST raw pixels inputs~\citep{zharmagambetov2019experimental} like BNeuralT (cf. Table~\ref{tab:BNeuralT_MNSIT}). 
    	
\begin{figure}
   		\centering
   		\includegraphics[width=0.98\linewidth]{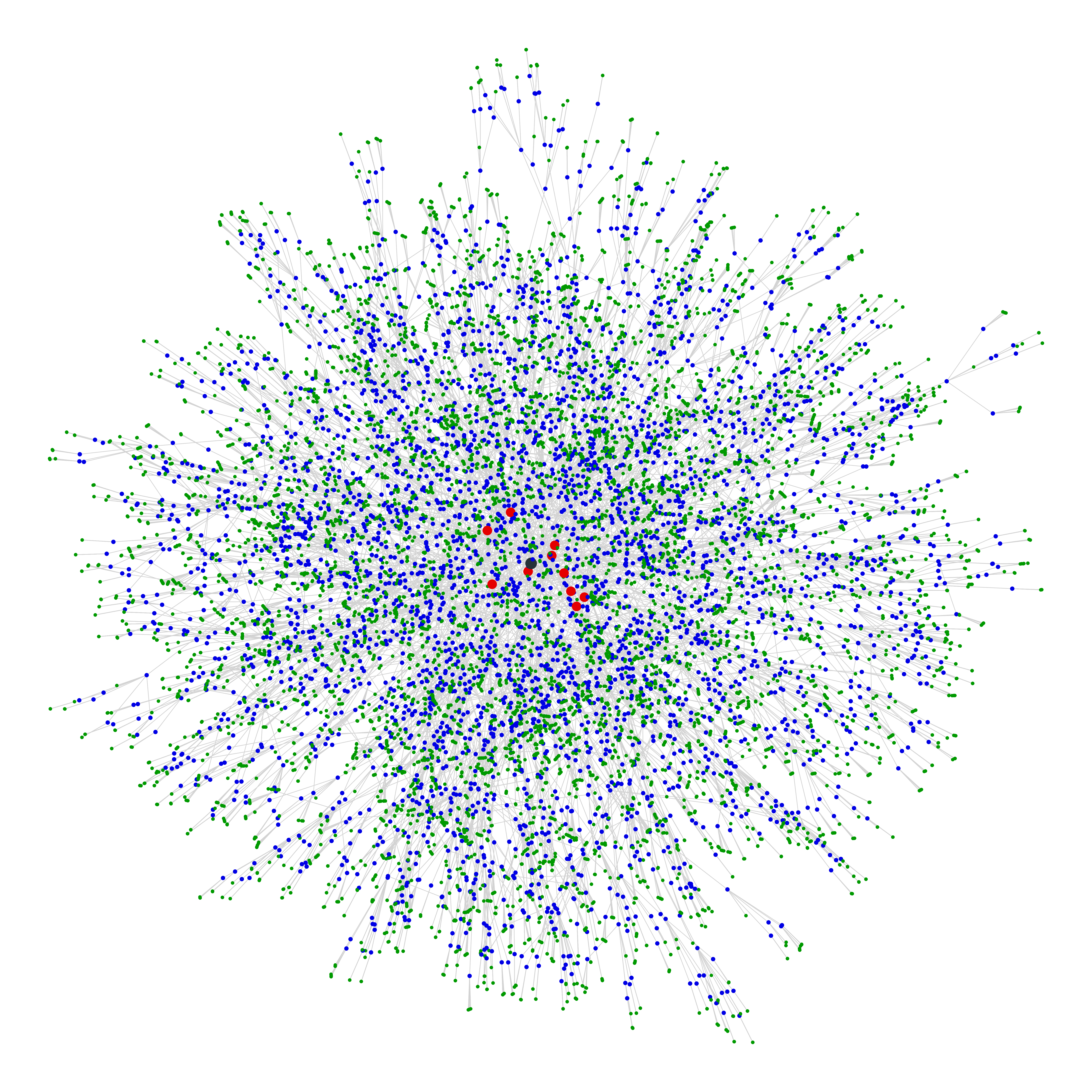}
   		\caption{BNeuralT-20K (pixels) MNIST model (tree structure). This model has 3,664 function nodes (blue nodes), 16,507 leaf nodes (green nodes), and ten class nodes (red nodes in the inner circle), and the root node (in black) in the center. This model has 6,738 edges (gray lines connecting nodes). These lines also represent neural weights. Each blue node also has its bias. Edge weights and bias together make 23,835 tree's trainable parameters. This model has a test accuracy of 94\% (an error rate of 6.08\%).
   		\label{fig:tree_img_mnist}}
   		%% 3664, 3664, 16507

   	\end{figure}

    We compare BNeuralT with biologically plausible models of \cite{jones2021might} that performed binary classification on MNIST's two classes (class 3 and class 5). This is, however, a trivial comparison as BNeuralT works on all classes and uses sigmoidal dendritic  nonlinearities, whereas \cite{jones2021might}'s models work on binary class and use Leaky ReLU as dendritic  nonlinearities. They obtained an error rate of $7.8\%$, $3.65\%$, and $8.89\%$ respectively with 1-tree ($\textbf{w}=2,047$), 32-tree ($\textbf{w}=65,504$), and A-32-tree ($\textbf{w}=65,504$) models. In contrast to \cite{jones2021might}'s models,  BNeuralT performs classification on all ten classes of MNIST pixels. Obviously, some classes are easier to learn than others (see Fig.~\ref{fig:BNeuralT_MNIST_ROC}), and training a binary classifier presents an entirely different difficulty level than a multi-class classification. However, although a one-to-one comparison is not possible in such a scenario, it may be worth noting that BNeuralT obtained an error rate of $7.74\%$ ($\textbf{w}=11,987$) and $6.08\%$ ($\textbf{w}=23,835$) on all ten classes. Therefore,  the sparse stochastic structure of BNeuralT (e.g., Fig.~\ref{fig:tree_img_mnist}) stands competitive with the models of  \cite{jones2021might}. 
    	
    %%%%These tree-based models also worked on MNIST LeNet5 convolutional inputs~\citep{zharmagambetov2019experimental}.  Clearly, except TAO (LeNet-5 convolutional input) all tree-based algorithm with pixel-based input has performed much worse than the  based BNeuralT. The model TAO (pixels, oblique split) has a performance similar to BNeuralT. The results (test error rate) of CNN models, however, better than BNeuralT. It is important to note that this performance of BNeuralT comes with 241,999 trainable parameters and only 2098 activation function computation compared to CNN-based models. 
    	
    %%%%The BNeuralT is a new and first attempt to train stochastic model architecture using a recursive backpropagation algorithm with a minimal experiment setup. With this setup, BNeuralT produces high-performing models with extremely low complexity. A BNeuralT with \textit{23,835 trainable parameters} has only 3.9\% and 4.6\% difference in its accuracy compared to models with \textit{trainable parameters in millions and uses convolutional} show its robustness in learning ability with a very few parameters. 
    %%%%%%%And, only a 1.3\% difference in performance with 2-layer neural nets. 
    	
    Moreover, BNeuralT models show a linear relation between trainable parameters and their accuracy (cf. Table~\ref{tab:BNeuralT_MNSIT}). Hence, BNeuralT models with relatively higher parameters and exhaustive hyperparameter tuning are able to produce efficient results. 
    %%BNeuralT is a new class of algorithms capable of performing with very high accuracy and low complexity. 
    %%%ROC
    Fig.~\ref{fig:BNeuralT_MNIST_ROC} shows an example BNeuralT (20K) model's training convergence and MNIST character classification performance on a receiver operating characteristic curve plot, where BNeuralT (20K) model for all classes produces a very \textit{high sensitivity} (true-positive rate) and \textit{very low specificity} (low false-positive rate). Of all classes, we may arrange classes on the scale of hardness of learnability in the order of ``easiest to hardest'' to learn as $c1, c6, c0, c7, c5, c4, c9, c2, c3,$ and $c8$ (cf.  Fig.~\ref{fig:BNeuralT_MNIST_ROC}).
    	
    \begin{minipage}[t]{0.48\textwidth} 
   		\centering
   		\begin{figure}[H]
   			\centering
   			\includegraphics[width=0.98\linewidth]{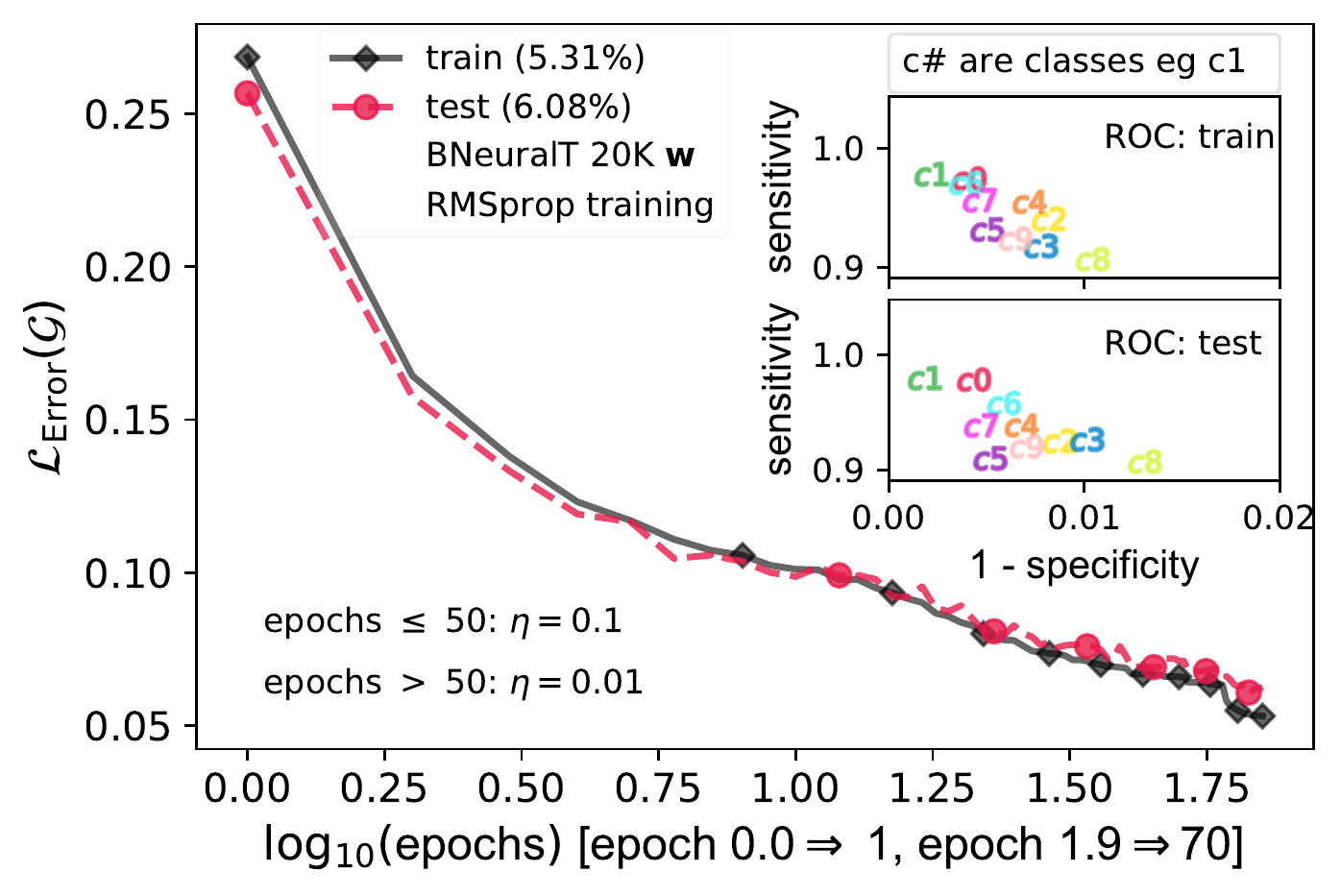}
   			\label{fig:mnist_performance}
   			\caption{BNeuralT-20K (23,835 trainable parameters $ \mathbf{w} $) model's RMSprop training and test error over 70 epochs on the MNIST dataset. Zoom-in for BNeuralT performance on receiver operating characteristic curve plots on training (\textit{inner top}) and test (\textit{inner bottom}) sets.}
   			\label{fig:BNeuralT_MNIST_ROC}
   		\end{figure}
   	\end{minipage}\quad
   	\begin{minipage}[t]{0.48\textwidth} 
   		\centering
   		\begin{table}[H]
   			\centering
   			\small
   			\renewcommand{\arraystretch}{1}
   			\setlength{\tabcolsep}{5pt}
   			\caption{Test error rate~\% of ad hoc BNeuralT  ($\mathcal{G}$) models with a varied number of trainable parameters on the MNIST dataset. All models are trained for \textit{70 epochs}, and where denoted $ \dagger $ is for \textit{25 epochs}. The decision tree models are reported in \citep{zharmagambetov2019experimental}. \label{tab:BNeuralT_MNSIT}}
   			\begin{tabular}{llr}
   				\toprule
   				\multicolumn{1}{c}{} & 
   				\multicolumn{1}{l}{Algorithms} & 
   				\multicolumn{1}{c}{$ \text{Error} $(\%)} \\ 
   				\midrule
   				\parbox[t]{2.8mm}{\multirow{4}{*}{\rotatebox[origin=c]{90}{BNeuralTs}}} 
   				& BNeuralT-10K (pixels) & 7.74 \\
   				& BNeuralT-18K (pixels) & 6.58 \\
   				& BNeuralT-20K (pixels) & 6.08 \\
   				& BNeuralT-200K$^\dagger$ (pixels)  & \textbf{5.19} \\[5pt]
   				\cline{1-2}
   				\parbox[t]{2.8mm}{\multirow{8}{*}{\rotatebox[origin=c]{90}{Classification Trees}}} 
   				& GUIDE (pixels, oblique split) & 26.21  \\
   				& OC1 (pixels, oblique split) & 25.66  \\
   				& GUIDE (pixels)    & 21.48  \\
   				& CART-R (pixels) & 11.97  \\
   				& CART-P (pixels) & 11.95  \\
   				& C5.0 (pixels)   & 11.69  \\
   				& TAO (pixels)    & 11.48  \\
   				& TAO (pixels, oblique split)  & 5.26 \\
   				\bottomrule
   				\hline 
   			\end{tabular} 
   		\end{table}
   	\end{minipage}

    \subsection{BNeuralT convergence analysis}
    \label{sec:BNeuralT_covergence}
\begin{figure}
        \centering
        \includegraphics[width=0.98\linewidth]{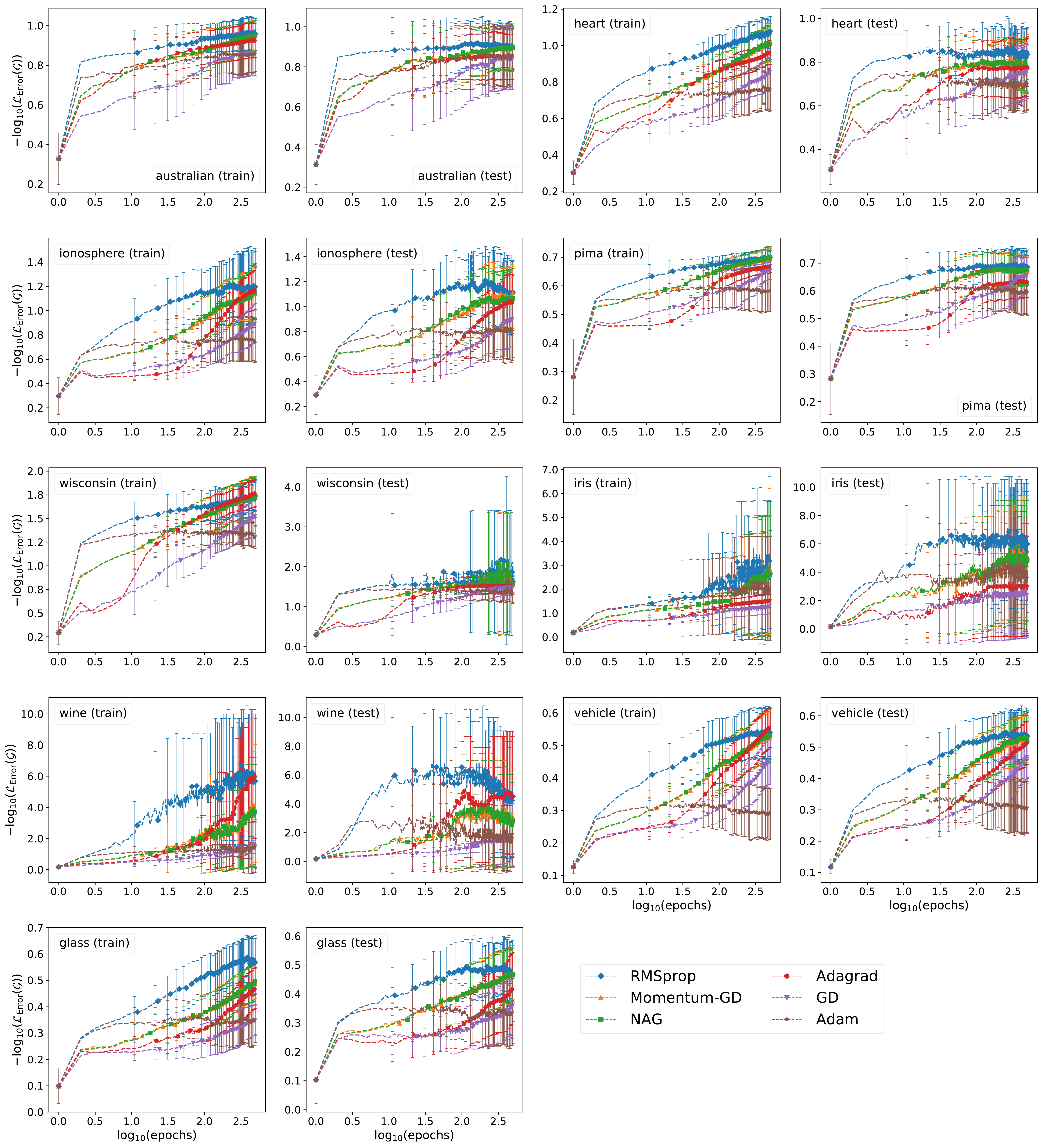}
        \caption{Early-stopping training of BNeuralT having \textit{sigmoid} nodes, $0.1$ learning rate, and $0.4$ leaf generation rate. BNeuralT average convergence trajectory performance computed over 30 independent runs for six optimizers over nine classification problems. The x-axis, $\log_{10}(\mbox{epochs})$ has the range $[0.0, 2.7]$ and is the training epochs $1$ to $500$. The y-axis, $-\log_{10}(\mathcal{L}_{\text{Error}}(\mathcal{G}))$ has the range $[0,10]$ and is the log scale of the training and test  accuracies. An error of $0.01$ (accuracy $99$\%) on the  $-\log_{10}(\mathcal{L}_{\text{Error}}(\mathcal{G}))$ scale has a value of $2.0$ and an accuracy of $90$\% has a value of $1.0$. Thus, a higher value on the y-axis is better. Error bar is the standard deviation of $-\log_{10}(\mathcal{L}_{\text{Error}}(\mathcal{G}))$ and it indicates stochasticity of the convergence that helps an optimizer escape local minima better. Thus, a larger length is better. For each data, training and test convergence pair are plotted for 500 epochs (on the log scale, $2.7$). RMSprop, MGD, NAG, Adagrad, GD, and Adam are respectively indicated in blue, orange, green, red, purple, and brown colors, respectively, with symbols diamond, triangle, circle, downward triangle, and star.}%In all Fig., RMSprop is (shown in blue line) is the fastest converging. Bottom rightmost plot BNeuralT (mode parameter 241,999) training on MNIST dataset for 15 epochs.
        \label{fig:BneuralT_class_covg}
    \end{figure}

\begin{figure}
        \centering
        %\subfigure[{\scriptsize  baseball}]
        \includegraphics[width=0.98\linewidth]{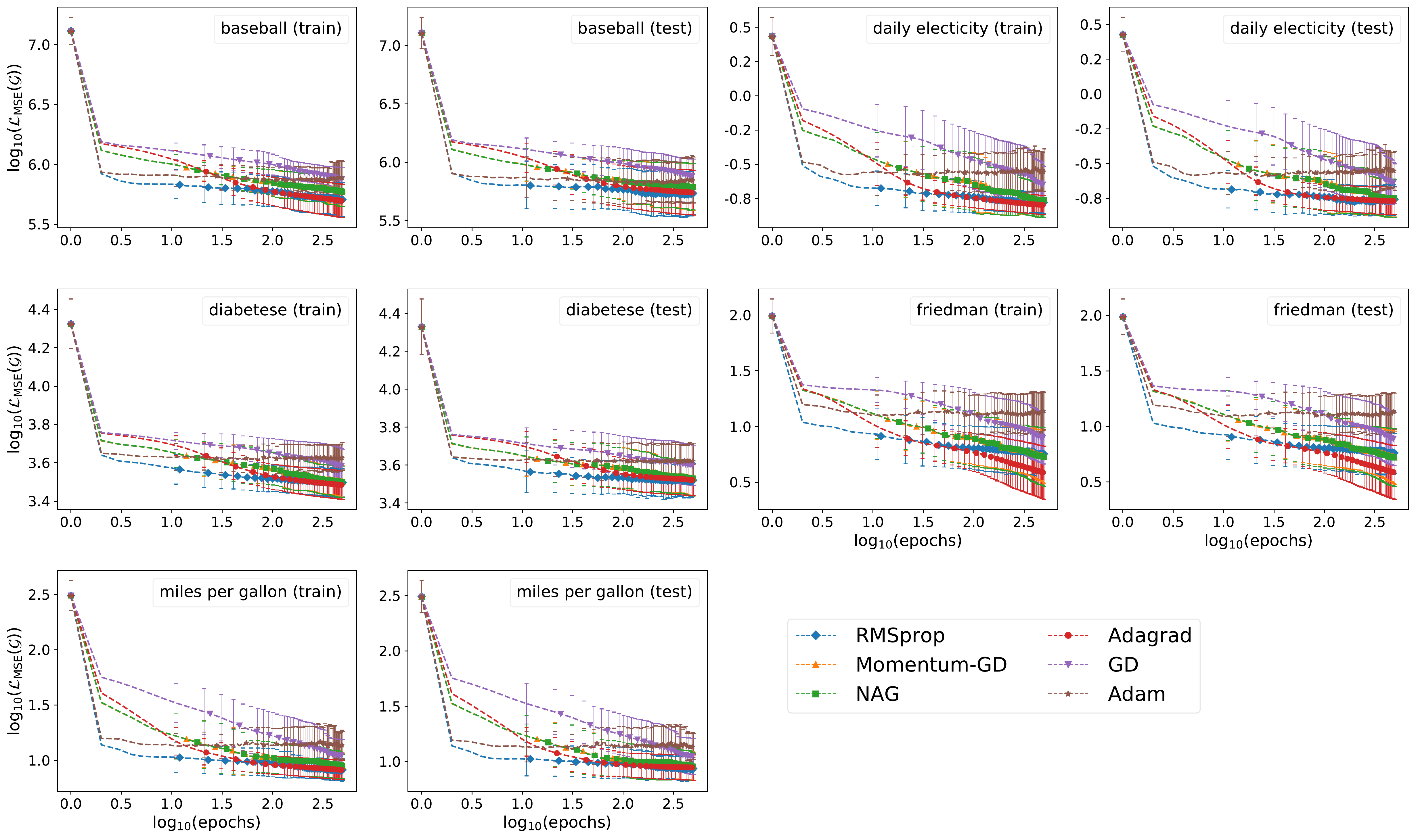}
        \caption{BNeuralT average convergence trajectory performance computed over 30 independent runs for six optimizers over five regression problems. The x-axis, $\log_{10}(\mbox{epochs})$ has the range $[0.0, 2.7]$ and is the training epochs $1$ to $500$. The y-axis, $\log_{10}(\mathcal{L}_{\mbox{MSE}}(\mathcal{G}))$ is the training and test sets mean square error (MSE) on the log scale. An MSE $0.01$ on the log scale has a value of $-2.0$. Thus, a lower value is better. Error bar is the standard deviation of $\log_{10}(\mathcal{L}_{\text{MSE}}(\mathcal{G}))$ and it indicates stochasticity of the convergence that helps an optimizer escape local minima better. Thus, a larger length is better. For each dataset, training and test convergence pair are plotted for 500 epochs (on the log scale, $2.7$). The optimizers RMSprop, MGD, NAG, Adagrad, GD, and Adam are respectively indicated in blue, orange, green, red, purple, and brown colors with respective symbols diamond, triangle, circle, down triangle, and star.}%Similar to classification dataset, RMSprop (blue line) is the best, followed by Adam, MGD, and NAG, Adagrad and GD.  Adagrad in red is stable and finds a regression fit for low complexity models.
        \label{fig:BneuralT_reg_all}
    \end{figure}

\begin{figure}
        \centering
        \subfigure[{\scriptsize  BNeuralT: Sigmod, $\eta = 0.1$}]{
            \includegraphics[width=0.32\linewidth]{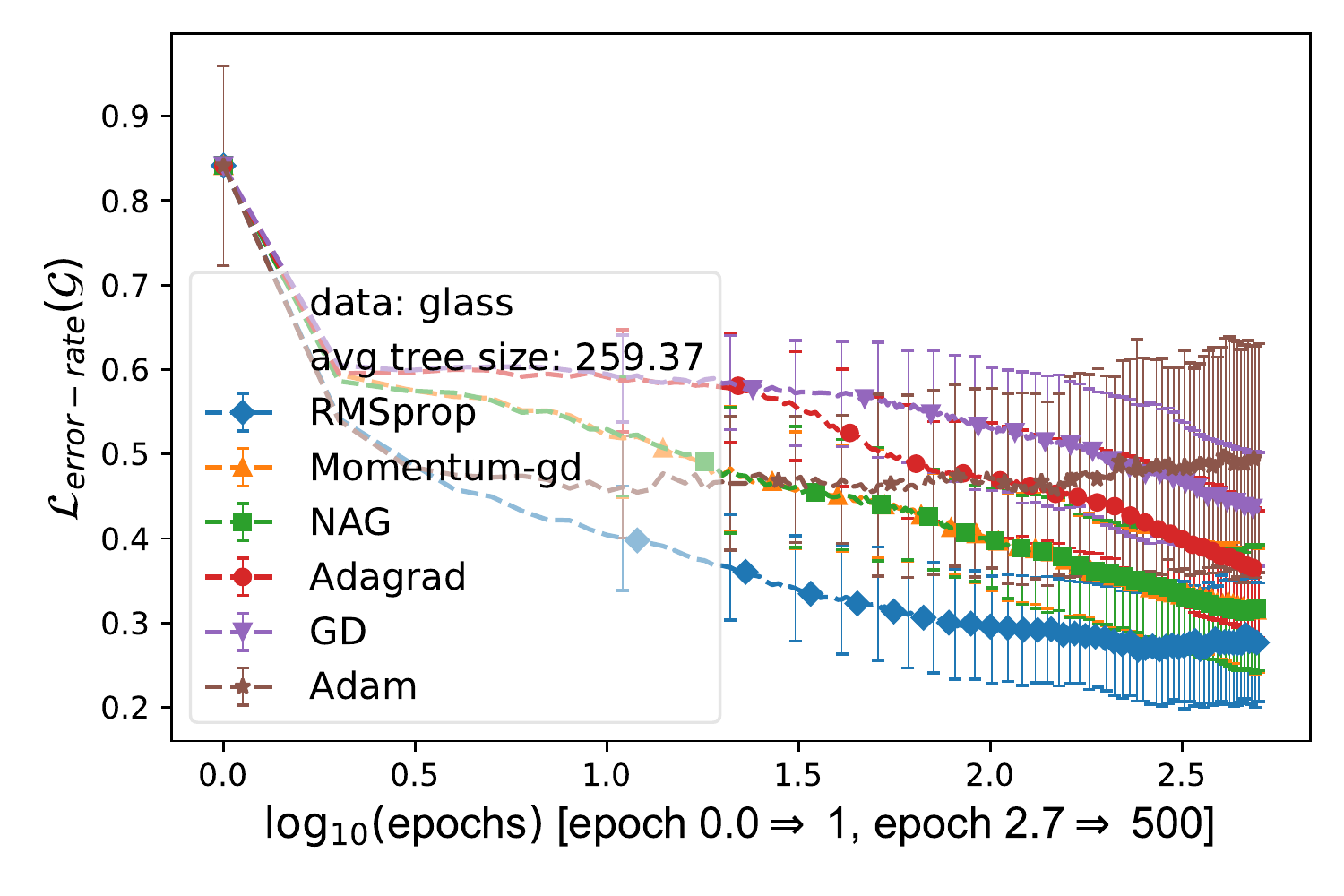}\label{sfig:bcp1}}
        \subfigure[{\scriptsize  BNeuralT: Sigmod, $\eta = default$}]{
            \includegraphics[width=0.32\linewidth]{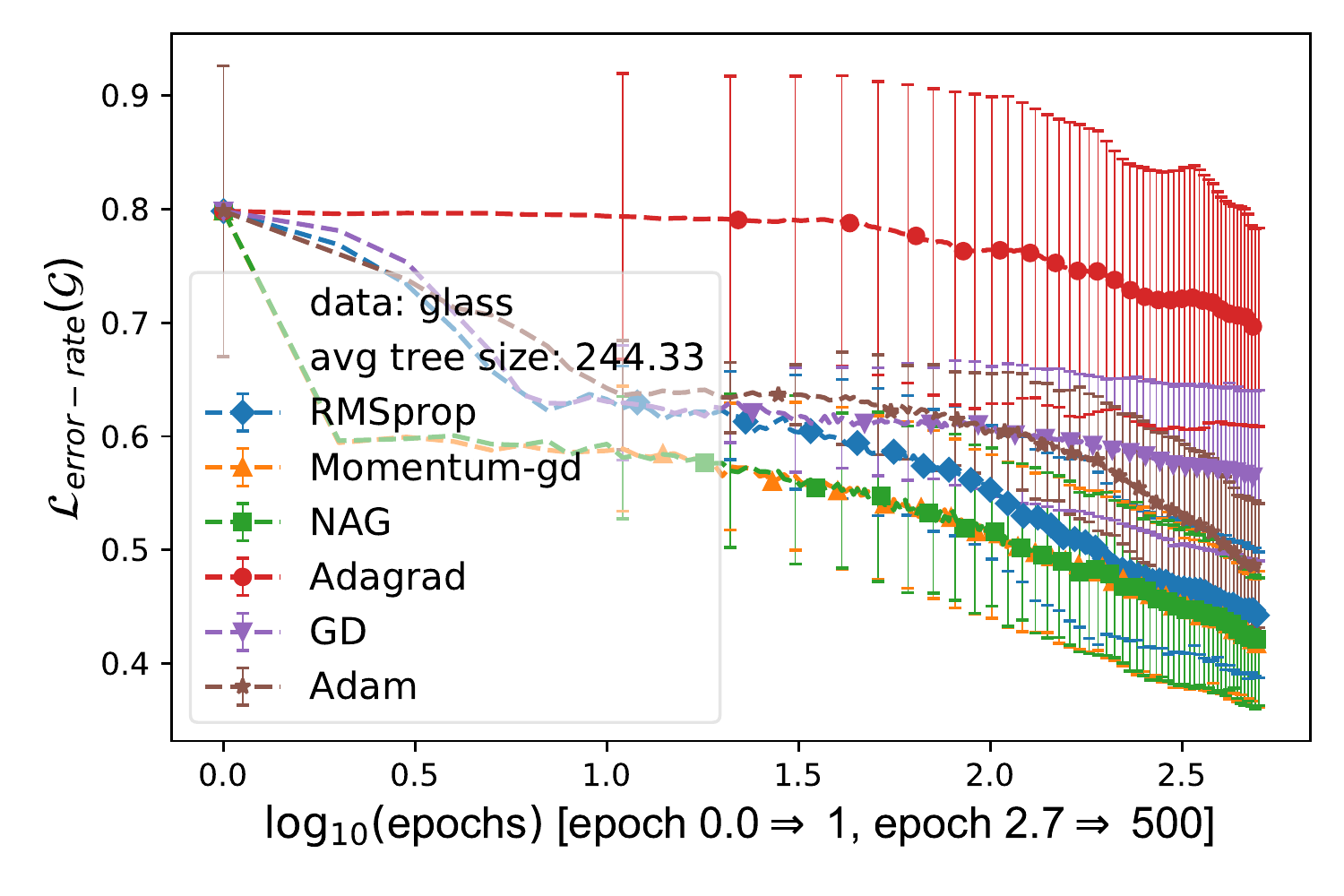}\label{sfig:bcpd}}
        \subfigure[{\scriptsize  BNeuralT: ReLU, $\eta = 0.1$}]{
            \includegraphics[width=0.32\linewidth]{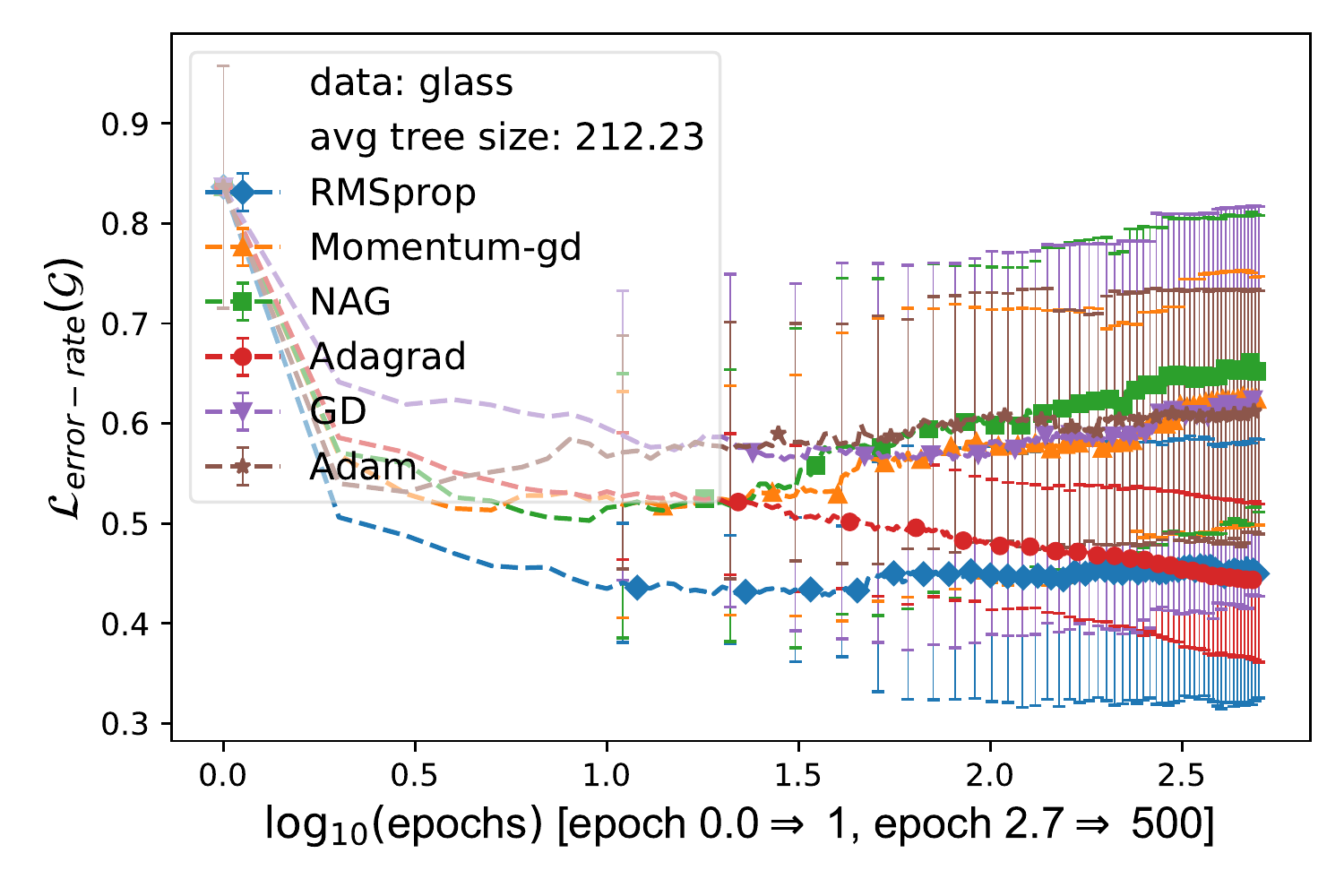}\label{sfig:bcr}}

        \subfigure[{\scriptsize  MLP: Sigmod, $\eta = 0.1$}]{
            \includegraphics[width=0.32\linewidth]{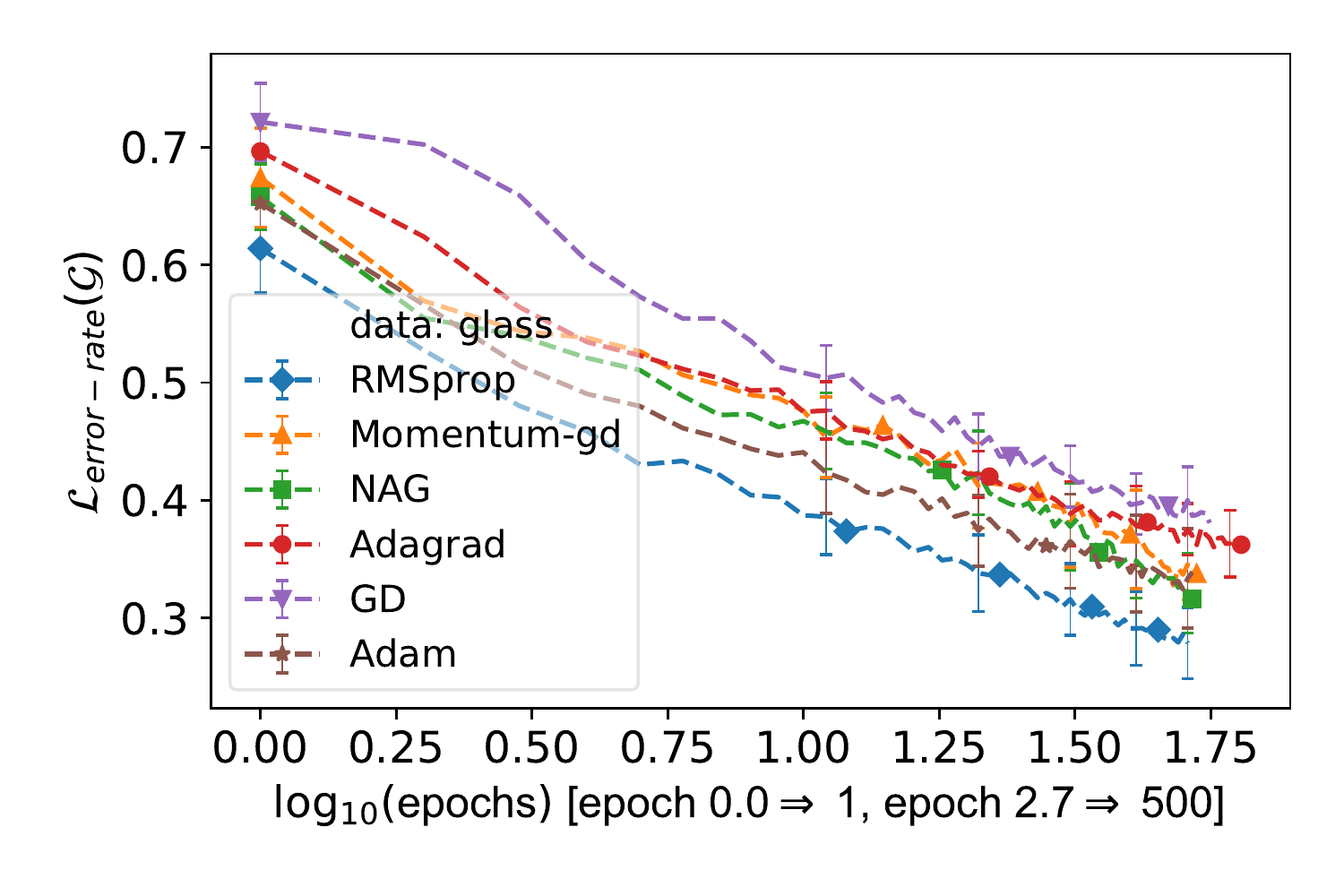}\label{sfig:mcp1}}
        \subfigure[{\scriptsize  MLP: Sigmod, $\eta = default$}]{
            \includegraphics[width=0.32\linewidth]{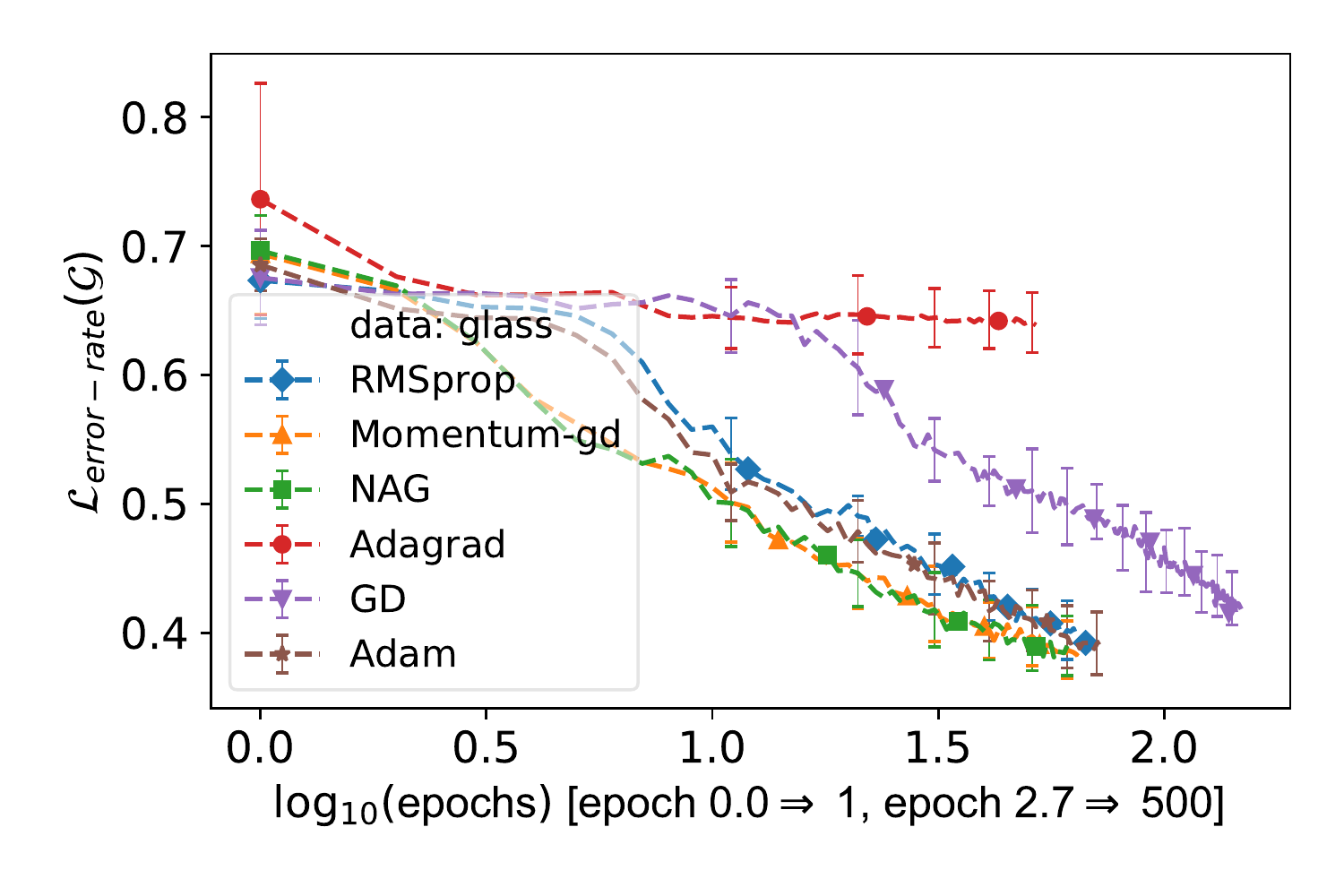}\label{sfig:mcpd}}
        \subfigure[{\scriptsize  MLP: ReLU, $\eta = 0.1$}]{
            \includegraphics[width=0.32\linewidth]{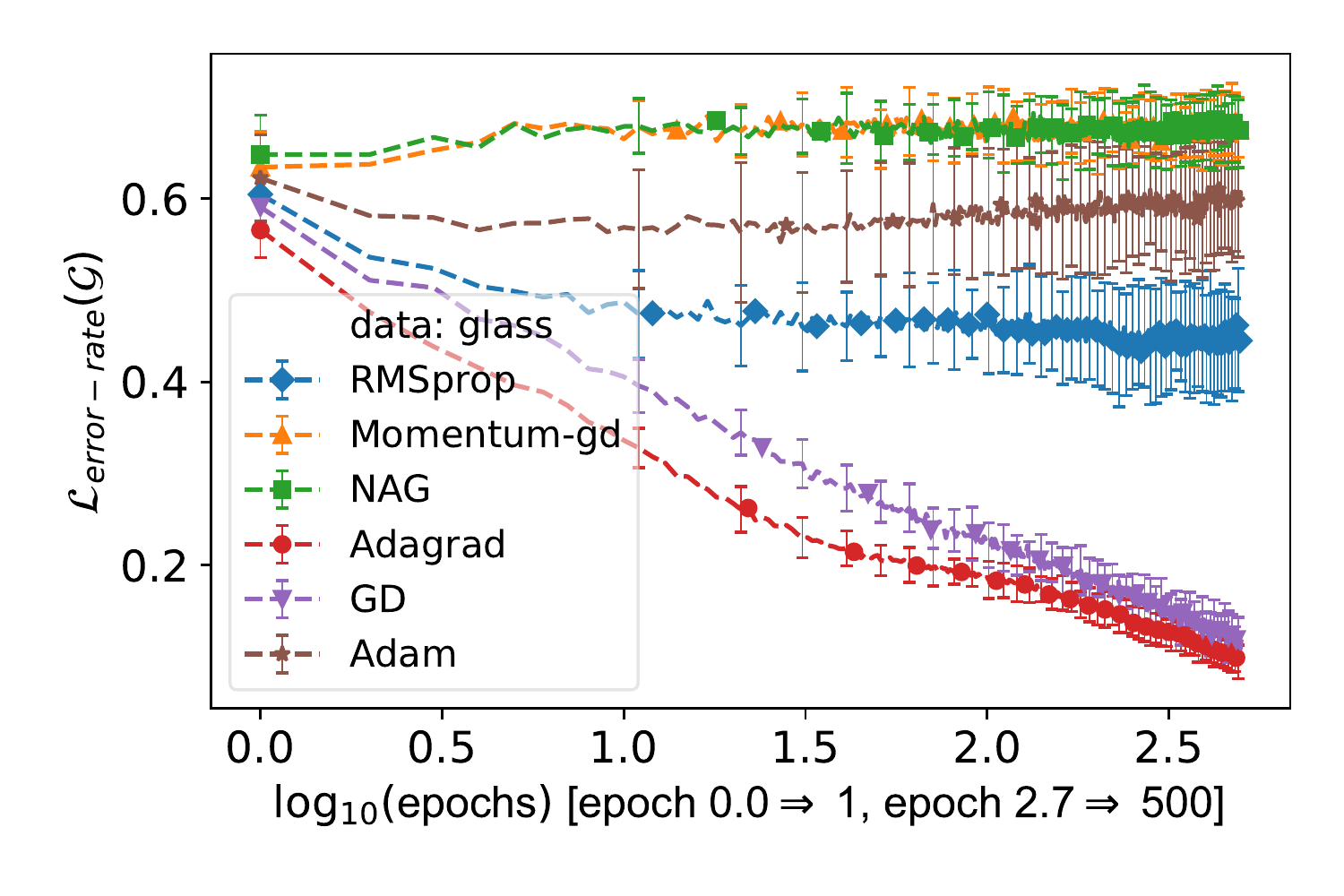}\label{sfig:mcr}}
        %\caption{Comparison of convergence of optimizer between BNeuralT and MLP and between optimizers over varied learning rate setting and activation function usage.\label{fig:class_comparison}}
        %\end{figure} 
        %
        %\begin{figure}
        
        \subfigure[{\scriptsize  BNeuralT: Sigmod, $\eta = 0.1$}]{
            \includegraphics[width=0.32\linewidth]{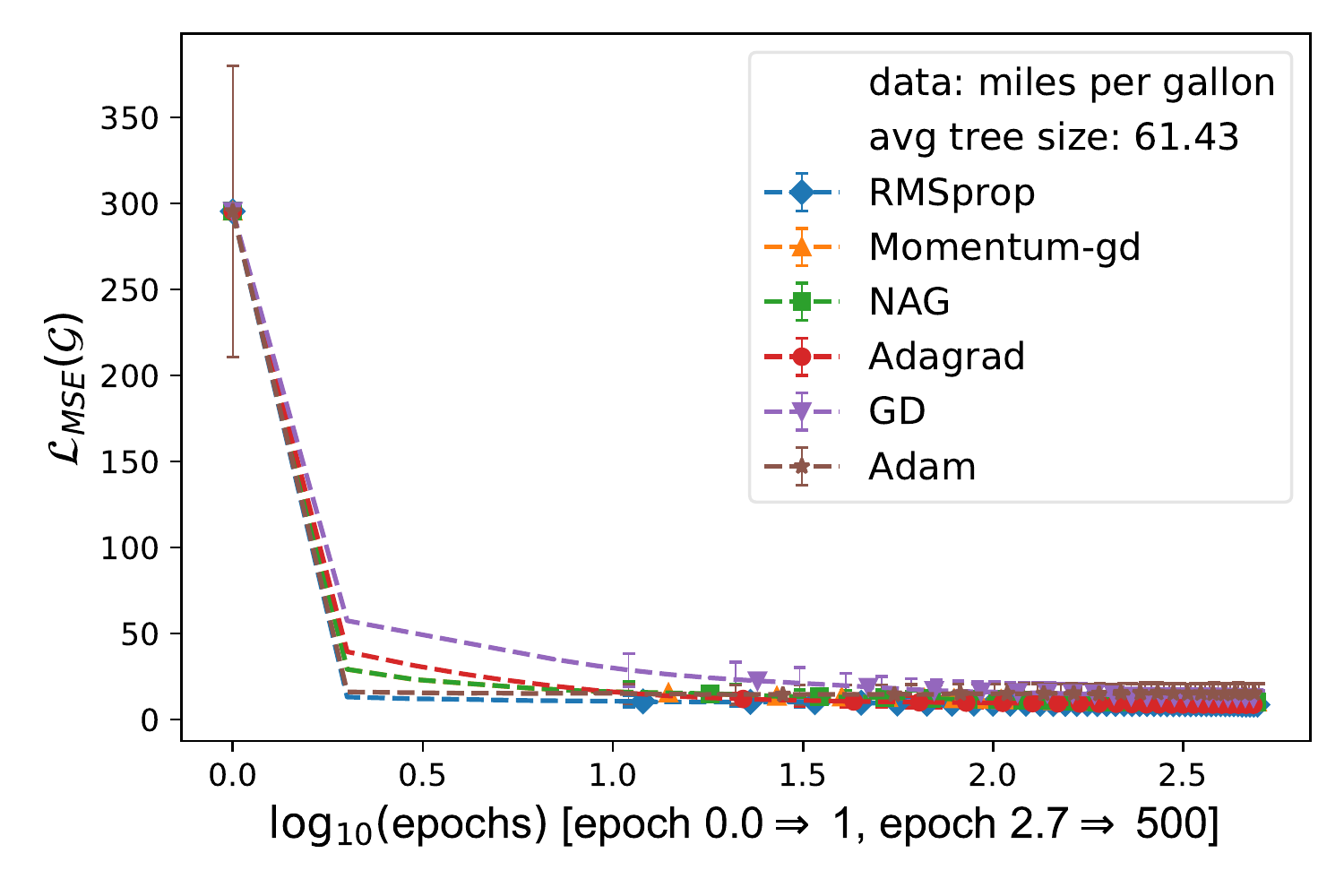}\label{sfig:brp1}}
        \subfigure[{\scriptsize  BNeuralT: Sigmod, $\eta = default$}]{
            \includegraphics[width=0.32\linewidth]{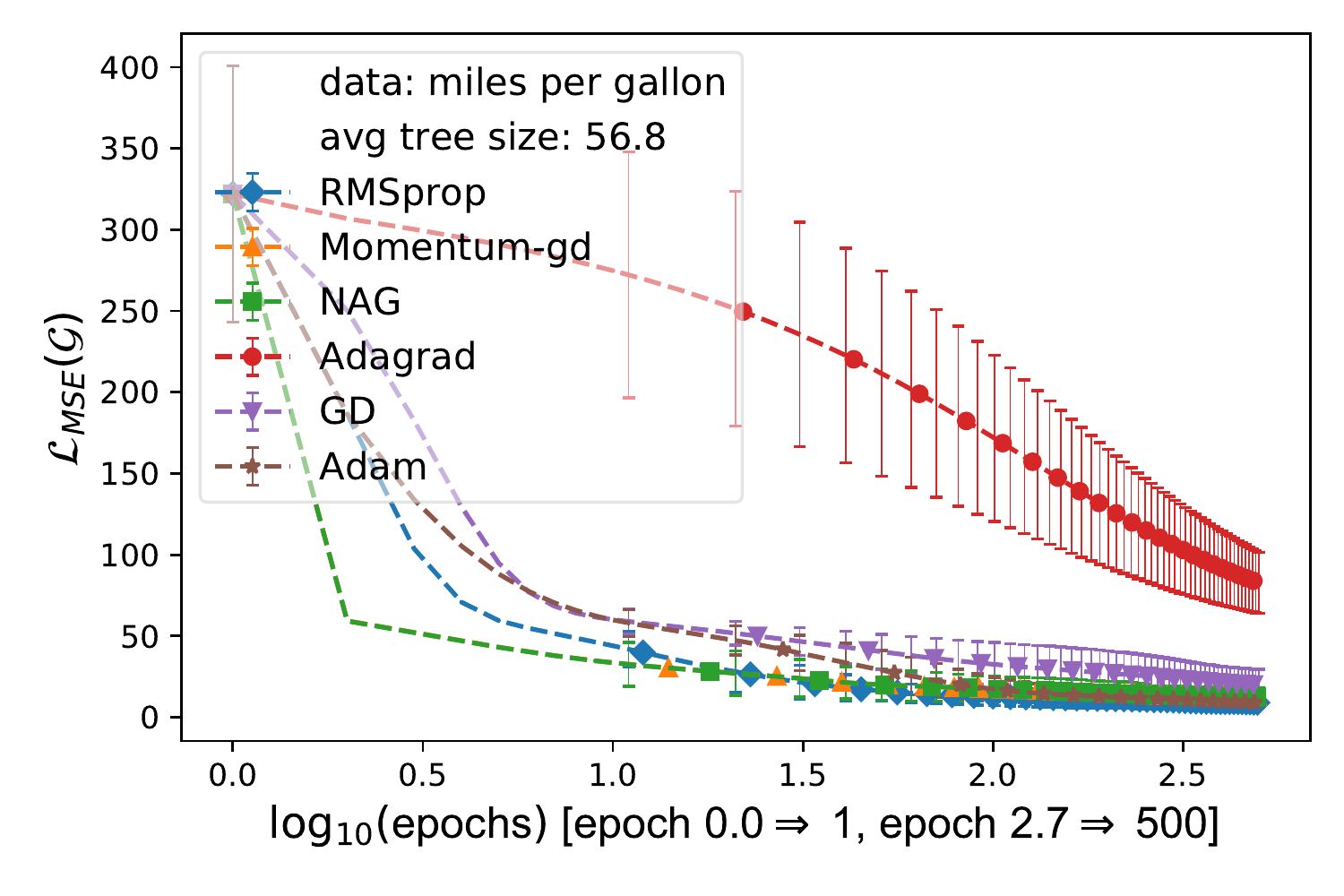}\label{sfig:brpd}}
        \subfigure[{\scriptsize  BNeuralT: ReLU, $\eta = 0.1$}]{
            \includegraphics[width=0.32\linewidth]{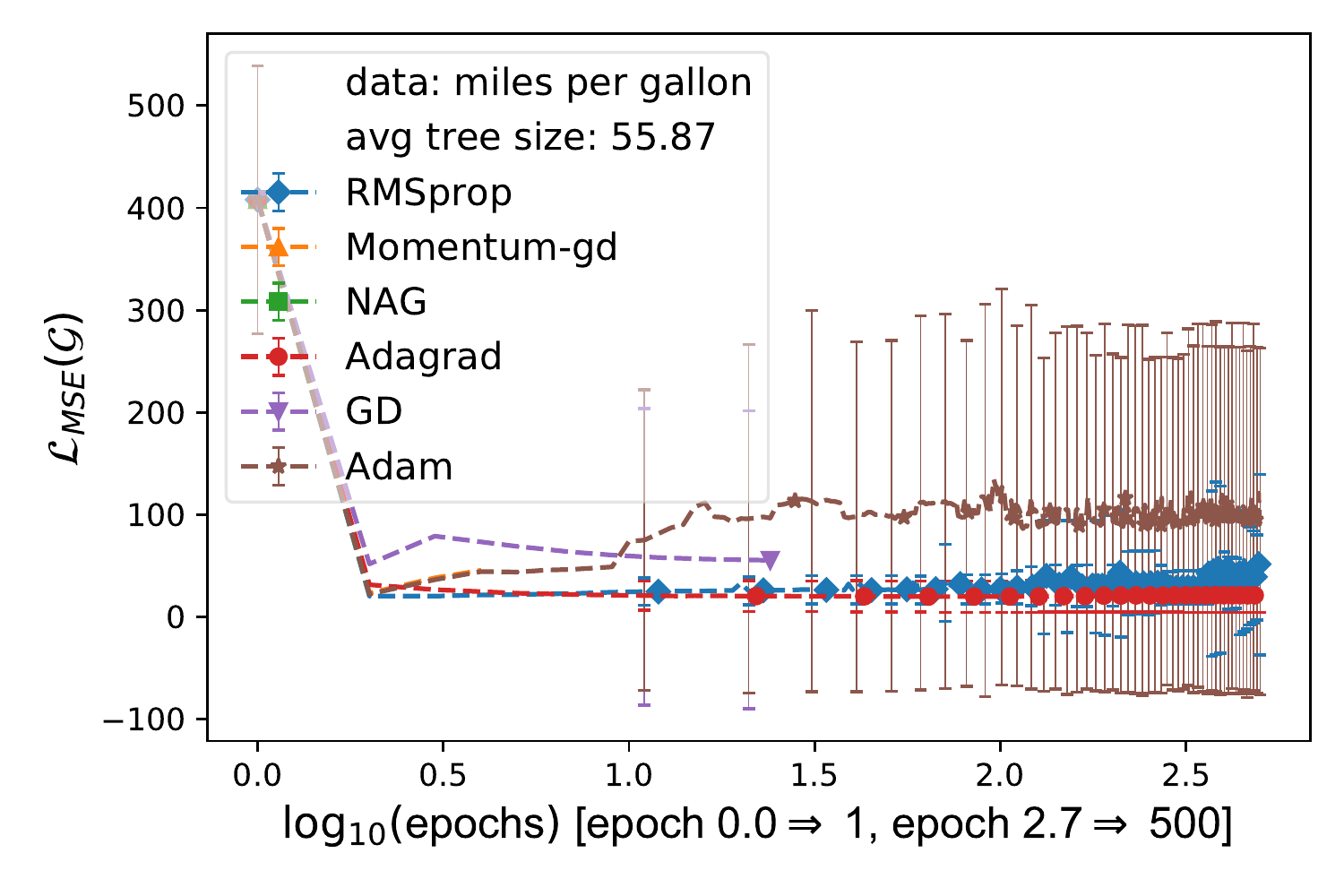}\label{sfig:brr}}
        
        \subfigure[{\scriptsize  MLP: Sigmod, $\eta = 0.1$}]{
            \includegraphics[width=0.32\linewidth]{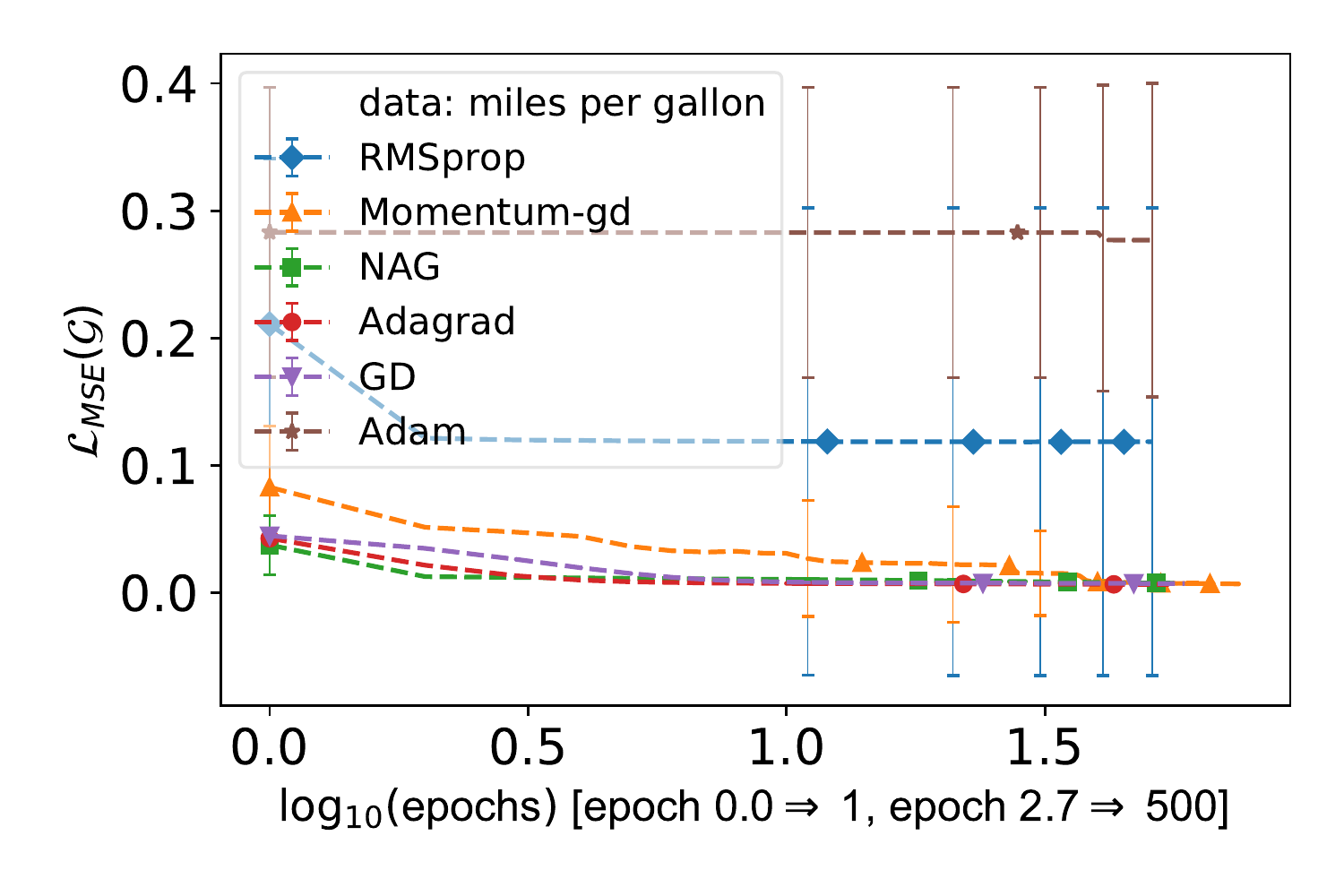}\label{sfig:mrp1}}
        \subfigure[{\scriptsize   MLP: Sigmod, $\eta = default$}]{
            \includegraphics[width=0.32\linewidth]{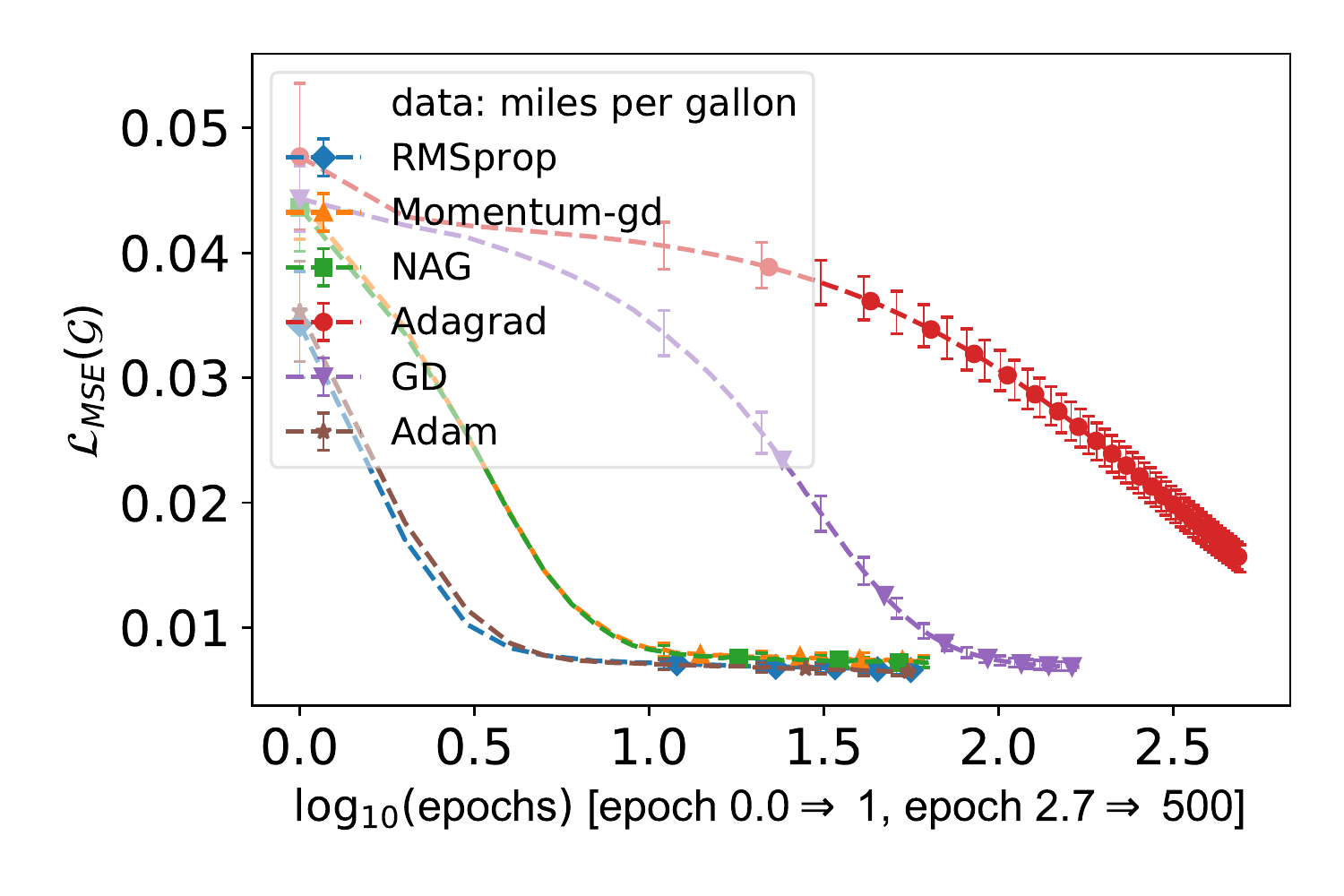}\label{sfig:mrpd}}
        \subfigure[{\scriptsize   MLP: ReLU, $\eta = 0.1$}]{
            \includegraphics[width=0.32\linewidth]{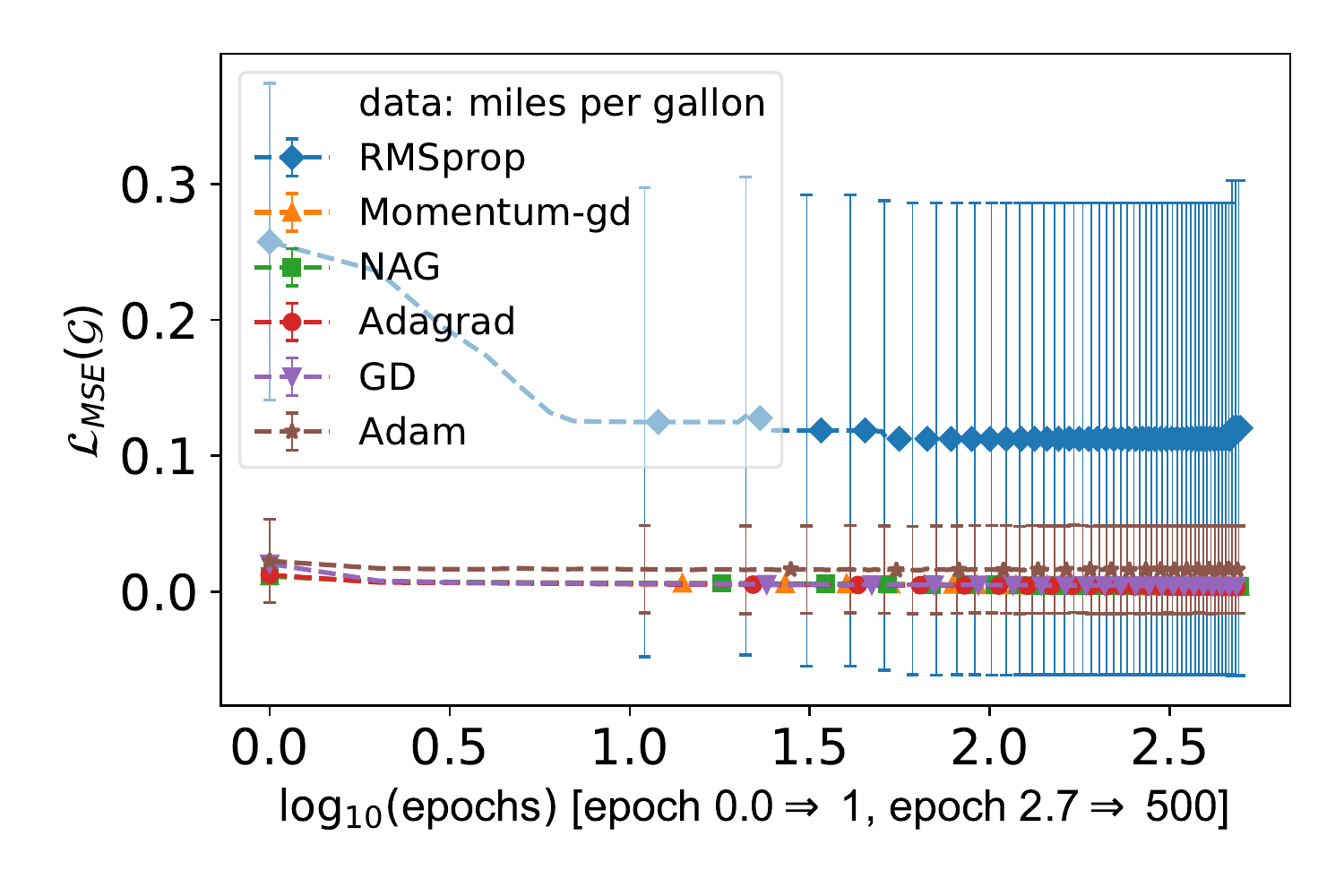}\label{sfig:mrr}}
        \caption{Comparison of convergence of optimizers for optimizing BNeuralT and MLP and  optimizers convergence over varied learning rate settings and activation function usage. (a)--(f) Classification problem where y-axis is an error rate. (g)--(l) Regression problems where y-axis is an MSE. The x-axis, $\log_{10}(\mbox{epochs})$ has the range $[0.0, 2.7]$ and is the training epochs $1$ to $500$. For MLP, early-stopping method show values for optimizes only upto the epochs where training stopped. Learning rate $\eta$ value $default$ indicates that RMSprop, Adam, and Adagrad has a value of $0.001$ as their learning rate and GD, NAG, and MGD has a value of $0.01$.
            \label{fig:active_sgd_comparison}}
    \end{figure}

\begin{figure}[]
        \centering
        \includegraphics[width=0.98\linewidth]{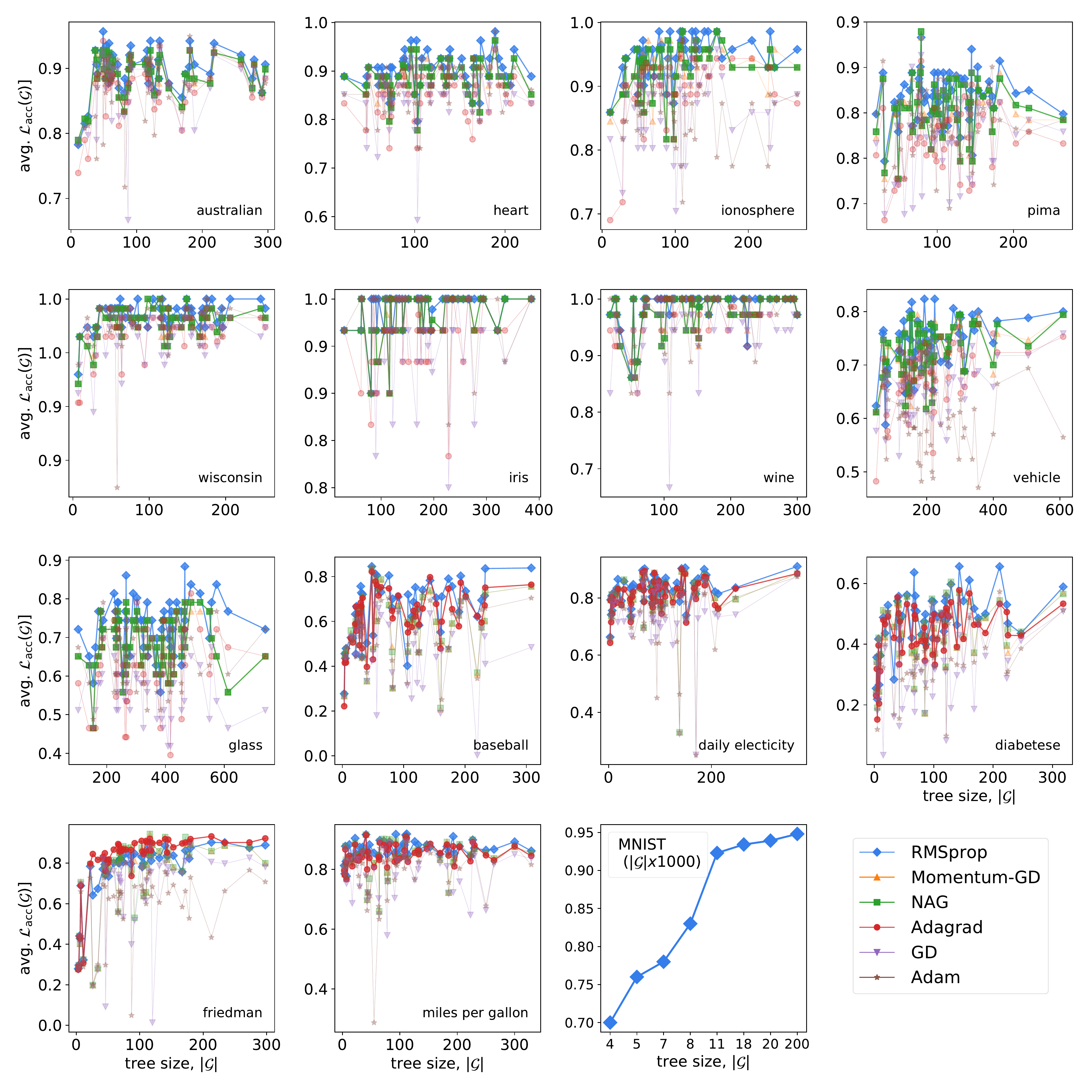}
        \caption{BNeuralT model's tree size $|\mathcal{G}|$ (x-axis) compared with its average (avg.) accuracy $\mathcal{L}_{\text{acc}}(\mathcal{G}) = (1 - \mathcal{L}_{\text{Error}}(\mathcal{G}))$ (y-axis has range $[0,1]$) on the training sets of nine classification problems, five regression problems, and one pattern recognition problem.  For the pattern recognition problem, BNeuralT model's tree size $|\mathcal{G}|$ (x-axis) is in the multiple of $1000$. The optimizers RMSprop, MGD, NAG, Adagrad, GD, and Adam are respectively indicated in blue, orange, green, red, purple, and brown colors, respectively, with symbols diamond, triangle, circle, downward triangle, and star.  For a few cases, convergence is linear to tree size, however for a few, high accuracy is achieved with smaller trees. For MNIST, RMSprop has 10-epoch of stochastic online training and shows a linear relation with tree size. In all tasks, the convergence of RMSprop (blue diamonds) is the best, followed by NAG (green squares) and MGD (yellow triangles). Adagrad (red circles) shows competitive performance with  RMSprop (blue diamonds) for regression problems. Classification datasets have green and blueish hues because NAG and RMSprop are top optimizers, and for regression datasets, plots appear to be red and blueish hue because of RMSprop and Adagrad are top optimizers.
        \label{fig:BneuralT_tree_size_vs_acc}}
        
    \end{figure}

    We evaluated average asymptotic convergence profiles of all six SGD optimizers for optimizing BNeuralT on classification and regression problems (cf. Figs.~\ref{fig:BneuralT_class_covg},  ~\ref{fig:BneuralT_reg_all}, and \ref{fig:active_sgd_comparison}). For such an analysis, we recorded training and test accuracies of each training epoch. Since we ran algorithms for 30 independent instances, we analyzed the average trajectory of all 30 runs. In each run, an ad hoc BNeuralT architecture was generated, which could vary  in tree size between a minimum ``outputs $\times$ 2'' nodes to a maximum $(m^{p+1} - 1)/(m-1)$ nodes. Hence, BNeuralT architecture and trainable parameters varied stochastically at each instance of the experiment. Such high entropy network architectures pose difficulties for SGDs to perform well consistently. We, therefore, investigated BNeuralT models' accuracy against their architecture (number of parameters) (cf.~Fig.~\ref{fig:BneuralT_tree_size_vs_acc}).
    
    \textbf{BNeuralT classification models convergence.} Fig.~\ref{fig:BneuralT_class_covg} shows convergence (training and test errors) profiles of ES training of BNeuralT having \textit{sigmoid} nodes, $0.1$ learning rate, and $0.4$ leaf generation rate. With this BNeuralT's setting, we observe that BNeuralT's RMSprop converges the fastest among all SGDs. RMSprop also outperformed all other optimizers. NAG and MGD were asymptotically closer to RMSprop optimizer. Like RMSprop, NAG and MGD showed monotonically increasing training convergence. However, on the test sets, we observe that the models started overfitting.  This motivated us to use early-stopping with restore best. Adagrad showed the most interesting convergence profile as initially, it had worse convergence among all optimizers, and while approaching higher epochs, it started rapidly improving its convergence. Thus, over an asymptotic behavior, Adagrad converged to a similar accuracy to that of RMSprop's accuracy. The optimizers NAG and MGD behave equivalently. Adam and GD were found sensitive to BNeuralT architecture (and trainable parameters).
    
    \textbf{BNeuralT regression models convergence.} Fig.~\ref{fig:BneuralT_reg_all} shows convergence (training and test errors) profiles of ES training of BNeuralT having \textit{sigmoid} nodes, $0.1$ learning rate, and $0.4$ leaf generation rate.  The convergence profiles of optimizers on five regression problems suggest that RMSprop and Adagrad were better converging optimizers. Similar to its classification problems profile, RMSprop converged faster than other optimizers for regression problems. Adagrad showed slower convergence than RMSprop. However, unlike its performance on classification problems, Adagrad showed a more stable convergence profile for regression. Contrary to classification problems, overfitting occurred only occasionally for  regression problems when comparing training and test convergence profiles.
    
    \textbf{BNeuralT and MLP settings convergence.}
    In Fig.~\ref{fig:active_sgd_comparison}, we compare convergence of six optimizers for optimizing both BNeuralT and MLP on various settings. We show this comparison on ``glass'' and ``miles per gallon'' datasets as an example. (Supplementary shows convergence of all other datasets on various settings.) In Fig.~\ref{fig:active_sgd_comparison}, we observe that the learning rate 0.1 produces stable convergence for all optimizers. For learning rate 0.1, Adam does not converge as good as other algorithms (cf. Figs.~\ref{sfig:bcp1} and~\ref{sfig:mcp1}). However, Adam does converge when learning rate is 0.001 (cf. Figs.~\ref{sfig:bcpd}, \ref{sfig:mcpd}, \ref{sfig:brpd}, and \ref{sfig:mrpd}). For learning rate 0.001, Adagrad does not converge as good as others. It may be observed that Adagrad's adaptively decreasing learning rate property made the convergence very slow and it may require more epochs to converge (cf. Figs.~\ref{sfig:bcpd}, ~\ref{sfig:mcpd}, ~\ref{sfig:brpd}, and \ref{sfig:mrpd}). 
    
    \textbf{Convergence of optimizers on ReLU.}
    For \textit{ReLU} activation function, Adagrad and GD being the slowest converging optimizers performed better than other faster converging optimizers like RMSprop, Adam, and NAG (cf. Figs.~\ref{sfig:bcr}, ~\ref{sfig:mcr}, ~\ref{sfig:mrr} and \ref{sfig:mrr}). In fact, when using ReLU activation function for regression problems, BNeuralT suffered from exploding gradient issues when using optimizers like GD, MGD, NAG, and Adam during some instances of runs of some datasets. Adagrad, however, remained unaffected by exploding gradient issue. This is due to its decreasing convergence speed. BNeuralT's performance with ReLU, due to its high sparsity, was affected by exploding gradient effect more than the MLP, which showed more tolerance to exploding gradient effect due to its large number of parameters (cf. Supplementary Fig.~\ref{fig:PRM_BP_ReLU_ES_p5}). %Hence, BNeuralT showed} a decline in the model's performance for ReLU (cf. supplementary Fig.~\ref{fig:PRM_BP_ReLU_ES_p5}).
    
    \textbf{Convergence of accuracy against trainable parameters.} BNeuralT's tree size (proportional to trainable parameter) and test accuracy in Fig.~\ref{fig:BneuralT_tree_size_vs_acc} suggest that RMSprop compared to other optimizers can optimize ad hoc structure better. We observed that the accuracy of BNeuralT increases with increasing tree size. However, accuracy dropped for some outliers in the connected scatter plot in Fig.~\ref{fig:BneuralT_tree_size_vs_acc}. This was because many points were within a specific range.  For classification problems, except for RMSprop, NAG was another better optimizer. For regression problems, along with RMSprop, Adagrad was another better performing optimizer. BNeuralT's RMSprop optimizer showed rather more stable performance for stochastically varying architectures compared to other optimizers. 
    For the pattern recognition MNIST dataset, RMSprop optimizer was used, and it showed a linear increase in accuracy for increasing order of tree size.

%        \subsection{BNeuralT tree structure analysis} 
%        \lipsum[1-2]

    \section{Discussion} 
    \label{sec:discussion}   
     %%%%%%%%%%%   DISCUSSION    %%%%%%%%%%%%%%%
    We designed and investigated a learning system called BNeuralT capable of solving three classes of machine learning problems: classification, regression, and pattern recognition. We assessed the capability of this neural tree algorithm as a single neuron model approximating computational dendritic tree-like behavior (cf. Figs.~\ref{fig:bio_trees} and Fig.~\ref{fig:BNeuralT}). This algorithm can also be considered a highly sparse NN trained using SGD optimizers. To train BNeuralT using SGDs, we designed a recursive backpropagation algorithm. Therefore, we broadly assessed three aspects of a learning system, i.e., its performance on (i) stochastically generated highly sparse models, (ii) sigmoid and ReLU functions and their dendritic interactions with internal nodes, and (iii) optimizers asymptotic convergence behavior. We had a diverse range of classification and regression problems and algorithms to compare BNeuralT's capabilities over these dimensions. 
    
    Since BNeuralT resembles a highly sparse NN, its performance was assessed against MLP (and MLP with dropout rate similar to the  probability of keeping nodes in BNeuralT) for their similar versions of SGD training. Six classification trees of BNeuralT, among all other algorithms and experiments, were top-performing models with a very low number of parameters. In fact, BNeuralT performed better against MLP dropout on classification problems, and it statistically had a similar performance on regression problems. This BNeuralT's performance against MLP's dropout regularization technique confirms that 
    \begin{quote}
    	    \textit{stochastic gradient descent training of any a priori arbitrarily ``thinned'' network has the potential to solve machine learning tasks with equivalent or better degree of accuracy than a fully connected symmetric and systematic NN architecture}. 
    \end{quote}

    We used six different SGD optimizers for optimizing BNeuralT and MLP. Each optimizer behaved differently in terms of their asymptotic convergence depending on what problem they solve, how their learning rate behaved over the training epochs, and what activation function was used  (cf. Figs.~\ref{fig:BneuralT_class_covg}, \ref{fig:BneuralT_reg_all}, and \ref{fig:active_sgd_comparison}). For example, with a $0.1$ learning rate, RMSprop was the best among others for BNeuralT optimization over classification problems. For regression problems, both RMSprop and Adagrad performed well. Adagrad, however, was slow on classification problems. Since optimizers had to optimize the same architecture in an instance, it may be the continuous-variable output in the case of regression problems has helped Adagrad perform better than the discrete variable output in classification problems.
    
    The use of activation functions influenced the performances of SGD optimizers. The sigmoid function proved to be more efficient with RMSprop, NAG, and MGD.  Whereas ReLU proved to be efficient with Adagrad. This  may be related to Adagrad's slow convergence speed that avoided weights to explode too quickly compared to faster converging optimizers like RMSprop (cf. Fig.~\ref{fig:active_sgd_comparison}(a-b), \ref{fig:active_sgd_comparison}(d-e), \ref{fig:active_sgd_comparison}(g-h)). This phenomenon of Adagrad may be confirmed since GD being the slowest converging SGD, was also found efficient when ReLU is used (cf. Fig.~\ref{fig:active_sgd_comparison}(c, f, and e)).
    Additionally, Adagrad converged better with a learning rate of 0.1 than 0.001 (e.g.  Fig.~\ref{fig:active_sgd_comparison}(a-b)). This is because Adagrad was too slow at earlier epochs that prevented it from converging within a fixed number of training epochs. 
    
    BNeuralT is operationally similar to HFNT and MONT algorithms. The HFNT and MONT algorithms model structures were genetically optimized  as opposed to BNeuralT structure. The better performance of BNeuralT compared to HFNT and MONT shows that the stochastic structure of BNeuralT has a high potential to solve machine learning problems (cf. Table~\ref{tab:BNeuralT_all_results}). However, this performance comparison also shows that BNeuralT models can be further compacted because both HFNT and MONT on classification problems had smaller average tree sizes than BNeuralT. This confirms that optimization of structure made HFNT and MONT more compact, although their accuracies were slightly compromised. On regression problems, however, BNeuralT performed better than HFNT both in terms of tree size and regression fit.  
    
    BNeuralT's performance compared to MLP's (with and without dropout) models and genetically optimized HFNT and MONT models confirms \textit{Occam's razor principle of parsimony} for machine learning model selection that the simple models possess better generalization capability than the complex models~\citep{blumer1987occam}. Indeed, it is similar to the \textit{sparsity of the biological brain} that a sparse network generalizes better or as good as a dense network~\citep{friston2008hierarchical,herculano2010connectivity,hoefler2021sparsity}. Moreover, it has been argued that a dense network is often overparameterized, and only a minute fraction of it is required for generalization~\citep{denil2013predicting}. Our result is in a similar line because BNeuralT, with only an average of $222$ parameters, which is \textit{only} $13.5\%$ of parameters than that of MLP's average $1638$ parameters, is able to generalize machine learning problems better or with similar accuracy than MLP. Additionally, the sparsity and compactness of BNeuralT models reduce memory usage and CO$_2$ footprint as they require less memory and computational resources than dense networks.
    
    The decision tree algorithms DT and RF (ensemble of DTs) computationally have dedicated paths from the root to leaves~\citep{breiman1984classification,breiman2001random}. BNeuralT  computationally also has dedicated information processing paths but from leaves to root. Although these algorithms differ in how nodes propagate information, a performance comparison suggests that BNeuralT has superior or competitive performances compared with DT and RF (cf. Table~\ref{tab:BNeuralT_all_results}). This performance is noticeable since RF is an ensemble algorithm that, using bootstrapping, combines 100 DTs to construct a predictor~\citep{breiman2001random}. Hence, the better performance of a standalone randomly generated BNeuralT model shows its high capabilities. Especially when RF being an ensemble of many trees, is more complex than a small and compact BNeuralT tree. Moreover, DTs are symbolic machine learning algorithms whose models offer inference ability as opposed to the black-box nature of NNs because of their ability to induce data using dedicated paths from the root to leaves. Likewise as shown in Figs.~\ref{fig:tree_image_class} and \ref{fig:tree_image_reg}, 
    \begin{quote}
    \textit{BNeuralT has dedicated information processing paths from leaves to root, and such paths related to particular subsets of inputs may be analyzed, which potentially may offer inference ability to BNeuralT.}
      \end{quote}
    
    Thus, BNeuralT models are potentially inferable as opposed to NNs. However, this is a challenging task since BNeuralT's nodes combine inputs and perform a nonlinear or linear transformation. 
    
    We assessed BNeuralT performance against GP, NBC, and SVM. These three algorithms take Gaussian kernels. That is, these algorithms have a powerful approach towards prediction. GP and NBC algorithms are robust and powerful algorithms if input data follow a normal distribution. Similarly, SVM uses Gaussian kernels to project input to high dimensions, increasing the separability of data points to help to classify them~\citep{cortes1995support}. Better performance of BNeuralT compared to these algorithms on classification and regression problems (cf. Table~\ref{tab:BNeuralT_all_results}) suggests that BNeuralT offers an efficient alternative to these algorithms as BNeuralT does not make any assumption about data to generate a hypothesis (model)  when fitting or classifying data.  
    
    The biologically plausible design of BNeuralT comes from its structural arrangement that takes random repeated input and has a computational dendritic tree-like organization with sigmoidal nonlinearities or ReLU linearity through its internal nodes~\citep{london2005dendritic}. The biologically plausible computational dendritic tree-like models 1-tree and 32-tree have a regular structural arrangement where repeated inputs are fed to a neuron systematically to form a tree structure~\citep{jones2021might}. Whereas BNeuralT takes randomly generated inputs and takes a non-systematic stochastic approach to its tree construction (cf. Fig.~\ref{fig:bio_trees}). Moreover, BNeuralT works on multi-class classification; and 1-tree, 32-tree, and  A-32-tree models work on binary classification~\citep{jones2021might}.
    
    BNeuralT's comparison with 1-tree,  32-tree, A-32-tree models, although limited, presents a noticeable performance. The error rate of BNeuralT on the MNIST dataset on all ten classes classification was $6.08\%$ with $23835$ parameters. The error rates of 1-tree, 32-tree, A-32-tree models on the binary classification of classes 3 and 5 of the MNIST dataset were reported as $7.8\%$,  $3.65\%$,  and  $8.89\%$, respectively, and they had $2047$, $65504$, $65504$ parameters, respectively. This result confirms BNeuralT's potential to produce capable learning systems, especially when BNeuralT's structural randomness (cf. Fig.~\ref{fig:bio_trees}) is closer to the  randomness (if any) of biological computational dendritic-tree~\citep{travis2005regional}.

    \section{Conclusions}
    \label{sec:con}
    %%%%%%%%%%%   CONCLUSION    %%%%%%%%%%%%%%%
    We propose a new algorithm Backpropagation Neural Tree (BNeuralT).  
    Our BNeuralT algorithm plausibly has a \textit{biological dendritic tree-like} modeling capability. It has a \textit{single neuron-like} model with sigmodal dendritic nonlinearities or rectified linear unit (ReLU) based dendritic linearity. It uses random repeated inputs at the leaves of subtrees attached to a single neuron, which is the root of a tree. BNeuralT uses stochastic gradient descent (SGD) optimizers to optimize stochastically generated sparse tree structures that are potentially minimal subsets of neuron networks (NNs). We propose a \textit{recursive error backpropagation algorithm} to apply SGDs to train trees that require pre-order and post-order traversal in a depth-first-search manner for their forward pass and backward pass computations.
    
    The results showed that our stochastically generated biologically plausibly tree structure and recursive error backpropagation algorithm have the capacity to learn a wide variety of machine learning problems. Moreover, we show that any stochastically generated tree structures can learn machine learning problems with high accuracy, and structure optimization may only be required for making models more compact. However, there is a trade-off for compacting models as we found that making the models more compact means compromising on accuracy. Additionally, BNeuralT's strong performance compared to MLP's dropout regularization technique confirms that SGD training of any ``\textit{a priori}'' arbitrarily ``\textit{thinned network}'' (spares tree structures) has the potential to solve machine learning tasks with equivalent or better degree of accuracy.

    The sigmoidal dendritic nonlinearities (sigmoid function used at tree’s root and internal nodes) performed obviously better than a linear dendritic tree (sigmoid function used at tree’s root and ReLU at internal nodes). However, the linear dendritic tree differed from the best performing nonlinear dendritic tree by only about 10\% accuracy. Nevertheless, it was comparable with a few nonlinear dendritic tree models, especially with those trained with gradient descent (GD), momentum GD, and Adam. This shows that \textit{purely single node} BNeuralT models might solve machine learning problems efficiently.
 
    On MNIST (pixels) character classification dataset, BNeuralT, when loosely compared with 1-tree and 32-tree biologically plausible dendritic tree algorithms, was found competitive. Moreover, BNeuralT performed best among select tree-based classifiers for the classification of MNIST characters. On classification and regression problems, the overall performance of BNeuralT was better than some varied types of well-known algorithms: decision tree, random forest, Gaussian process, naïve Bayes classifier, and support vector machine. Such a performance of BNeuralT came from a minimal hyperparameter setup. Therefore, this work shows that our newly designed learning algorithm generates high-performing and parsimonious (therefore sustainable) models balancing the complexity with descriptive ability.

    %\section*{Declarations}
    %\label{sec:dec}
    %Authors declare no conflicts of interest. Supplementary provides all source code and data repository links. %Both authors contributed equally to this research.  

    %Authors declare no conflict of inters
    %All manuscripts must contain the following sections under the heading 'Declarations'.
    %
    %If any of the sections are not relevant to your manuscript, please include the heading and write 'Not applicable' for that section.
    %
    %To be used for non-life science journals
    %
    %Funding (information that explains whether and by whom the research was supported)
    %
    %Conflicts of interest/Competing interests (include appropriate disclosures)
    %
    %Availability of data and material (data transparency)
    %
    %Code availability (software application or custom code)
    %
    %Authors' contributions (optional: please review the submission guidelines from the journal whether statements are mandatory)

    {\small
        %\bibliographystyle{spmpsci}      % mathematics and physical sciences NOT THIS
        %\bibliographystyle{splncs04}
        \bibliographystyle{agsm} % Harvard Style 

        \bibliography{BNeuralT}
    }

    \clearpage
    \appendix
    \normalsize

    \renewcommand{\thetable}{\Alph{section}\arabic{table}}
    \renewcommand{\thefigure}{\Alph{section}\arabic{figure}}
    \setcounter{table}{0}
    \setcounter{figure}{0}
    
    \section{Supplementary}
    Supplementary has three Sections: Sec~\ref{sec:source_code} contains URLs of source code repositories, README, trained models, and links to data. Sec.\ref{sec:convergece_figs} offers convergence trajectories of BNeuralT and MLP. Sec.\ref{sec:results_table_other} supplements Table~\ref{tab:BNeuralT_all_results}.

    \subsection{Source code, scripts, pre-trained models, and data}
    \label{sec:source_code}
    \subsubsection{BNeuralT Code}  
    BNeuralT algorithm source code repository:
    \url{https://github.com/vojha-code/BNeuralT}
    has the following items, including training and evaluation entry points for re-producing results in Table~\ref{tab:BNeuralT_all_results} and Table~\ref{tab:BNeuralT_MNSIT}.
    
    \subsubsection{MLP and Other Algorithms Code and Scripts} 
    Scripts of MLP and other algorithms are available at:\\
    \url{https://github.com/vojha-code/BNeuralT/tree/master/source_mlp_tf}
    
    \subsubsection{Pre-trained Models (of Table 2 and Table 4)}
    Experiments produced pre-trained models and results are available at:\\
    \url{https://github.com/vojha-code/BNeuralT/tree/master/trained_models}
    
    \subsubsection{Data} 
    Classification and regression learning problems:\\ \url{https://github.com/vojha-code/BNeuralT/tree/master/data}\\
    For these datasets, the exact sequence of all 30 independent runs for training and test can be found in models files pre-trained models.
    
    Pattern recognition problem (MNIST):
    %\url{https://drive.google.com/file/d/10FtlGn6m1RkzCp4s6GfDlWzCvsPjcuVG/view}\\
    \url{http://yann.lecun.com/exdb/mnist/}

    \subsection{Supplementary Figures}
    \label{sec:convergece_figs}
    Figs.~\ref{fig:PRM_BP_Sig_ES_p5}, 
    \ref{fig:PRM_BP_Sig_ES_p5_default}, 
    \ref{fig:PRM_BP_ReLU_ES_p5}, 
    \ref{fig:MLP_Sig_ES_50_lr1},
    \ref{fig:MLP_Sig_ES_50_lrD}, 
    \ref{fig:MLP_Sig_ES_No_lr1}, and 
    \ref{fig:MLP_Sig_ES_No_lrD} 
    are the average training and test convergence performance computed over 30 independent runs for six optimizers of BNeuralT  and MLP over nine classification and five regression datasets. The x-axis in 
    Figs.~\ref{fig:PRM_BP_Sig_ES_p5}, 
    \ref{fig:PRM_BP_Sig_ES_p5_default}, 
    \ref{fig:PRM_BP_ReLU_ES_p5}, 
    \ref{fig:MLP_Sig_ES_50_lr1},
    \ref{fig:MLP_Sig_ES_50_lrD}, 
    \ref{fig:MLP_Sig_ES_No_lr1}, and 
    \ref{fig:MLP_Sig_ES_No_lrD}  
    are $\log_{10}(\mbox{epochs})$ that has range $[0.0, 2.7]$ and is the training epoch range $[1, 500]$. Optimizers RMSprop, MGD, NAG, Adagrad, GD, and Adam are indicated in blue, orange, green, red, purple, and brown, respectively, with symbols diamond, triangle, circle, downward triangle, and star. In each plot is labeled with the name of the dataset and set type. For example, iris (train) and iris (test) represent training set and test set convergence of optimizers on iris data.
    
    The performance of BNeuralT and MLP of classification problems is shown in %
    Figs.~\ref{fig:PRM_BP_Sig_ES_p5}(a), 
    \ref{fig:PRM_BP_Sig_ES_p5_default}(a), 
    \ref{fig:PRM_BP_ReLU_ES_p5}(a), 
    \ref{fig:MLP_Sig_ES_50_lr1}(a),
    \ref{fig:MLP_Sig_ES_50_lrD}(a), 
    \ref{fig:MLP_Sig_ES_No_lr1}(a), and 
    \ref{fig:MLP_Sig_ES_No_lrD}(a).
    The y-axis of each plot is ``$-\log_{10}(\mathcal{L}_{\text{Error}}(\cdot))$'' that has the range $[0,10]$ and is the log scale of the training and test  accuracies. An  accuracy of $99$\% (an error of $0.01$) on $-\log_{10}(\mathcal{L}_{\text{Error}}(\cdot))$ scale has a value of $2.0$, and an accuracy of $90$\% (an error $0.1$) has a value of $1.0$. Thus, a \textit{higher} value on the y-axis is better. Error bar is the standard deviation of $-\log_{10}(\mathcal{L}_{\text{Error}}(\cdot))$ value. 
    The performances of regression problems are shown in Figs.~\ref{fig:PRM_BP_Sig_ES_p5}(b), 
    \ref{fig:PRM_BP_Sig_ES_p5_default}(b), 
    \ref{fig:PRM_BP_ReLU_ES_p5}(b), 
    \ref{fig:MLP_Sig_ES_50_lr1}(b),
    \ref{fig:MLP_Sig_ES_50_lrD}(b), 
    \ref{fig:MLP_Sig_ES_No_lr1}(b), and 
    \ref{fig:MLP_Sig_ES_No_lrD}(b).  
    The y-axis of each plot is $\log_{10}(\mathcal{L}_{\mbox{MSE}}(\cdot))$, and it is the training and test sets MSE on the log scale. An MSE $0.01$ on the log scale has a value of $-2$. Thus, a lower value is better. Error bar is the standard deviation of $\log_{10}(\mathcal{L}_{\text{MSE}}(\cdot))$. In both classification and regression plots, a larger length of error bar shows higher stochasticity of an optimizer that indicates an optimizer's higher ability to skip local minima. Thus, it shows an optimizer's better convergence ability.
    \subsubsection{BNeuralT Convergence Trajectories}
\begin{figure}
        \centering
        \includegraphics[width=0.7\linewidth]{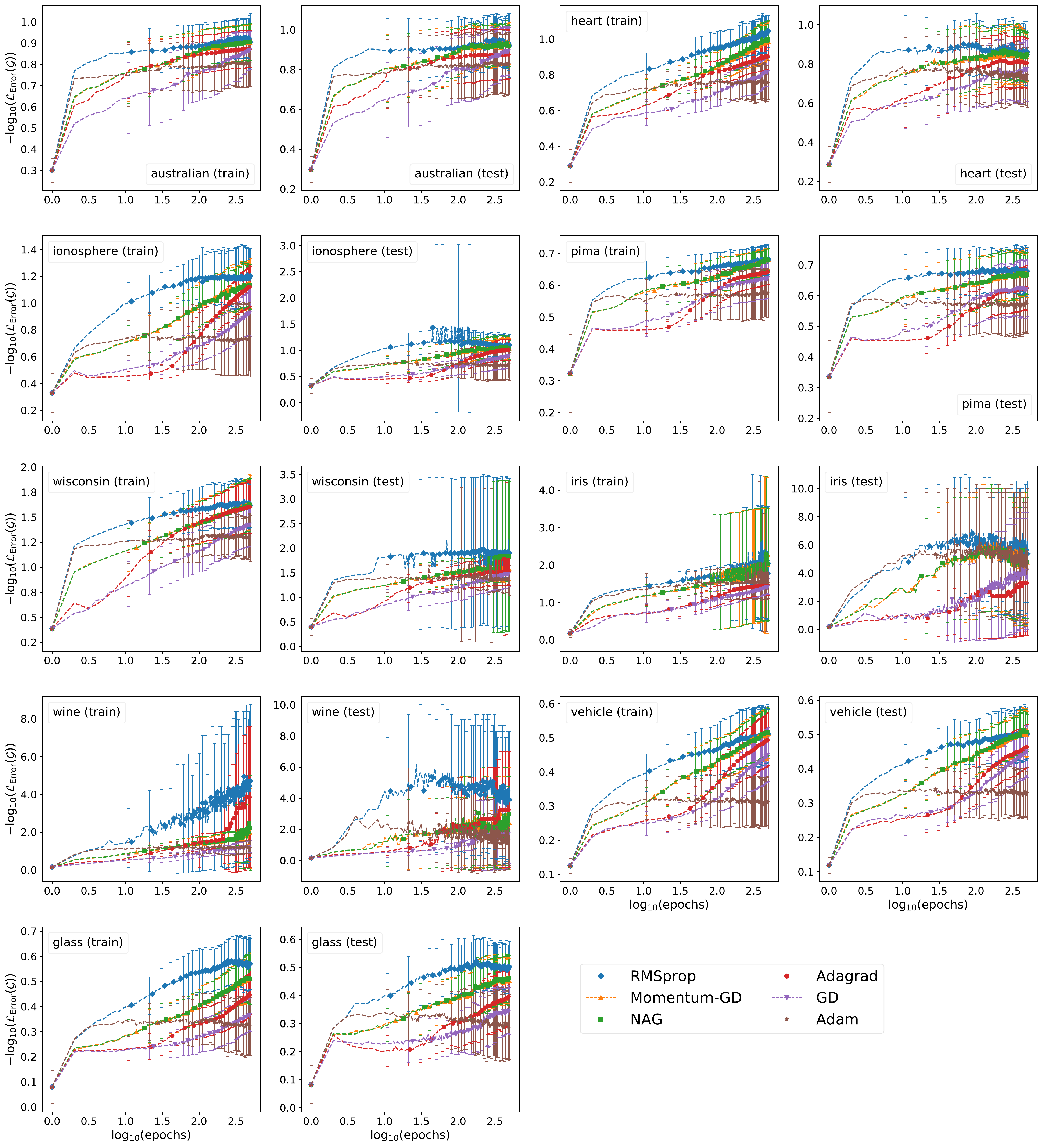}
        %\caption{BNeuralT classifiers with sigmoid internal nodes and 0.5 leaf generation rate at lower tree depth trained with learning rate 0.1 and early-stopping strategy.}
        %\label{fig:PRM_BP_Sig_ES_p5_class}
        
        (a)
        
        \centering
        \includegraphics[width=0.7\linewidth]{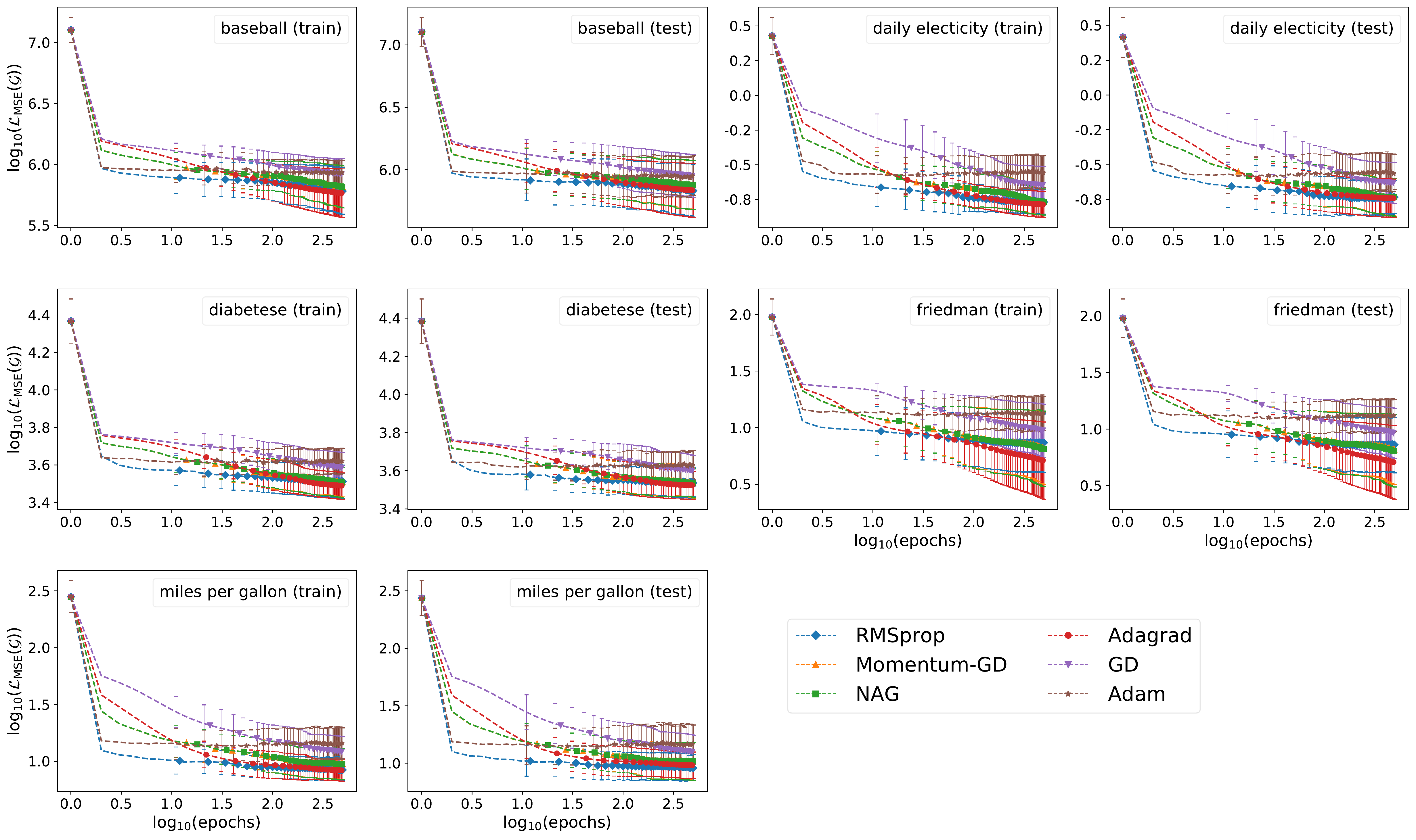}
        
        (b)
        \caption{BNeuralT (a) classification (b) regression models having \textit{sigmoid} nodes, $0.5$ leaf generation rate, $0.1$ \textit{learning rate}, and \textit{early-stopping} strategy setting.
        (a) RMSprop in blue is the fastest converging optimizer as it appears at the top line in plots. Adagrad in red shows slow convergence in the beginning and only rapidly improves at higher epochs. Adam and RMSprop have similar trajectories at earlier epochs, but Adam stays at local minima. (b) Adagrad and RMSprop both appear as bottom lines in graphs showing better convergence than others.  
        \label{fig:PRM_BP_Sig_ES_p5}}
    \end{figure}

\begin{figure}
        \centering
        \includegraphics[width=0.7\linewidth]{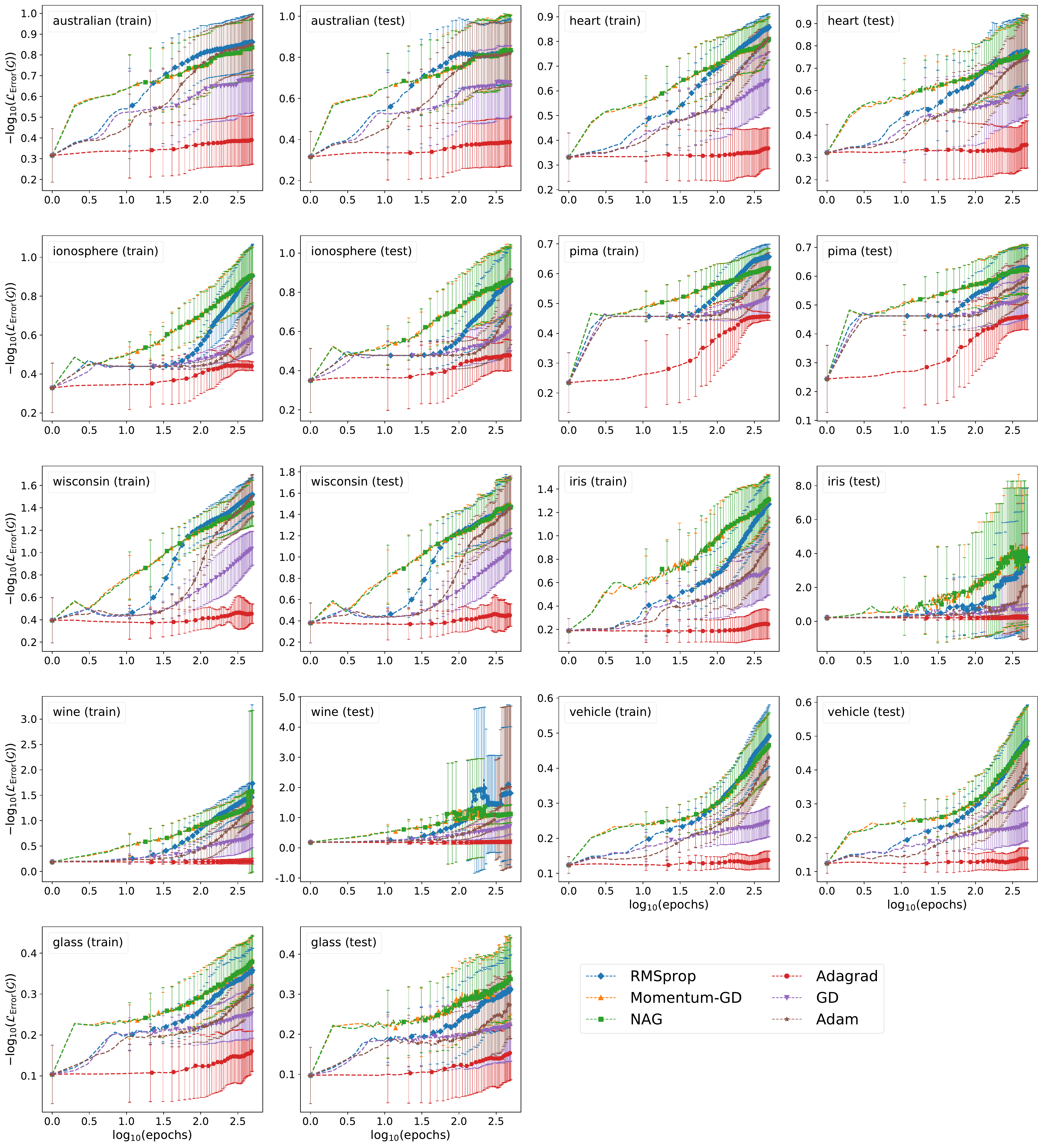}
        
        (a)
        %\caption{BNeuralT Sigmod P5 Lr default (low) Class}
        %\label{fig:PRM_BP_Sig_ES_p5_default_class}
        
        \centering
        \includegraphics[width=0.7\linewidth]{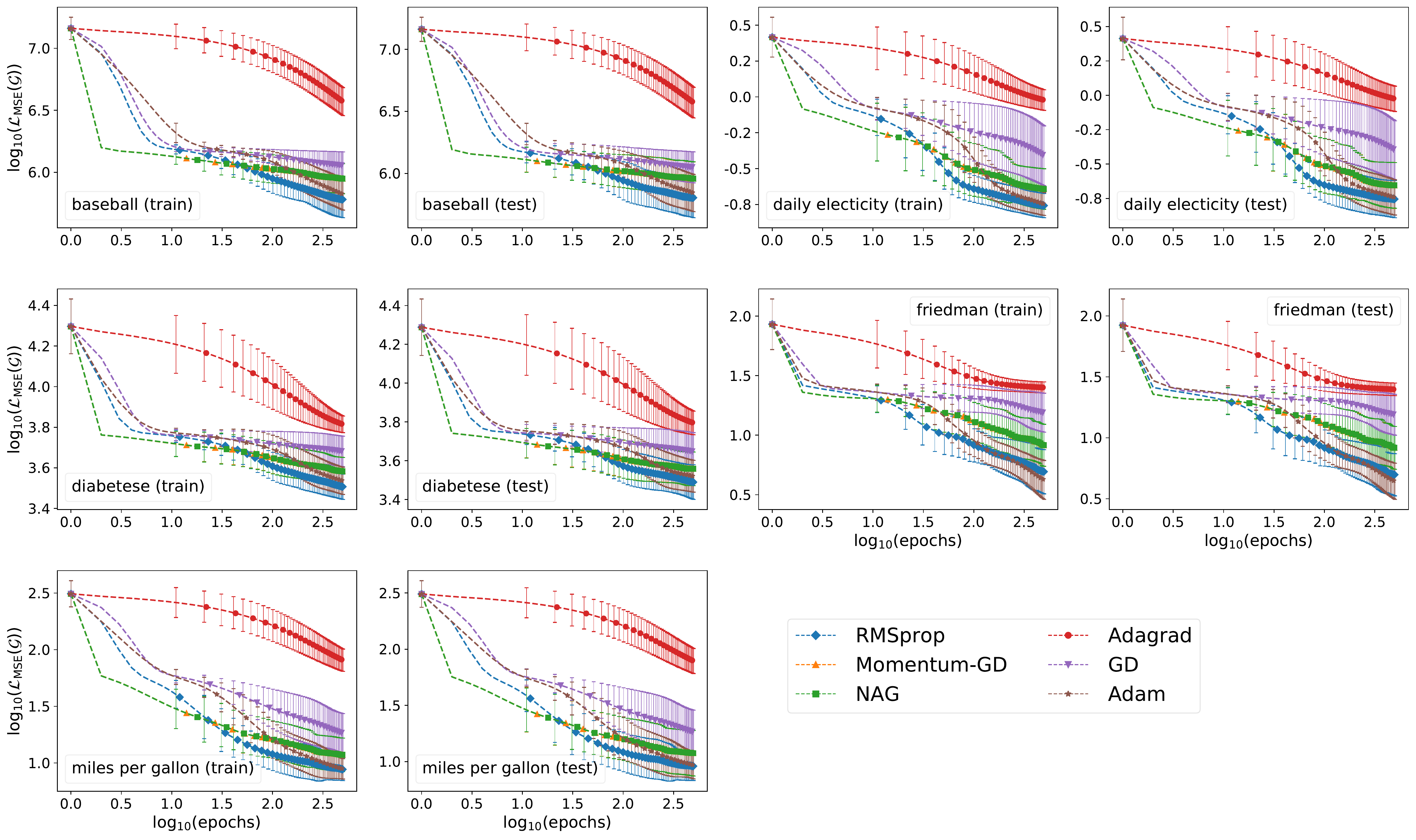}
        
        (b)
        \caption{BNeuralT (a) classification (b) regression models having sigmoid nodes, $0.5$ leaf generation rate, \textit{default learning rate}, and \textit{early-stopping} strategy setting.
        (a) NAG in green is the fastest converging and most dominant  optimizer in plots. Adagrad in red is the slowest. Adam and RMSprop have a similar trajectory, and both show rapid convergence at higher epochs. (b) RMSprop and NAG both appear as bottom lines in graphs showing better convergence than others. 
        \label{fig:PRM_BP_Sig_ES_p5_default}}
    \end{figure}

\begin{figure}
        \centering
        \includegraphics[width=0.7\linewidth]{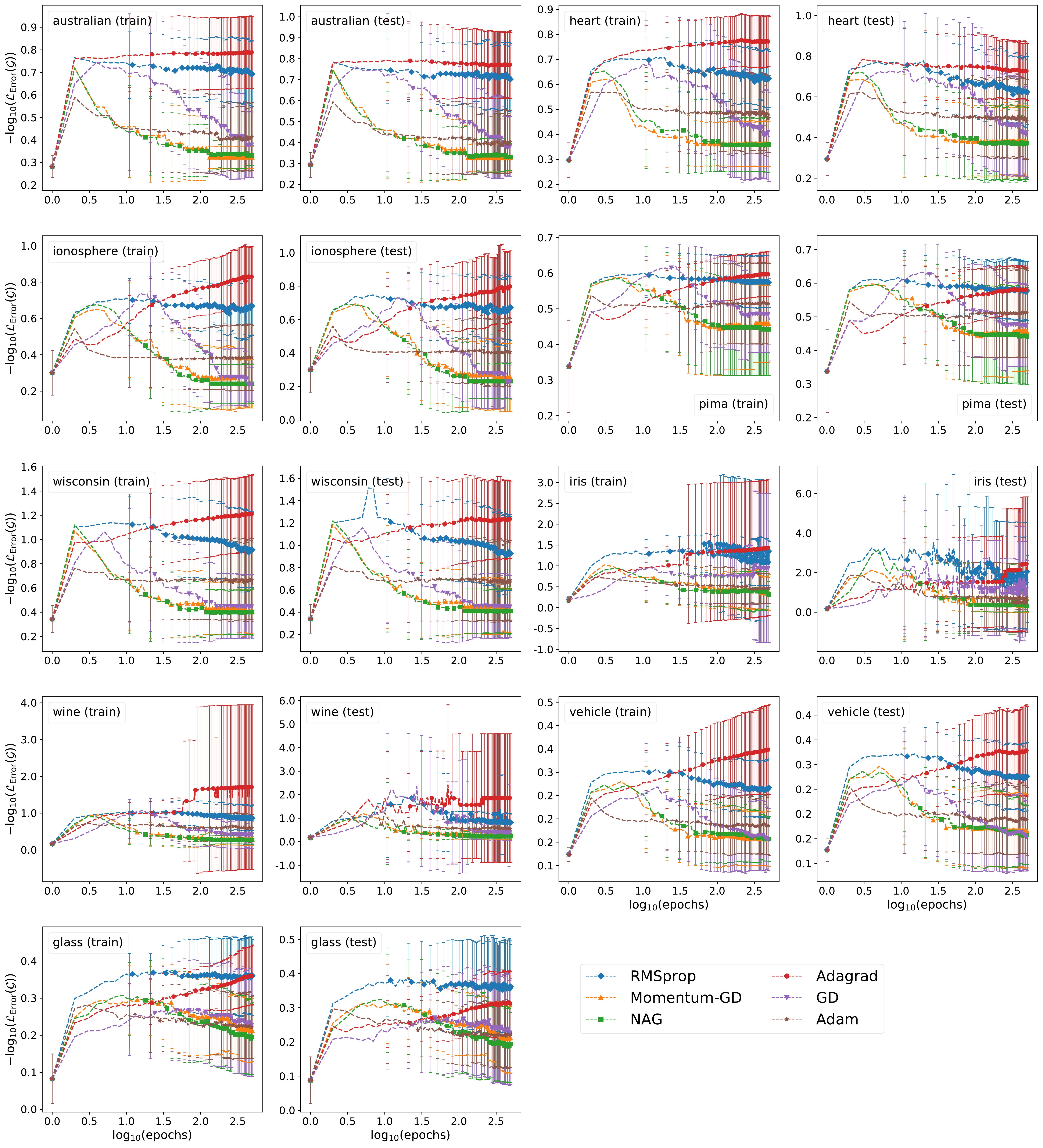}
        %\caption{BNeuralT ReLU P5 Lr 0.1 Class}
        %\label{fig:PRM_BP_ReLU_ES_p5_class}
        
        (a)
        
        \centering
        \includegraphics[width=0.7\linewidth]{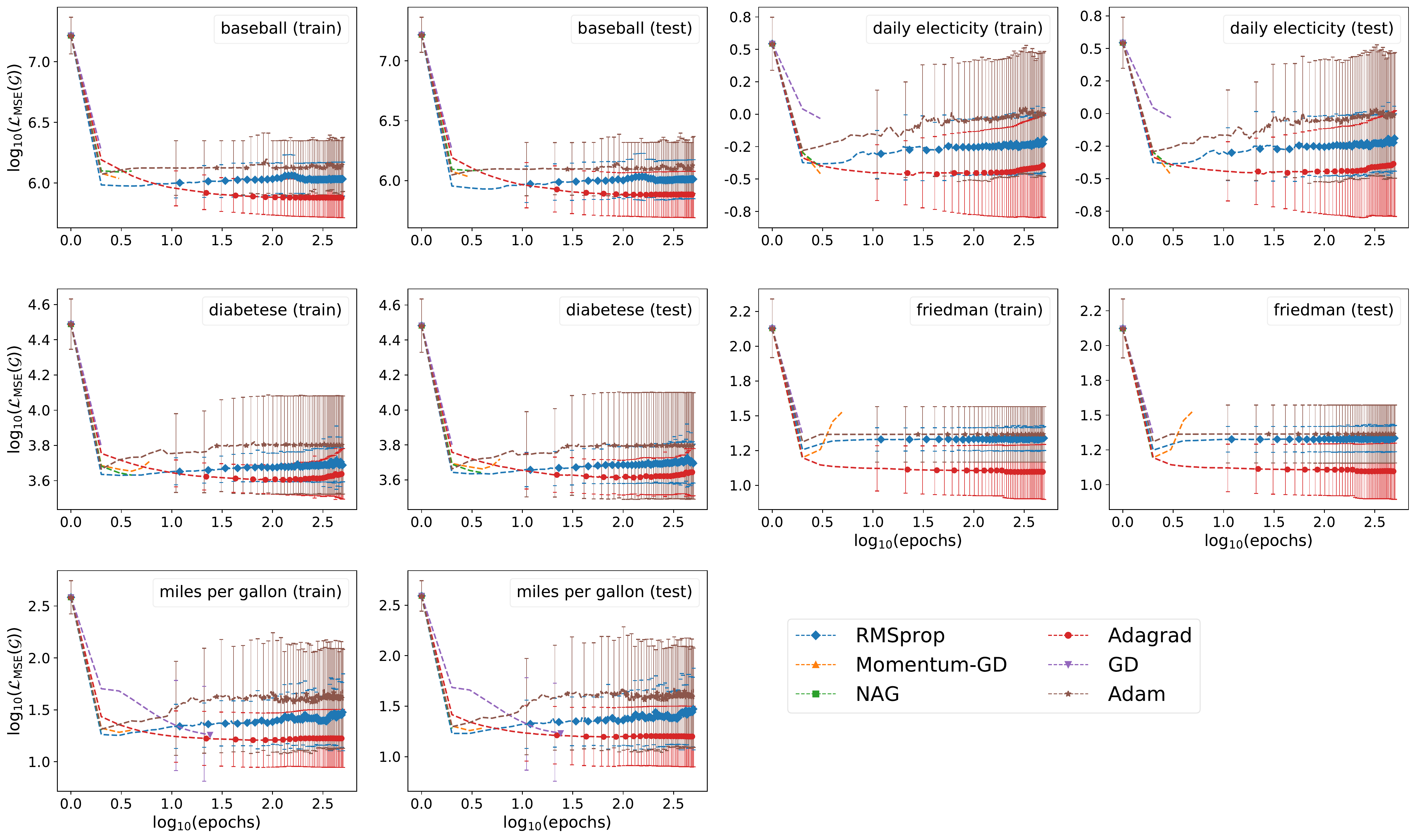}
        
        (b)
        \caption{BNeuralT (a) classification (b) regression models having ReLU nodes, $0.5$ leaf generation rate, $0.1$ \textit{learning rate}, and early-stopping strategy setting.
        (a) Adagrad in red is the fastest converging and most dominant  optimizer in plots. The downward curve of other algorithms, except Adagrad, indicates that, when ReLU was used, not all structures could be trained. This is because of exploding gradient issue. This is because of exploding gradient issue. Hence the average trajectories show a downward curve (trajectory) in performance after certain epochs. This phenomenon is indicated with a sudden upward curve in plots (b) for regression. (b) Adagrad in red is the fastest converging and most dominant optimizer in plots.
        \label{fig:PRM_BP_ReLU_ES_p5}}
    \end{figure}

    Figs.~\ref{fig:PRM_BP_Sig_ES_p5}, 
    \ref{fig:PRM_BP_Sig_ES_p5_default}, and
    \ref{fig:PRM_BP_ReLU_ES_p5}
    are BNeuralT models with their leaf generation rate at lower tree depth set to $0.5$, and they were trained with an early-stopping strategy. However, they varied in the following ways:
    % (i)
    Fig.~\ref{fig:PRM_BP_Sig_ES_p5} has the sigmoid functions as its internal nodes. All six SGDs are trained with a learning rate of $0.1$. 
    % (ii)
    Fig.~\ref{fig:PRM_BP_Sig_ES_p5_default} has the sigmoid functions as its internal nodes. In this setting, the optimizers RMSprop, Adam, and Adagrad had a learning rate of $0.001$, and optimizer MGD, NAG, and GD  had a learning rate of $0.01$.
    %(iii)
    Fig.~\ref{fig:PRM_BP_ReLU_ES_p5} has the ReLU functions as its internal nodes. All six SGDs are trained with a learning rate of $0.1$. 
    
    Figs.~\ref{fig:PRM_BP_Sig_ES_p5}, \ref{fig:PRM_BP_Sig_ES_p5_default}, and
    \ref{fig:PRM_BP_ReLU_ES_p5} 
    offer convergence profiles for both classification and regression problems for BNeuralT setting with higher leaf generation rates, i.e., $0.5$. For the higher leaf generation rate (smaller model size), BNeuralT with sigmoid node and $0.1$ learning rate convergence is similar to the lower leaf generation rate (larger models). However, for the \textit{default} learning rate (lower learning rate), BNeuralT convergence shows a different profile (cf.  Fig.~\ref{fig:PRM_BP_Sig_ES_p5_default}). For a lower learning rate, RMSprop (and Adam) is much slower at the beginning of training but converges very fast at the reminder of the training epochs. NAG and MGD show a monotonically increasing and stable learning profile. Interestingly, Adam (as evident from literature~\citep{diederik2015adam}) performed the best with a learning rate of $0.001$,  and as shown in Fig.~\ref{fig:BneuralT_class_covg}, this was not the case where the learning rate was $0.1$. Adam has a similar profile to RMSprop, but RMSprop produced better accuracy than Adam. Adagrad, with a lower learning rate, had worse performance among  all six SGDs.  
    
    BNeuralT's performance with \textit{ReLU} due to its high sparsity and loss nonlinearity show a decline in the model's performance (cf. Fig.~\ref{fig:PRM_BP_ReLU_ES_p5}). In fact, it seems to suffer from exploding gradient issues for some optimizers like GD, MGD, NAG, and Adam. Adagrad, however, remains unaffected by this issue when the ReLU function was used.
    
    %%%     BNeuralT  ES  p5 lr 0.1
    
    %%%     BNeuralT  ES  p5 lr default
    
    %%%     BNeuralT  ES  p5 lr 0.1 ReLU
    
    \clearpage
    %TODO %%%%%%%%%%%%%%%%%%%%%%%%%%%%%%%%%
    \subsubsection{MLP Convergence Trajectories}
\begin{figure}
        \centering
        \includegraphics[width=0.7\linewidth]{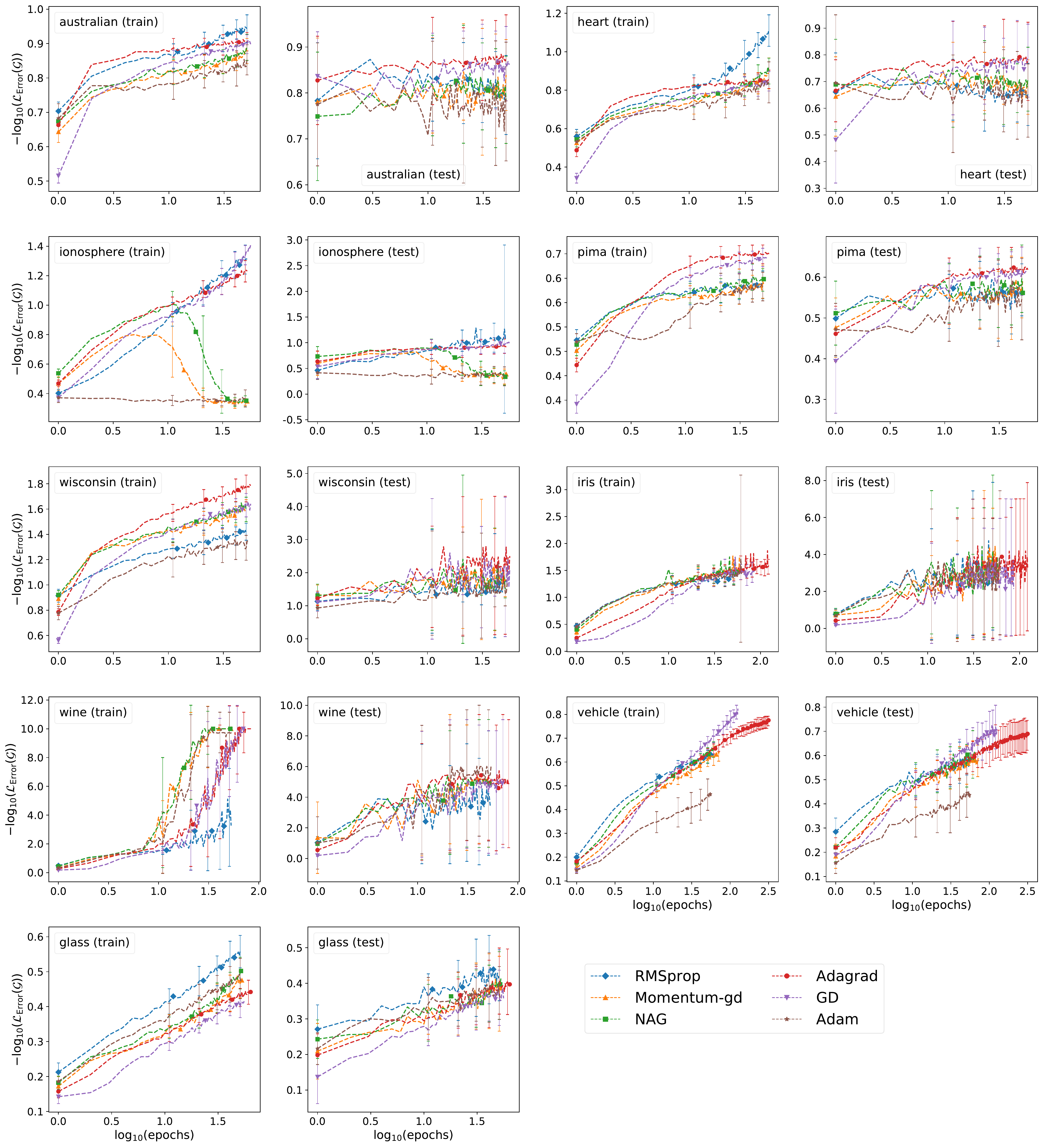}
        %\caption{MLP ReLU P5 Lr 0.1 Class}
        %\label{fig:MLP_class_sigmoid_ES_50_Reg_No}
        
        (a)
        
        \centering
        \includegraphics[width=0.7\linewidth]{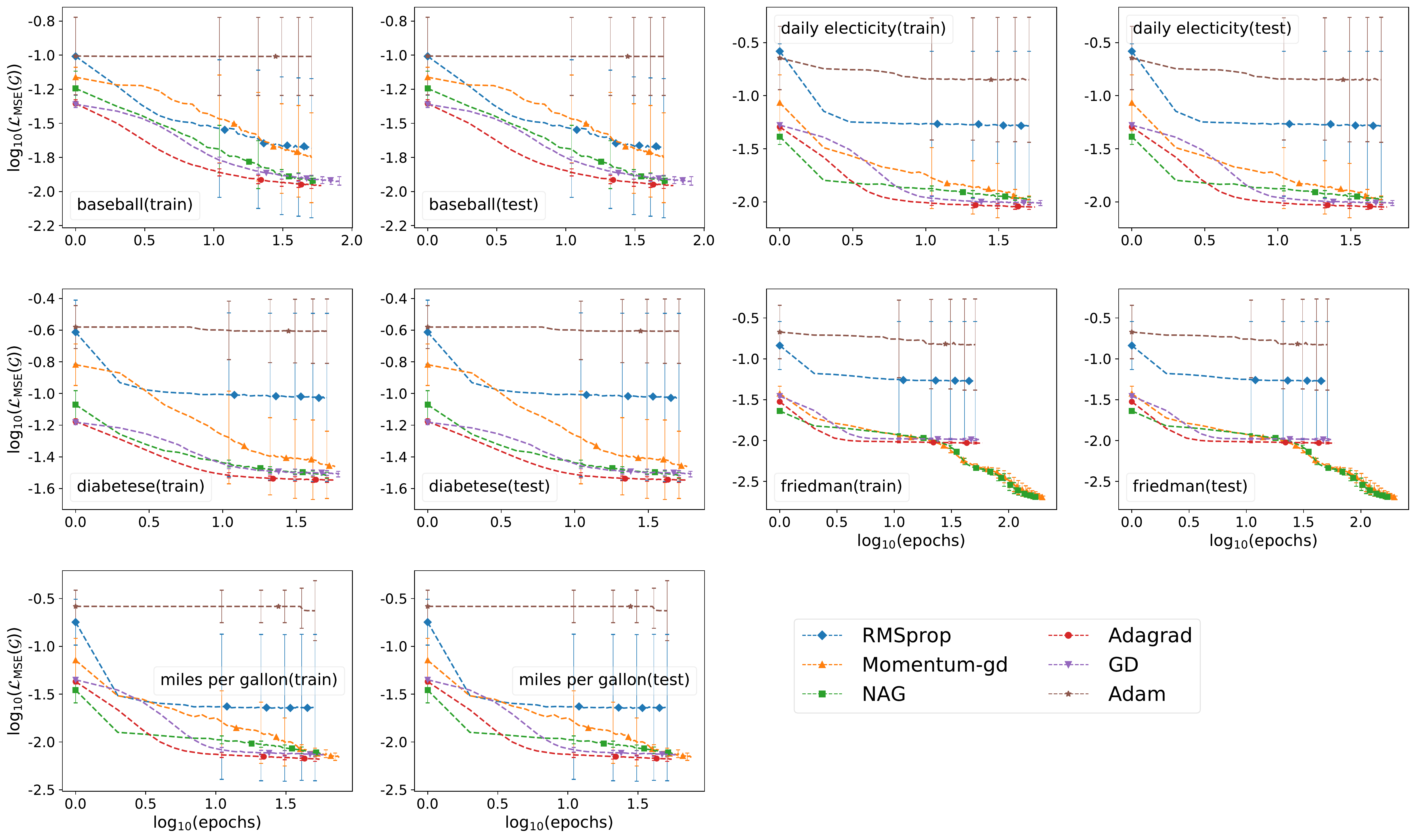}
        
        (b)
        \caption{MLP (a) classification (b) regression with \textit{sigmoid} nodes trained with 0.1 learning rate and \textit{early-stopping} strategy. (a) Adagrad in red in this setting seems taking more iteration as red lines in test sets plots show more fluctuation and longer epochs and lines are on the upper side of plots. (b) NAG is green shows better performance as lines are lower in plots.
        \label{fig:MLP_Sig_ES_50_lr1}}
    \end{figure}

\begin{figure}
        \centering
        \includegraphics[width=0.7\linewidth]{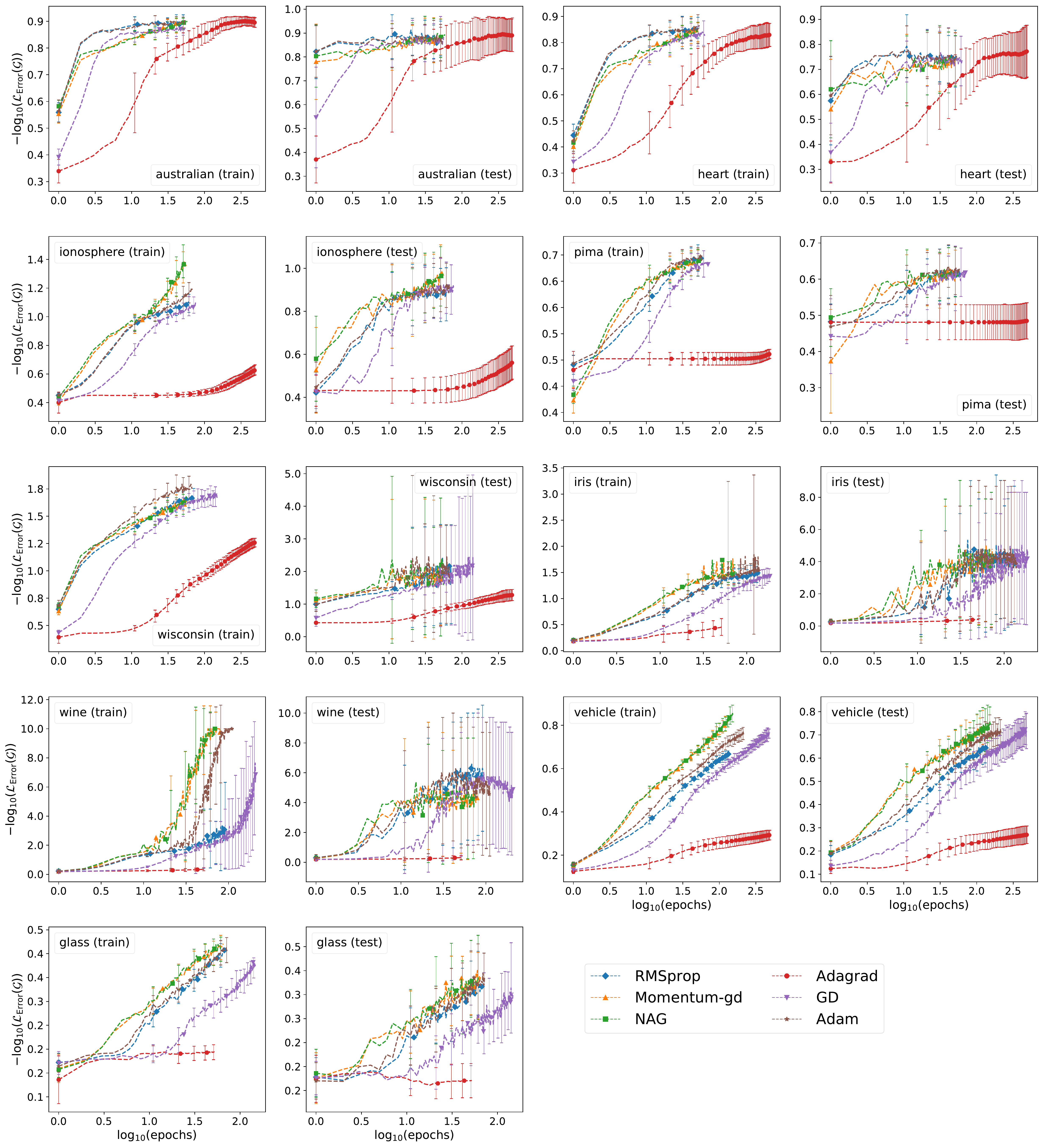}
        %\caption{MLP ReLU P5 Lr 0.1 Class}
        %\label{fig:..}
        
        (a)
        
        \centering
        \includegraphics[width=0.7\linewidth]{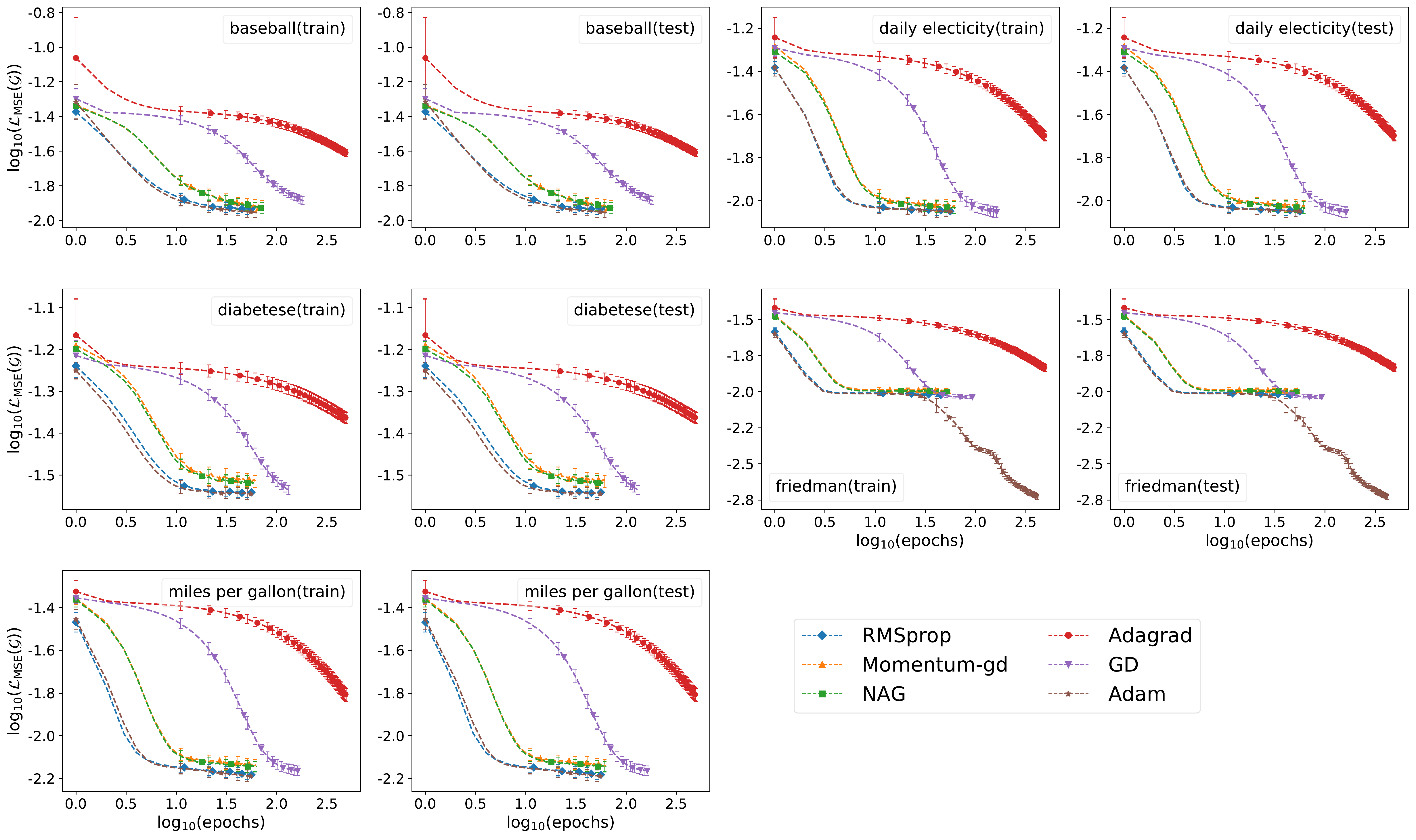}
        
        (b)
        \caption{MLP (a) classification (b) regression with \textit{sigmoid} nodes trained with \textit{default} learning rate  and \textit{early-stopping} strategy. (a) NAG and MGD, in green and orange in this setting, show better performance as on the test sets, and they are on the upper side of plots. Adam is competitively closer to NAG and MGD. (b) Adam, GD, and RMSprop in respective order show better performance as lines are lower in plots, and GD takes longer epochs to converge in early stopping.
        \label{fig:MLP_Sig_ES_50_lrD}}
    \end{figure}

\begin{figure}
        \centering
        \includegraphics[width=0.7\linewidth]{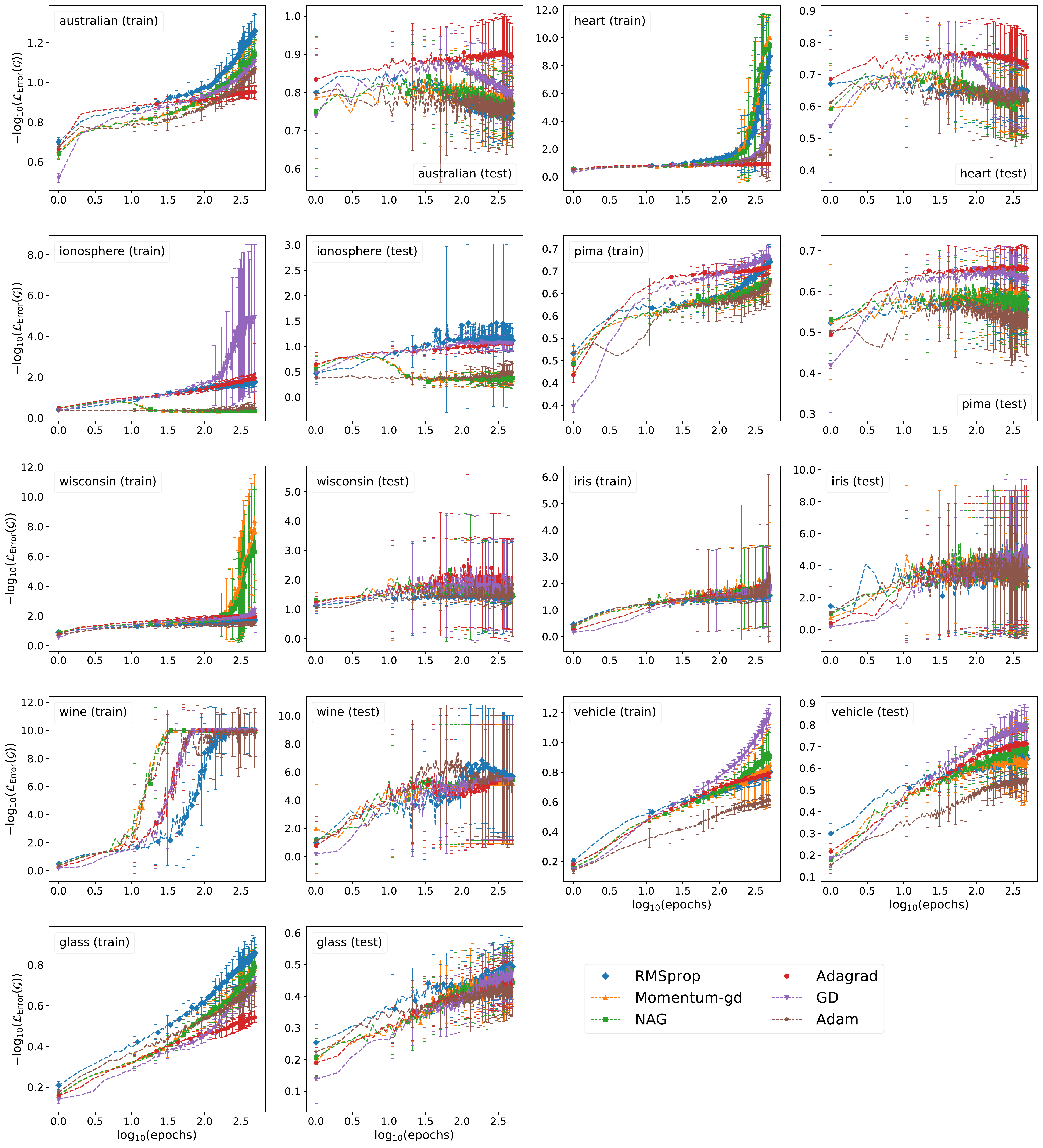}
        \label{fig:..}
        %\caption{MLP ReLU P5 Lr 0.1 Class}
        %\label{fig:..}
        
        (a)
        \centering
        \includegraphics[width=0.7\linewidth]{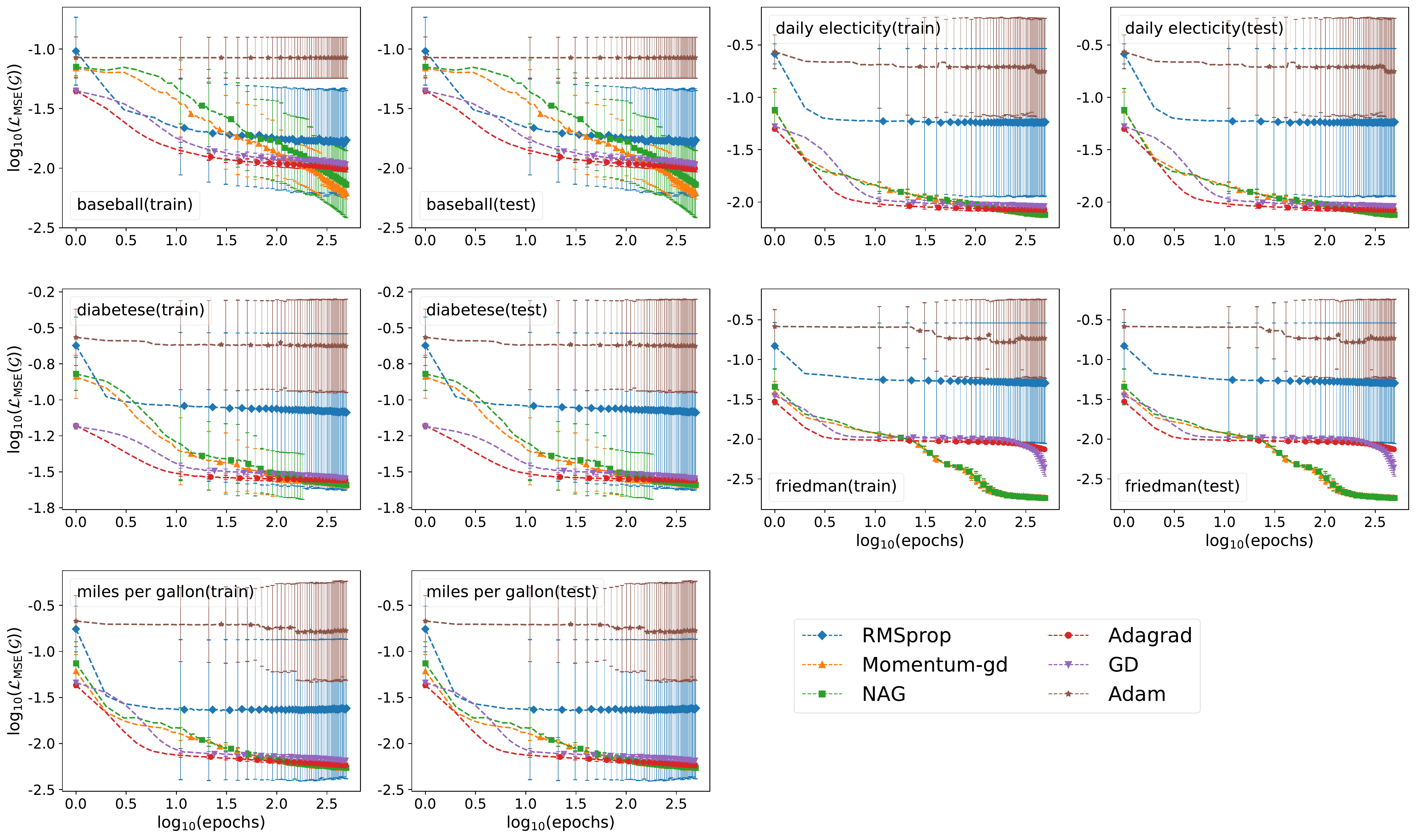}
        
        (b)
        \caption{MLP (a) classification (b) regression with \textit{sigmoid} nodes trained with \textit{0.1} learning rate  and without an \textit{early-stopping} strategy. (a) Adagrad and GD in red and purple in this setting show more fluctuation as lines are on the upper side of plots. (b) GD and NAG in purple and green show better performance as lines are lower in plots.
        \label{fig:MLP_Sig_ES_No_lr1}}
    \end{figure}

\begin{figure}
        \centering
        \includegraphics[width=0.7\linewidth]{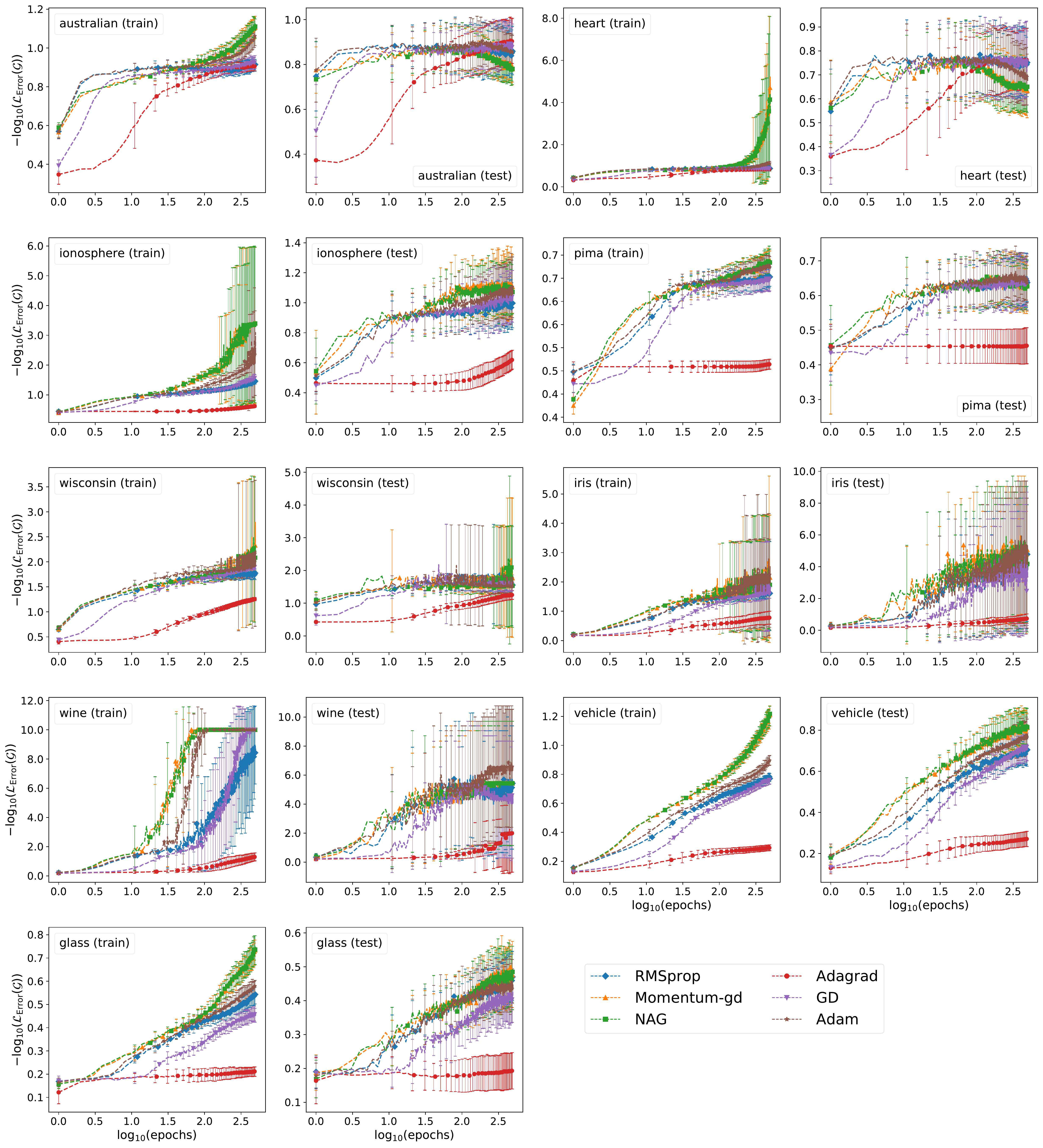}
        %\caption{MLP ReLU P5 Lr 0.1 Class}
        %\label{fig:..}
        
        (a)
        \centering
        \includegraphics[width=0.7\linewidth]{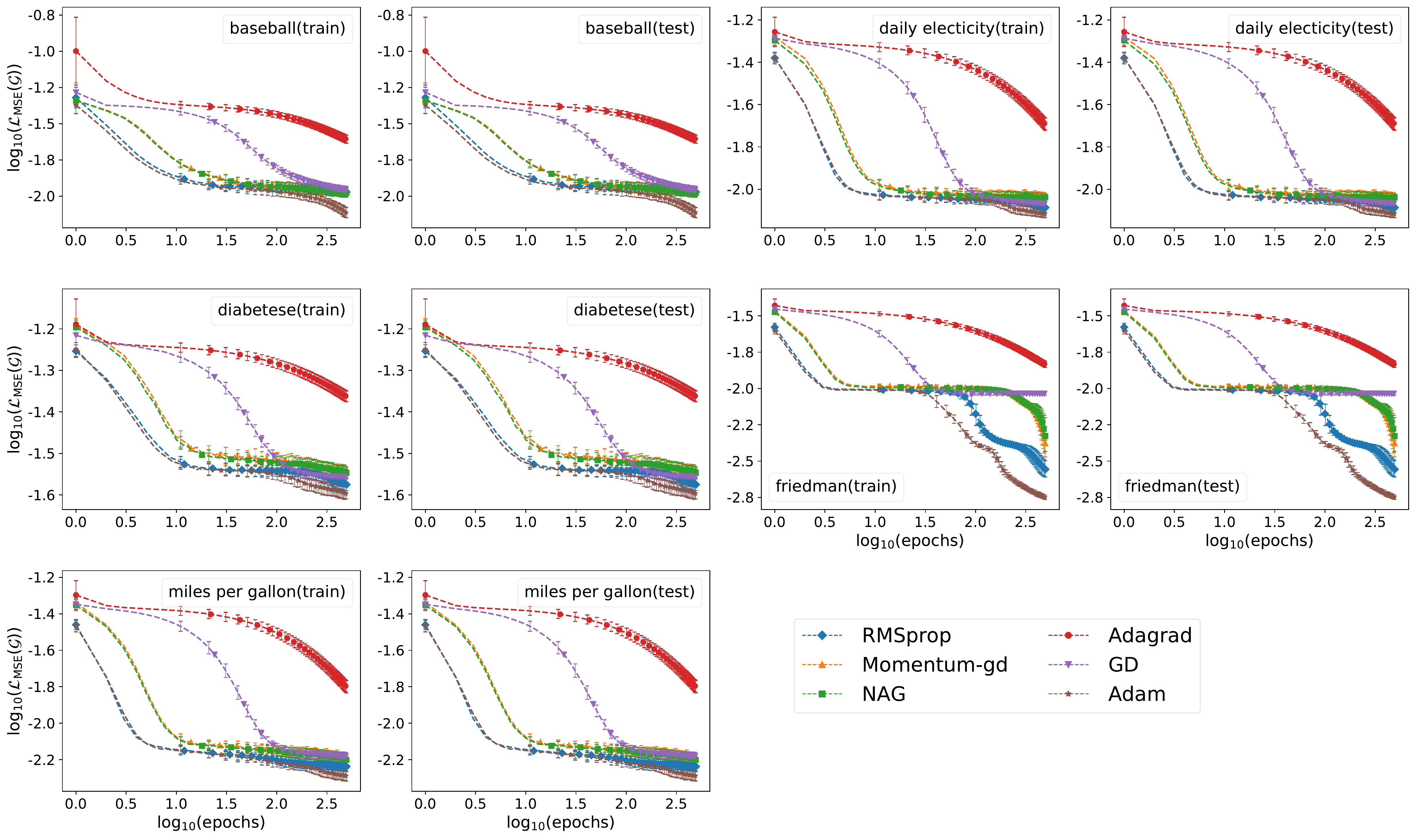}
        
        (b)
        \caption{MLP (a) classification (b) regression with \textit{sigmoid} nodes trained with \textit{default} learning rate  and without an \textit{early-stopping} strategy. (a) Adam and NAG in brown and green in this setting show more fluctuation as lines are on the upper side of plots. (b) Adam, RMSprop, and NAG in brown, blue, and green show better performance as lines are lower in plots.
        \label{fig:MLP_Sig_ES_No_lrD}}
    \end{figure}

    Figs.~\ref{fig:MLP_Sig_ES_50_lr1},
    \ref{fig:MLP_Sig_ES_50_lrD}, 
    \ref{fig:MLP_Sig_ES_No_lr1}, and 
    \ref{fig:MLP_Sig_ES_No_lrD}
    are MLP algorithm settings, each with sigmoid activation functions. However, they  varied as per early-stopping and learning rate usage strategy:
    % (i)
    Fig.~\ref{fig:MLP_Sig_ES_50_lr1} was trained with early-stopping and with a flat $0.1$ learning rate for all optimizers. 
    % (ii)
    Fig.~\ref{fig:MLP_Sig_ES_50_lrD} was trained with early-stopping, but  RMSprop, Adam, and Adagrad had a learning rate of $0.001$, and MGD, NAG, and GD  had a learning rate of $0.01$.
    %(iii)
    Fig.~\ref{fig:MLP_Sig_ES_No_lr1} was trained without early-stopping and with a flat $0.1$ learning rate for all optimizers. 
    % (iv)
    Fig.~\ref{fig:MLP_Sig_ES_No_lrD} was trained without early-stopping, but the optimizers RMSprop, Adam, and Adagrad had a learning rate of $0.001$ and optimizer MGD, NAG, and GD  had a learning rate of $0.01$.
    
    The convergence profiles of the optimizers for MLP were not consistent, with one optimizer outperforming all other optimizers across  all datasets [cf. Figs. \ref{fig:MLP_Sig_ES_50_lr1} (only upto ES epochs),
    \ref{fig:MLP_Sig_ES_50_lrD} (only upto ES epochs), 
    \ref{fig:MLP_Sig_ES_No_lr1} (full epochs), and 
    \ref{fig:MLP_Sig_ES_No_lrD} (full epochs)].  
    RMSprop for learning rate $0.1$ performed better on four datasets. Adagrad performed relatively consistent among all optimizers. Adam performed worse in cases of regression problems with a learning rate of $0.1$. Adam, however, does perform well with a \textit{default} ($0.001$) learning rate. Adagrad shows poor converges with a slower learning rate as it does in the cases of BNeuralT. NAG and MGD, in the cases of both classification and regression problems, show a stable convergence profile.
    
    Considering the convergence profiles of different optimizers on MLP training for early stopping and learning rate $0.1$, we observed the following: Adagrad with early stopping and higher initial learning rate offered better accuracy on both classification and regression problems. This is because Adagrad's small step at higher epochs allowed networks to find a better early stopping point than that of the other algorithms whose larger step size made networks converge to a premature early stopping point (cf. Figs.~\ref{fig:MLP_Sig_ES_50_lr1} and \ref{fig:MLP_Sig_ES_No_lr1}).
    
    With an initial small learning rate of $0.001$, Adagrad is too slow to converge within a predefined number of epochs ($500$). In this setting, Adam seems to have an  appropriate small step size to converge to a proper early-stopping point. The performances of RMSprop, MGD, NAG, and GD are next to Adam's performance (cf. Figs.~\ref{fig:MLP_Sig_ES_50_lrD} and \ref{fig:MLP_Sig_ES_No_lrD}).

    %%%     MLP  ES 50 lr 0.1
    
    %TODO %%     MLP  ES 50 lr defalult
    
    %TODO %%     MLP  ES No lr 0.1
    
    %TODO %%     MLP  ES NO lr default
    
    %TODO  %%%%%%%%%%%%%%%%%%%%%%%%%%%%%%%%
    \newpage
    \subsection{Supplementary Tables}
    \label{sec:results_table_other}
    \begin{table}[H]
        \centering
        \footnotesize 		
        \caption{Hyperparameter definition and choices used during the experiments of the algorithms. BNeuralT and MLP parameters setting are listed in detail. Some of the  basic settings as recommended in the library or literature of HFNT, DT/RF, GP, NCB, and SVM are listed in this table.}
        \label{tab:parameters}
        \renewcommand{\arraystretch}{1.2}
        \setlength{\tabcolsep}{0.1cm}
        \begin{tabular}{llp{6cm}p{3.5cm}l}
            \toprule
            & Parameter & Definition & Default Rang &  Choices used \\
            \midrule
            \parbox[t]{3mm}{\multirow{6}{*}{\rotatebox[origin=c]{90}{BNeuralT}}}
            & Scaling & Input-features scaling range. & $[min,max]$, & $[0,1]$ \\	
            & Tree height & Maximum depth (layers) of a tree model. & $ \left\lbrace p \in  \mathbb{Z} > 1 \right\rbrace $  & $\{5,10\}$ \\			
            & Tree arity  & Maximum arguments of a node $ m $. & $ \left\lbrace m \in  \mathbb{Z} \ge 2 \right\rbrace $  & $\{5,10\}$ \\			
            & Tree edge  & Initialization of neural weights of tree & $[w_l,w_u],$ $ w_l\in \mathbb{R}$,  $w_u \in \mathbb{R}$  & $[0,1]$ \\			
            & $P[\text{leaf}_p <p]$ & Leaf generation at depth $< p$ & $P[\text{leaf}_p <p] \in [0,1]$ & $\{0.4, 0.5\}$\\
            & Internal nodes  & An activation function & \parbox[t]{3.5cm}{\raggedright \{Sigmoid, ReLU, tanh\} } &\parbox[t]{2.5cm}{\raggedright \{Sigmoid, ReLU\} }\\
            \cmidrule{1-1}
            \parbox[t]{3mm}{\multirow{3}{*}{\rotatebox[origin=c]{90}{MLP}}}	
            & Layers  & Number of hidden layers in  architecture &  & 1\\
            & Hidden nodes  & Number of nodes at hidden layer &  & 100\\
            & Activation nodes  & An activation function & \parbox[t]{3.5cm}{\raggedright \{Sigmoid, ReLU, tanh\}}& \parbox[t]{2.5cm}{\raggedright \{Sigmoid, ReLU\} }\ \\
            \cmidrule{1-1}
            \parbox[t]{3mm}{\multirow{6}{*}{\rotatebox[origin=c]{90}{Optimizers}}}
            & Learning rate & Optimizer's learning rate & $\eta \in [0,1]$ & $\{0.1,0.01,0.001\}$\\
            %&                      & GD                   & $\eta \in [0,1]$ & $\{0.1,0.01\}$\\
            %&                      & MGD & $\eta \in [0,1]$ & $\{0.1,0.01\}$\\
            %&                      & NAG                 & $\eta \in [0,1]$ & $\{0.1,0.01\}$\\
            %&                      & Adagrad           & $\eta \in [0,1]$ & $\{0.1,0.001\}$\\
            %&                      & RMSprop          & $\eta \in [0,1]$ & $\{0.1,0.001\}$\\
            %&                      & Adam               & $\eta \in [0,1]$ & $\{0.1,0.001\}$\\
            & Momentum rate & Optimizers momentum rate & $\gamma \in [0,1]$ & $\{0.9\}$\\
            & Training epochs  & Condition for algorithm termination & $ \left\lbrace Epochs \in  \mathbb{Z} > 1 \right\rbrace $ &  $ 500 $ \\
            & Early-stopping & Stopping algorithm to avoid overfitting & $[0, Epochs]$ & $\{50\}$\\
            & 					   & Method of early-stopping & Fallback $\in$ [True, False] & True\\
            & Algorithms & Algorithms  for optimization & \multicolumn{2}{l}{\parbox[t]{6cm}{\raggedright \{GD, MGD, NAG, Adagrad, RMSprop, Adam, L-BFGS~\citep{liu1989limited}\}} } \\
            \cmidrule{1-1}
            \parbox[t]{3mm}{\multirow{4}{*}{\rotatebox[origin=c]{90}{HFNT}}}
            & Tree setup & \multicolumn{3}{l}{Height = 10, arity = 5, edge = $[0,1]$, leaf generation is attribute dependent} \\			
            & Internal node & \multicolumn{3}{l}{Takes arbitrarily any function from
            \{Gaussian, Sigmoid, tanh, fermi, etc.\}} \\
            & Structure & \multicolumn{3}{l}{Single and multi objective genetic programming with 50 population and 500 generations} \\
            & Parameter  & \multicolumn{3}{l}{Differential evaluation with 100 population and 1000 generations, 0.9 crossover rate} \\
            \cmidrule{1-1}
            \parbox[t]{3mm}{\multirow{6}{*}{\rotatebox[origin=c]{90}{DT/RF}}}	
            & Tree depth  & Restriction max depth of the tree & $ \left\lbrace p \in  \mathbb{Z} > 1 \right\rbrace $  & Unbounded \\
            & Max leaf node & Restriction on max leaf node of a tree & & Unbounded\\
            & Split & Strategy for splitting a decision node & & ``best'' \\
            & Criteria  & Decision criteria about an input &  Classification & ``gini index'' \\
            &              &   &  Regression & MSE \\
            & Random forest & Data sampling method for tree generation &  & bootstrap \\ 
            & Estimators & The number of trees in the forest & & 100\\
            \cmidrule{1-1}
            \parbox[t]{3mm}{\multirow{5}{*}{\rotatebox[origin=c]{90}{GP}}}	
            & Optimizer  & Algorithm used for kernel optimization  &  &  L-BFGS \\
            & Training iteration  & Condition for algorithm termination & $ \left\lbrace Epochs \in  \mathbb{Z} > 1 \right\rbrace $ &  $ 100 $ \\
            & Classification & Method of multi-class classification &\parbox[t]{3cm}{\raggedright \{1 vs rest, 1 vs 1\}} & \{1 vs rest\} \\
            & Kernel & Gaussian kernel used for learning & Classification & radial basis \\
            &              &                                                & Regression & \parbox[t]{2.5cm}{\raggedright dot product + white noise}    \\
            \cmidrule{1-1}
            \parbox[t]{3mm}{\multirow{5}{*}{\rotatebox[origin=c]{90}{SVM}}}	
            & Kernel & Kernel used for learning &  & radial basis  \\
            & Classification & Method of multi-class classification &\parbox[t]{3cm}{\raggedright \{1 vs rest, 1 vs 1 \}} & \{1 vs rest\} \\
            & Penalty & Regularization to avoid overfitting &\{L1, L2\} & L1\\
            & Training loss & Loss function used during training & & \parbox[t]{2.cm}{\raggedright hinge loss } \\
            & Training iteration  & Condition for algorithm termination & $ \left\lbrace Epochs \in  \mathbb{Z} > 1 \right\rbrace $ &  $ 1000 $ \\
            \cmidrule{1-1}
            \multicolumn{2}{l}{NBC}	& It uses Gaussian function & &  \\
            \bottomrule
        \end{tabular}
    \end{table}

\begin{table}[]
    \tiny
    \centering 
    \renewcommand{\arraystretch}{1.2}
    \setlength{\tabcolsep}{5pt}
    \caption{Summary of top (based on overall performance) 55 of 110 experiments on all data, on all algorithms, and all settings . The values are an average of 30 independent runs of each setting. Names of experiments start with B indicate BNeuralT and M indicate MLP. The next letter is S or R indicates sigmoid and ReLU activation function, ESy indicates early-stopping and ESn indicates no early-stopping, Rn or Ry indicates no regularization or elastic net regularization; L indicates default learning rate or absence of L indicates a learning rate of $0.1$, Dr indicates dropout; p4 or p5 indicates leaf generation rate $0.4$ or $0.5$, and optimizers GD, MGD, NAG, Adagrad, RMSprop, and Adam are  indicated with G, M, N, A, R, D, respectively.
    \label{tab:All_experiments_A}}
    \begin{tabular}[t]{lrrrlrrrlrrr}
        \toprule
        Overall & $Acc$ & $\sigma_{Acc}$ & $\textbf{w}$ & Classification & $r_2$ & $\sigma$ & $\textbf{w}$ & Regression & $Acc$ & $\sigma_{Acc}$ & $\textbf{w}$ \\
        \midrule
        \textbf{B-S-ESy-p4-R} & .832 & .15 & 222 & \textbf{B-S-ESy-p4-R} & .891 & .10 & 261 & M-S-ESy-Rn-N & .775 & .16 & 1041  \\
        M-S-ESy-Rn-N & .832 & .14 & 1638 & \textbf{B-S-ESy-p5-R} & .888 & .11 & 157 & M-S-ESy-Rn-L-D & .772 & .16 & 1041  \\
        M-S-ESy-Rn-L-D & .831 & .14 & 1638 & \textbf{B-S-ESy-p4-N} & .875 & .11 & 261 & M-R-ESn-Rn-L-G & .761 & .17 & 1041  \\
        M-S-ESy-Rn-G & .823 & .14 & 1638 & \textbf{B-S-ESy-p4-M} & .874 & .11 & 261 & M-S-ESy-Rn-M & .759 & .22 & 1041  \\
        M-S-ESy-Rn-M & .822 & .17 & 1638 & \textbf{B-S-ESy-p5-N} & .871 & .11 & 157 & M-S-ESn-Rn-L-Dr-D & .754 & .15 & 1041  \\
        \textit{M-S-ESn-Rn-L-Dr-D} & .821 & .14 & 1638 & \textbf{B-S-ESy-p5-M} & .871 & .11 & 157 & M-S-ESy-Rn-L-R & .752 & .16 & 1041  \\
        M-S-ESy-Rn-L-M & .821 & .14 & 1638 & GP & .868 & .10 &  & M-S-ESn-Rn-L-D & .749 & .19 & 1041  \\
        M-S-ESy-Rn-L-N & .820 & .14 & 1638 & M-S-ESy-Rn-L-M & .868 & .12 & 1970 & M-S-ESn-Rn-L-R & .745 & .17 & 1041  \\
        RF & .818 & .14 &  & M-S-ESy-Rn-L-N & .868 & .11 & 1970 & M-S-ESy-Rn-G & .744 & .14 & 1041  \\
        M-S-ESy-Rn-L-R & .818 & .14 & 1638 & M-S-ESy-Rn-G & .867 & .12 & 1970 & M-S-ESy-Rn-L-M & .737 & .14 & 1041  \\
        M-S-ESn-Rn-L-D & .817 & .15 & 1638 & M-S-ESy-Rn-L-Dr-N & .865 & .12 & 1970 & M-S-ESn-Rn-L-Dr-R & .737 & .15 & 1041  \\
        M-S-ESn-Rn-L-R & .815 & .14 & 1638 & M-S-ESy-Rn-N & .865 & .12 & 1970 & M-S-ESy-Rn-A & .736 & .14 & 1041  \\
        M-S-ESy-Rn-A & .814 & .14 & 1638 & RF & .864 & .10 &  & M-S-ESn-Rn-G & .735 & .16 & 1041  \\
        M-S-ESy-Rn-L-Dr-D & .813 & .14 & 1638 & M-S-ESy-Rn-L-D & .863 & .12 & 1970 & RF & .735 & .17 &   \\
        B-S-ESy-p5-R & .812 & .18 & 131 & M-S-ESy-Rn-L-Dr-M & .863 & .13 & 1970 & M-S-ESy-Rn-L-N & .733 & .14 & 1041  \\
        M-S-ESn-Rn-G & .812 & .14 & 1638 & M-S-ESn-Rn-L-Dr-M & .861 & .11 & 1970 & M-S-ESn-Rn-M & .733 & .20 & 1041  \\
        M-S-ESn-Rn-A & .811 & .14 & 1638 & M-S-ESy-Rn-L-Dr-D & .861 & .12 & 1970 & SVM & .733 & .20 &   \\
        B-S-ESy-p4-N & .808 & .17 & 222 & M-S-ESn-Rn-L-Dr-D & .859 & .11 & 1970 & \textbf{B-S-ESy-p4-R} & .727 & .17 & 152  \\
        M-S-ESn-Rn-L-Dr-R & .808 & .14 & 1638 & M-S-ESn-Rn-Dr-G & .860 & .11 & 1970 & M-S-ESn-Rn-A & .727 & .14 & 1041  \\
        M-S-ESy-Rn-L-Dr-N & .808 & .15 & 1638 & M-S-ESn-Rn-L-Dr-N & .858 & .11 & 1970 & M-S-ESy-Rn-L-Dr-D & .725 & .14 & 1041  \\
        M-S-ESn-Rn-L-N & .808 & .15 & 1638 & M-R-ESn-Rn-A & .858 & .11 & 1970 & M-S-ESn-Rn-L-M & .725 & .17 & 1041  \\
        M-S-ESn-Rn-Dr-A & .807 & .14 & 1638 & M-S-ESy-Rn-Dr-G & .858 & .13 & 1970 & M-S-ESn-Rn-L-N & .724 & .17 & 1041  \\
        B-S-ESy-p4-M & .807 & .17 & 222 & M-S-ESn-Rn-A & .857 & .11 & 1970 & M-S-ESn-Rn-Dr-N & .721 & .16 & 1041  \\
        M-S-ESn-Rn-L-Dr-M & .807 & .14 & 1638 & M-S-ESy-Rn-A & .857 & .12 & 1970 & M-S-ESn-Rn-Dr-A & .719 & .14 & 1041  \\
        M-S-ESn-Rn-L-M & .807 & .15 & 1638 & M-S-ESy-Rn-M & .857 & .13 & 1970 & M-S-ESn-Rn-N & .717 & .26 & 1041  \\
        M-S-ESy-Rn-L-Dr-M & .807 & .15 & 1638 & M-S-ESn-Rn-Dr-A & .857 & .12 & 1970 & M-S-ESy-Rn-L-Dr-R & .717 & .14 & 1041  \\
        GP & .806 & .14 &  & M-S-ESn-Rn-L-D & .856 & .11 & 1970 & M-S-ESy-Rn-L-G & .716 & .14 & 1041  \\
        M-S-ESn-Rn-L-Dr-N & .805 & .14 & 1638 & M-S-ESn-Rn-G & .854 & .11 & 1970 & M-S-ESy-Rn-Dr-A & .715 & .14 & 1041  \\
        M-S-ESy-Rn-L-G & .805 & .14 & 1638 & M-S-ESy-Rn-Dr-A & .854 & .12 & 1970 & M-S-ESn-Rn-L-Dr-N & .710 & .14 & 1041  \\
        M-S-ESy-Rn-Dr-G & .805 & .15 & 1638 & M-S-ESn-Rn-L-N & .854 & .11 & 1970 & M-S-ESn-Rn-L-Dr-M & .709 & .14 & 1041  \\
        M-S-ESy-Rn-Dr-A & .804 & .14 & 1638 & M-S-ESn-Rn-L-R & .854 & .11 & 1970 & B-S-ESy-p4-A & .708 & .19 & 152  \\
        M-S-ESn-Rn-Dr-G & .804 & .15 & 1638 & M-S-ESy-Rn-L-G & .854 & .12 & 1970 & M-S-ESy-Rn-Dr-G & .708 & .14 & 1041  \\
        SVM & .802 & .16 &  & M-S-ESy-Rn-L-R & .854 & .12 & 1970 & M-S-ESn-Rn-L-G & .707 & .14 & 1041  \\
        M-S-ESy-Rn-L-Dr-R & .801 & .14 & 1638 & M-S-ESn-Rn-L-M & .852 & .11 & 1970 & M-S-ESy-Rn-L-Dr-M & .705 & .14 & 1041  \\
        M-R-ESn-Rn-A & .801 & .19 & 1638 & B-S-ESy-p4-A & .852 & .13 & 261 & M-R-ESn-Rn-G & .704 & .24 & 1041  \\
        B-S-ESy-p4-A & .801 & .17 & 222 & M-S-ESn-Rn-L-G & .852 & .12 & 1970 & M-S-ESn-Rn-Dr-G & .704 & .14 & 1041  \\
        M-S-ESn-Rn-L-G & .800 & .15 & 1638 & M-S-ESy-Rn-R & .850 & .15 & 1970 & M-R-ESn-Rn-L-M & .704 & .25 & 1041  \\
        M-R-ESn-Rn-L-G & .799 & .14 & 1569 & M-R-ESn-Rn-G & .850 & .11 & 1970 & M-S-ESy-Rn-L-Dr-N & .703 & .14 & 1041  \\
        M-R-ESn-Rn-G & .798 & .18 & 1638 & M-S-ESy-Rn-L-Dr-R & .848 & .13 & 1970 & M-R-ESn-Rn-A & .698 & .25 & 1041  \\
        B-S-ESy-p5-N & .790 & .19 & 131 & M-S-ESn-Rn-L-Dr-R & .848 & .12 & 1970 & M-R-ESn-Rn-L-N & .698 & .26 & 1041  \\
        B-S-ESy-p5-M & .789 & .19 & 131 & B-S-ESy-p4-D & .843 & .14 & 261 & GP & .695 & .13 &   \\
        NBC & .782 &  &  & M-S-ESn-Rn-L-Dr-G & .842 & .13 & 1970 & B-S-ESy-p5-L-R & .688 & .19 & 95  \\
        M-S-ESn-Rn-L-Dr-G & .779 & .16 & 1638 & M-S-ESy-Rn-L-Dr-G & .842 & .14 & 1970 & B-S-ESy-p4-M & .687 & .20 & 152  \\
        B-S-ESy-p5-A & .778 & .19 & 131 & SVM & .841 & .13 &  & M-R-ESn-Rn-L-A & .686 & .16 & 1041  \\
        M-S-ESy-Rn-L-Dr-G & .777 & .17 & 1638 & B-S-ESy-p5-A & .841 & .13 & 157 & B-S-ESy-p4-N & .686 & .20 & 152  \\
        B-S-ESy-p5-L-R & .776 & .17 & 138 & M-S-ESn-Rn-Dr-R & .841 & .12 & 1970 & B-S-ESy-p5-R & .676 & .21 & 85  \\
        M-S-ESn-Rn-M & .770 & .18 & 1638 & B-S-ESy-p5-D & .840 & .14 & 157 & M-R-ESn-Rn-L-R & .675 & .26 & 1041  \\
        M-R-ESn-Rn-L-M & .769 & .19 & 1569 & M-S-ESn-Rn-R & .838 & .12 & 1970 & B-S-ESy-p5-L-D & .673 & .21 & 95  \\
        M-S-ESn-Rn-N & .769 & .20 & 1638 & B-S-ESy-p4-G & .832 & .14 & 261 & B-S-ESn-A & .668 & .21 & 80  \\
        M-R-ESn-Rn-L-N & .767 & .19 & 1569 & M-S-ESy-Rn-Dr-R & .832 & .14 & 1970 & M-S-ESn-Rn-Dr-M & .667 & .39 & 1041  \\
        B-S-ESn-A & .765 & .19 & 128 & B-S-ESn-R & .831 & .12 & 154 & M-S-ESn-Rn-L-Dr-G & .665 & .14 & 1041  \\
        B-S-ESn-R & .765 & .18 & 128 & B-S-ESy-p5-G & .830 & .14 & 157 & B-S-ESy-p5-A & .664 & .22 & 85  \\
        B-S-ESy-p4-D & .765 & .20 & 222 & B-S-ESy-p5-L-N & .829 & .14 & 162 & M-S-ESy-Rn-L-Dr-G & .660 & .15 & 1041  \\
        M-R-ESn-Rn-L-R & .762 & .20 & 1569 & B-S-ESy-p5-L-M & .828 & .14 & 162 & M-R-ESn-Rn-M & .656 & .29 & 1041  \\
        B-S-ESn-N & .760 & .19 & 128 & M-R-ESn-Rn-L-G & .826 & .11 & 1946 & B-S-ESn-N & .651 & .22 & 80  \\
\bottomrule
    \end{tabular}
\end{table}     

\begin{table}[]
    \tiny
    \centering 
    \renewcommand{\arraystretch}{1.2}
    \setlength{\tabcolsep}{5pt}
    \caption{Summary of last (based on overall performance) 55 of 110 experiments on all data, on all algorithms, and all settings . The values are an average of 30 independent runs of each setting. Names of experiments start with B indicate BNeuralT and M indicate MLP. The next letter is S or R indicates sigmoid and ReLU activation function, ESy indicates early-stopping and ESn indicates no early-stopping, Rn or Ry indicates no regularization or elastic net regularization; L indicates default learning rate or absence of L indicates a learning rate of $0.1$, Dr indicates dropout; p4 or p5 indicates leaf generation rate $0.4$ or $0.5$, and optimizers GD, MGD, NAG, Adagrad, RMSprop, and Adam are  indicated with G, M, N, A, R, D, respectively.
    \label{tab:All_experiments_B}}
    \begin{tabular}[t]{lrrrlrrrlrrr}
        \toprule
        Overall & $Acc$ & $\sigma_{Acc}$ & $\textbf{w}$ & Classification & $r_2$ & $\sigma$ & $\textbf{w}$ & Regression & $Acc$ & $\sigma_{Acc}$ & $\textbf{w}$ \\
        \midrule
        B-S-ESn-M & .759 & .19 & 128 & B-S-ESy-p5-L-R & .825 & .15 & 162 & M-R-ESn-Rn-N & .650 & .30 & 1041  \\
        B-S-ESy-p5-L-D & .751 & .19 & 138 & M-R-ESn-Rn-L-R & .824 & .10 & 1946 & B-S-ESn-M & .649 & .22 & 80  \\
        B-S-ESy-p4-G & .750 & .21 & 222 & M-S-ESy-Rn-D & .821 & .15 & 1970 & B-S-ESn-R & .646 & .21 & 80  \\
        B-S-ESy-p5-D & .747 & .21 & 131 & B-S-ESn-N & .821 & .13 & 154 & B-S-ESy-p5-N & .644 & .22 & 85  \\
        M-R-ESn-Rn-L-D & .747 & .24 & 1569 & M-R-ESn-Rn-L-D & .820 & .11 & 1946 & M-R-ESn-Rn-L-D & .644 & .32 & 1041  \\
        M-S-ESn-Rn-Dr-N & .744 & .17 & 1638 & B-S-ESn-M & .820 & .13 & 154 & B-S-ESy-p5-M & .643 & .22 & 85  \\
        B-S-ESy-p5-L-N & .740 & .22 & 138 & B-S-ESn-A & .819 & .14 & 154 & M-S-ESy-Rn-Dr-N & .628 & .64 & 1041  \\
        B-S-ESy-p5-L-M & .740 & .22 & 138 & M-R-ESn-Rn-L-N & .817 & .11 & 1946 & B-S-ESy-p4-D & .623 & .21 & 152  \\
        B-S-ESy-p5-G & .737 & .22 & 131 & B-R-ESy-p5-R & .817 & .14 & 150 & B-S-ESy-p4-G & .602 & .23 & 152  \\
        M-S-ESy-Rn-Dr-N & .733 & .41 & 1638 & M-R-ESn-Rn-L-M & .816 & .11 & 1946 & B-S-ESn-G & .582 & .24 & 80  \\
        M-R-ESn-Rn-L-A & .732 & .16 & 1569 & DT & .813 & .11 &  & B-S-ESy-p5-L-M & .580 & .24 & 95  \\
        B-S-ESn-G & .720 & .21 & 128 & B-R-ESy-p5-G & .802 & .18 & 150 & B-S-ESy-p5-L-N & .580 & .24 & 95  \\
        M-S-ESn-Rn-Dr-M & .714 & .27 & 1638 & M-S-ESn-Rn-N & .798 & .15 & 1970 & B-S-ESy-p5-D & .579 & .21 & 85  \\
        HFNT$^{\text{S}}$ & .708 & .32 &  & B-S-ESn-G & .797 & .14 & 154 & B-S-ESy-p5-G & .570 & .24 & 85  \\
        B-R-ESy-p5-A & .706 & .22 & 125 & B-R-ESy-p5-A & .795 & .16 & 150 & HFNT$^{\text{M}}$ & .567 & .54 &   \\
        B-S-ESn-D & .705 & .21 & 128 & B-S-ESy-p5-L-D & .795 & .16 & 162 & HFNT$^{\text{S}}$ & .562 & .44 &   \\
        HFNT$^{\text{M}}$ & .704 & .37 &  & M-S-ESn-Rn-D & .792 & .15 & 1970 & B-S-ESn-D & .557 & .23 & 80  \\
        DT & .695 & .27 &  & M-S-ESy-Rn-Dr-N & .791 & .17 & 1970 & B-R-ESy-p5-A & .544 & .24 & 81  \\
        B-R-ESy-p5-R & .689 & .25 & 125 & M-S-ESy-Rn-Dr-D & .791 & .17 & 1970 & B-R-ESy-p5-N & .484 & .25 & 81  \\
        B-R-ESy-p5-G & .683 & .32 & 125 & M-S-ESn-Rn-M & .791 & .17 & 1970 & DT & .484 & .34 &   \\
        B-R-ESy-p5-N & .677 & .24 & 125 & M-S-ESy-Rn-Dr-M & .789 & .16 & 1970 & B-R-ESy-p5-M & .483 & .25 & 81  \\
        B-R-ESy-p5-M & .676 & .24 & 125 & HFNT$^{\text{S}}$ & .789 & .18 &  & M-S-ESy-Rn-L-A & .482 & .16 & 1041  \\
        M-S-ESn-Rn-L-A & .639 & .22 & 1638 & B-S-ESn-D & .788 & .15 & 154 & M-S-ESn-Rn-L-A & .479 & .16 & 1041  \\
        M-S-ESy-Rn-L-A & .635 & .22 & 1638 & B-R-ESy-p5-N & .784 & .16 & 150 & B-R-ESy-p5-G & .470 & .39 & 81  \\
        M-S-ESy-Rn-Dr-M & .622 & .75 & 1638 & B-R-ESy-p5-M & .782 & .15 & 150 & B-R-ESy-p5-R & .457 & .23 & 81  \\
        B-S-ESy-p5-L-G & .602 & .26 & 138 & NBC & .782 &  &  & B-S-ESy-p5-L-G & .394 & .26 & 95  \\
        B-R-ESy-p5-D & .583 & .28 & 125 & HFNT$^{\text{M}}$ & .779 & .18 &  & B-R-ESy-p5-D & .358 & .28 & 81  \\
        M-S-ESn-Rn-L-Dr-A & .569 & .24 & 1638 & M-S-ESn-Rn-Dr-D & .773 & .18 & 1970 & M-S-ESn-Rn-L-Dr-A & .342 & .12 & 1041  \\
        M-S-ESy-Rn-L-Dr-A & .556 & .25 & 1638 & M-R-ESn-Rn-L-A & .764 & .15 & 1946 & M-S-ESy-Rn-Dr-M & .320 & 1.18 & 1041  \\
        M-R-ESn-Rn-M & .543 & .23 & 1638 & M-R-ESn-Rn-R & .760 & .23 & 1970 & M-S-ESy-Rn-L-Dr-A & .320 & .15 & 1041  \\
        M-R-ESn-Rn-N & .527 & .24 & 1638 & M-S-ESn-Rn-Dr-N & .757 & .17 & 1970 & M-S-ESn-Ry-L-A & -0.016 & .02 & 1041  \\
        M-S-ESn-Ry-L-G & .468 & .42 & 1638 & M-S-ESn-Ry-L-G & .743 & .25 & 1970 & M-S-ESn-Ry-A & -0.019 & .03 & 1041  \\
        M-S-ESn-Ry-L-D & .442 & .40 & 1638 & M-S-ESn-Rn-Dr-M & .739 & .17 & 1970 & M-S-ESn-Ry-L-D & -0.025 & .04 & 1041  \\
        M-S-ESn-Ry-A & .431 & .39 & 1638 & M-S-ESn-Rn-L-A & .728 & .20 & 1970 & M-S-ESn-Ry-L-G & -0.027 & .04 & 1041  \\
        M-S-ESn-Ry-L-N & .429 & .43 & 1638 & M-S-ESy-Rn-L-A & .719 & .20 & 1970 & M-S-ESn-Ry-L-R & -0.040 & .05 & 1041  \\
        M-S-ESn-Ry-L-M & .429 & .41 & 1638 & B-S-ESy-p5-L-G & .718 & .18 & 162 & M-S-ESn-Ry-L-M & -0.058 & .08 & 1041  \\
        M-S-ESn-Ry-L-R & .426 & .41 & 1638 & M-S-ESn-Ry-L-N & .711 & .24 & 1970 & M-S-ESn-Ry-G & -0.074 & .12 & 1041  \\
        M-S-ESn-Ry-G & .425 & .42 & 1638 & B-R-ESy-p5-D & .707 & .18 & 150 & M-S-ESn-Ry-L-N & -0.077 & .10 & 1041  \\
        M-S-ESn-Ry-N & .337 & .37 & 1638 & M-S-ESn-Ry-G & .702 & .23 & 1970 & M-S-ESn-Ry-N & -0.081 & .12 & 1041  \\
        M-S-ESn-Ry-L-A & .304 & .27 & 1638 & M-S-ESn-Ry-L-D & .701 & .26 & 1970 & M-S-ESn-Ry-M & -0.118 & .28 & 1041  \\
        M-S-ESn-Ry-M & .274 & .37 & 1638 & M-S-ESn-Ry-L-M & .699 & .24 & 1970 & B-S-ESy-p5-L-A & -0.445 & .85 & 95  \\
        M-S-ESy-Rn-Dr-R & .267 & 1.15 & 1638 & M-S-ESn-Rn-L-Dr-A & .695 & .19 & 1970 & M-S-ESy-Rn-Dr-R & -0.748 & 1.44 & 1041  \\
        B-S-ESy-p5-L-A & .178 & .70 & 138 & M-S-ESy-Rn-L-Dr-A & .687 & .20 & 1970 & M-R-ESn-Rn-D & -1.338 & 4.05 & 1041  \\
        M-S-ESn-Rn-Dr-R & -0.027 & 1.82 & 1638 & M-S-ESn-Ry-L-R & .684 & .26 & 1970 & M-S-ESn-Rn-Dr-R & -1.589 & 2.34 & 1041  \\
        M-R-ESn-Rn-D & -0.072 & 2.60 & 1638 & M-S-ESn-Ry-A & .681 & .26 & 1970 & M-S-ESn-Rn-R & -2.096 & 4.00 & 1041  \\
        M-S-ESy-Rn-R & -0.208 & 2.89 & 1638 & M-R-ESn-Rn-D & .631 & .23 & 1970 & M-S-ESy-Rn-R & -2.112 & 4.22 & 1041  \\
        M-S-ESn-Rn-R & -0.210 & 2.77 & 1638 & M-S-ESn-Ry-N & .569 & .23 & 1970 & M-R-ESn-Rn-R & -2.750 & 4.03 & 1041  \\
        M-R-ESn-Rn-R & -0.494 & 2.94 & 1638 & B-S-ESy-p5-L-A & .523 & .18 & 162 & M-S-ESn-Ry-R & -2.829 & 4.03 & 1041  \\
        M-S-ESy-Rn-Dr-D & -0.563 & 2.32 & 1638 & M-S-ESn-Ry-R & .516 & .20 & 1970 & M-S-ESy-Rn-Dr-D & -3.000 & 2.41 & 1041  \\
        M-S-ESn-Ry-R & -0.679 & 2.89 & 1638 & M-S-ESn-Ry-D & .511 & .20 & 1970 & M-S-ESn-Rn-Dr-D & -3.932 & 2.95 & 1041  \\
        M-S-ESn-Rn-Dr-D & -0.908 & 2.86 & 1638 & M-S-ESn-Ry-M & .492 & .20 & 1970 & M-S-ESn-Ry-D & -4.080 & 2.99 & 1041  \\
        M-S-ESn-Rn-D & -0.983 & 3.05 & 1638 & M-S-ESn-Ry-L-A & .481 & .17 & 1970 & M-S-ESn-Rn-D & -4.179 & 3.17 & 1041  \\
        M-S-ESy-Rn-D & -1.014 & 3.28 & 1638 & M-R-ESn-Rn-M & .481 & .16 & 1970 & M-S-ESy-Rn-D & -4.316 & 3.64 & 1041  \\
        M-S-ESn-Ry-D & -1.129 & 2.84 & 1638 & M-R-ESn-Rn-N & .459 & .16 & 1970 & NBC &  &  &   \\
        \bottomrule
    \end{tabular}
\end{table}
    
    \begin{table} %[b]
        \small
        \centering
        \renewcommand{\arraystretch}{1.05}
        \setlength{\tabcolsep}{3pt}
        \caption{Wilcoxon signed-rank test on two samples: BNeuralT's RMSprop against all other algorithms for each data. The `stat,' `pval,' and `post,' respectively indicate Wilcoxon signed-rank statistic, two-tailed $p$-value, and Bonferroni correction post hoc adjusted $p$-value. The values are marked in bold where the null hypothesis that BNeuralT's RMSprop and other algorithms have no difference is rejected.}
        \label{tab:BNeuralT_RMSprop_VS_All_W_Rank_Test}%BNeuralT training is highly competitive with MLP despite having a very low number of parameters compared with MLPs.
        \begin{tabular}[t]{lrrrrrrrrrrrrrrrrr}
            \toprule
            \multicolumn{3}{c}{BNeuralT's} & \multicolumn{9}{c}{Classification} & ~~ & \multicolumn{5}{c}{Regression} \\
            \cline{4-12}\cline{14-18}
            \multicolumn{3}{c}{RMSprop vs.}  & Aus & Hrt & Ion & Pma & Wis & Irs & Win & Vhl & Gls & ~ & Bas & Dee & Dia & Frd & Mpg \\
            \midrule
            \parbox[t]{3mm}{\multirow{18}{*}{\rotatebox[origin=c]{90}{MLP}}}
            & GD & stat & 87.5 & 22 & 21 & 8 & 81 & 7 & 35 & 33 & 18 &  & 199 & 164 & 180 & 90 & 80 \\
            &  & pval & 0 & 0 & 0 & 0 & .01 & 0 & 0 & 0 & 0 &  & .49 & .16 & .28 & 0 & 0 \\
            &  & post & \textbf{.03} & \textbf{0} & \textbf{0} & \textbf{0} & \textbf{.05} & \textbf{0} & \textbf{.05} & \textbf{0} & \textbf{0} &  & 1 & 1 & 1 & \textbf{.03} & \textbf{.02} \\
            & MGD & stat & 112 & 28 & 70 & 35 & 116 & 22 & 25 & 3 & 43 &  & 157 & 183 & 197 & 219 & 112 \\
            &  & pval & .01 & 0 & 0 & 0 & .05 & .01 & .08 & 0 & 0 &  & .12 & .31 & .47 & .78 & .01 \\
            &  & post & .13 & \textbf{0} & \textbf{.01} & \textbf{0} & .47 & \textbf{.05} & .81 & \textbf{0} & \textbf{0 }&  & 1 & 1 & 1 & 1 & .12 \\
            & NAG & stat & 101 & 27 & 55 & 43 & 116 & 22 & 32 & 4 & 46 &  & 153 & 186 & 193 & 195 & 107 \\
            &  & pval & .01 & 0 & 0 & 0 & .05 & .01 & .11 & 0 & 0 &  & .10 & .34 & .42 & .44 & .01 \\
            &  & post & .07 & \textbf{0} & \textbf{0} & \textbf{0} & .47 & \textbf{.05} & 1 & \textbf{0} & \textbf{0} &  & .92 & 1 & 1 & 1 & .09 \\
            & Adagrad & stat & 83 & 22 & 0 & 0 & 0 & 0 & 0 & 0 & 0 &  & 2 & 0 & 2 & 39 & 0 \\
            &  & pval & 0 & 0 & 0 & 0 & 0 & 0 & 0 & 0 & 0 &  & 0 & 0 & 0 & 0 & 0 \\
            &  & post & \textbf{.02} & \textbf{0} & \textbf{0} & \textbf{0} & \textbf{0} & \textbf{0} & \textbf{0} & \textbf{0} & \textbf{0} &  & \textbf{0} & \textbf{0} & \textbf{0} & \textbf{0} & \textbf{0} \\
            & RMSprop & stat & 106 & 32 & 19 & 15 & 131 & 18 & 28 & 33 & 9 &  & 156 & 181 & 192 & 68 & 187 \\
            &  & pval & .01 & 0 & 0 & 0 & .10 & 0 & .21 & 0 & 0 &  & .12 & .29 & .40 & 0 & .35 \\
            &  & post & .09 & \textbf{0} & \textbf{0} & \textbf{0} & 1 & \textbf{.03} & 1 & \textbf{0} & \textbf{0} &  & 1 & 1 & 1 & .01 & 1 \\
            & Adam & stat & 123.5 & 27 & 19 & 37 & 155 & 18 & 24 & 9 & 34 &  & 145 & 198 & 203 & 0 & 199 \\
            &  & pval & .02 & 0 & 0 & 0 & .27 & 0 & .42 & 0 & 0 &  & .07 & .48 & .54 & 0 & .49 \\
            &  & post & .25 & \textbf{0} & \textbf{0} & \textbf{0} & 1 & \textbf{.03} & 1 & \textbf{0} & \textbf{0} &  & .65 & 1 & 1 & \textbf{0} & 1 \\
            %%%& $\mathcal{M}_{0.4}$NAG & stat & 106 &  60 &  90 &  21 &  108 &  9 &  24 &  18 &  14 &   & 186 &  72 &  174 &  73 &  94\\
            %%%& & pval & 0.01 & 0 & 0 & 0 & .01 & 0 & .42 & 0 & 0 &  & .34 & 0 & .23 & 0 & 0\\
            %%%& & post & 0.12 & .01 & .04 & 0 & .14 & .01 & 1 & 0 & 0 &  & 1 & .01 & 1 & .01 & .05\\
            &$\mathcal{M}_{0.4}$Adam & stat & 103 & 57 & 41 & 35 & 126 & 0 & 29 & 22 & 17 &  & 173 & 117 & 210 & 129 & 175  \\ 
            & & pval & .01 & 0 & 0 & 0 & .05 & 0 & .13 & 0 & 0 &  & .22 & .02 & .64 & .03 & .24  \\ 
            & & post & .10 & \textbf{0} & \textbf{0} & \textbf{0} & .62 & \textbf{0} & 1 & \textbf{0} & \textbf{0} &  & 1 & .21 & 1 & .40 & 1  \\ 
            \cmidrule{1-2}
            \parbox[t]{3mm}{\multirow{12}{*}{\rotatebox[origin=c]{90}{ Trees}}}
            & HFNT$^{\text{S}}$ & stat & 24 & 38 & 6 & 21 & 76 & 25 & 22 & 0 & 15 &  & 117 & 118 &  & 108 & 115 \\
            & & pval & 0 & 0 & 0 & 0 & 0 & 0 & 0 & 0 & 0 &  & .02 & .02 &  & .01 & .02 \\
            & & post & \textbf{0} & \textbf{0} & \textbf{0} & \textbf{0} & \textbf{.03} & \textbf{0} & \textbf{0} & \textbf{0} & \textbf{0} &  & \textbf{0} & \textbf{0} & \textbf{0} & \textbf{.01} & \textbf{0} \\
            & HFNT$^{\text{M}}$ & stat &  & 4 & 2 & 5 & 55 & 19 & 19 & 0 & 4 &  & 140 & 110 &  & 82 & 97 \\
            & & pval &  & 0 & 0 & 0 & 0 & 0 & 0 & 0 & 0 &  & .06 & .01 &  & 0 & .01 \\
            & & post &  & \textbf{0} & \textbf{0} & \textbf{0} & \textbf{0} & \textbf{0} & \textbf{0} & \textbf{0} & \textbf{0} &  & 1 & 1 & 1 & 1 & .12 \\
            & DT & stat & 1 & 3 & 23 & 0 & 0 & 5.5 & 0 & 69 & 130 &  & 16 & 0 & 1 & 61 & 0 \\
            &  & pval & 0 & 0 & 0 & 0 & 0 & 0 & 0 & 0 & .10 &  & 0 & 0 & 0 & 0 & 0 \\
            &  & post & \textbf{0} & \textbf{0} & \textbf{0} & \textbf{0} & \textbf{0} & \textbf{0} & \textbf{0} & \textbf{.01} & .96 &  & \textbf{0} & \textbf{0} & \textbf{0} & \textbf{0} & \textbf{0} \\
            &RF & stat & 46.50 & 7 & 92.50 & 1 & 20 & 0 & 42 & 191 & 96 &  & 225 & 201 & 144 & 30 & 226 \\
            & & pval & 0 & 0 & .01 & 0 & 0 & 0 & .02 & .57 & .03 &  & .88 & .52 & .07 & 0 & .89 \\
            & & post & .01 & 0 & .09 & 0 & 0 & 0 & .21 & 1 & .33 &  & 1 & 1 & .06 & 0 & 1 \\
            \cmidrule{1-2}
            \parbox[t]{3mm}{\multirow{9}{*}{\rotatebox[origin=c]{90}{Other}}}
            & GP & stat & 67 & 12 & 55 & 20.5 & 50 & 14 & 36.5 & 13 & 112 &  & 187 & 143 & 201 & 89 & 24 \\
            &  & pval & 0 & 0 & 0 & 0 & 0 & 0 & .09 & 0 & .04 &  & .35 & .07 & .52 & 0 & 0 \\
            &  & post & \textbf{.01} & \textbf{0} & \textbf{0} & \textbf{0} & \textbf{.01} & \textbf{0} & .93 & \textbf{0} & .38 &  & 1 & .59 & 1 & \textbf{.03} & \textbf{0} \\
            & NBC & stat & 1 & 30 & 13.5 & 10.5 & 0 & 10 & 28 & 0 & 0 &  &  &  &  &  &  \\
            &  & pval & 0 & 0 & 0 & 0 & 0 & 0 & .01 & 0 & 0 &  &  &  &  &  &  \\
            &  & post & \textbf{0} & \textbf{0} & \textbf{0} & \textbf{0} & \textbf{0} & \textbf{0} & .11 & \textbf{0} & \textbf{0} &  &  &  &  &  &  \\
            & SVM & stat & 36 & 21 & 5.5 & 4.5 & 67.5 & 4.5 & 37.5 & 145 & 6 &  & 194 & 228 & 97 & \textbf{0} & 175 \\
            &  & pval & {0} & {0} & {0} & {0} & {.01} & {0} & {.03} & .29 & {0} &  & .43 & .93 & {.01} & {0} & .24 \\
            &  & post & \textbf{0} & \textbf{0} & \textbf{0} & \textbf{0} & .06 & \textbf{0} & .30 & 1 & \textbf{0} &  & 1 & 1 & \textbf{.05} & \textbf{0} & 1 \\
            \bottomrule
        \end{tabular}
    \end{table}

    \begin{table} %[b]
        \small
        \centering
        \renewcommand{\arraystretch}{1.05}
        \setlength{\tabcolsep}{2pt}
        \caption{T-test on two independent samples: BNeuralT's RMSprop against all other algorithms for each data. The `stat,'`pval,' and `post,' respectively indicate T-test statistic, two-tailed $p$-value, and Bonferroni correction post hoc adjusted $p$-value. The values are marked in bold where null hypothesis that BNeuralT's RMSprop and other algorithms have the same expected (average) is rejected.}
        \label{tab:BNeuralT_RMSprop_VS_All_T_Test}
        \begin{tabular}[t]{lrrrrrrrrrrrrrrrrr}
            \toprule
            \multicolumn{3}{c}{BNeuralT's} & \multicolumn{9}{c}{Classification} & ~~ & \multicolumn{5}{c}{Regression} \\
            \cline{4-12}\cline{14-18}
            \multicolumn{3}{c}{RMSprop vs.}  & Aus & Hrt & Ion & Pma & Wis & Irs & Win & Vhl & Gls & ~ & Bas & Dee & Dia & Frd & Mpg \\
            \midrule
            \parbox[t]{3mm}{\multirow{18}{*}{\rotatebox[origin=c]{90}{MLP}}}
            & GD & stat & 3.21 & 6.87 & 7.51 & 6.17 & 2.52 & 3.93 & 2.55 & -5.39 & 7.02 &  & -1.22 & 1.38 & .47 & 1.42 & 2.70 \\
            &  & pval & 0 & 0 & 0 & 0 & .01 & 0 & .01 & 0 & 0 &  & .23 & .17 & .64 & .16 & .01 \\
            &  & post & \textbf{.02} & \textbf{0} & \textbf{0} & \textbf{0} & .14 & \textbf{0} & .13 & \textbf{0} & \textbf{0} &  & 1 & 1 & 1 & 1 & .08 \\
            & MGD & stat & 2.71 & 6.32 & 4.17 & 4.91 & 1.81 & 2.62 & .83 & -9.34 & 5.06 &  & -2 & .97 & .23 & -0.91 & 1.80 \\
            &  & pval & .01 & 0 & 0 & 0 & .08 & .01 & .41 & 0 & 0 &  & .05 & .33 & .82 & .37 & .08 \\
            &  & post & .09 & \textbf{0} & \textbf{0} & \textbf{0} & .75 & .11 & 1 & \textbf{0} & \textbf{0} &  & .45 & 1 & 1 & 1 & .70 \\
            & NAG & stat & 2.75 & 6.46 & 5.05 & 4.97 & 1.85 & 2.73 & .97 & -9.78 & 4.39 &  & -2.01 & .96 & .28 & -0.30 & 1.78 \\
            &  & pval & .01 & 0 & 0 & 0 & .07 & .01 & .33 & 0 & 0 &  & .05 & .34 & .78 & .76 & .08 \\
            &  & post & .08 & \textbf{0} & \textbf{0} & \textbf{0} & .70 & .08 & 1 & \textbf{0} & \textbf{0} &  & .44 & 1 & 1 & 1 & .73 \\
            & Adagrad & stat & 3.19 & 7.20 & 21.12 & 17.67 & 11.11 & 10.42 & 5.91 & 22.99 & 21.80 &  & 9.96 & 21.73 & 12.93 & 7.34 & 22.47 \\
            &  & pval & 0 & 0 & 0 & 0 & 0 & 0 & 0 & 0 & 0 &  & 0 & 0 & 0 & 0 & 0 \\
            &  & post & \textbf{.02} & \textbf{0} & \textbf{0} & \textbf{0} & \textbf{0} & \textbf{0} & \textbf{0} & \textbf{0} & \textbf{0} &  & \textbf{0} & \textbf{0} & \textbf{0} & \textbf{0} & \textbf{0} \\
            & RMSprop & stat & 2.69 & 6.07 & 8.03 & 6.40 & 2.08 & 3.22 & 0 & -5.13 & 6.90 &  & -1.98 & .95 & .22 & -2.89 & .21 \\
            &  & pval & .01 & 0 & 0 & 0 & .04 & 0 & 1 & 0 & 0 &  & .05 & .35 & .83 & .01 & .83 \\
            &  & post & .09 & \textbf{0} & \textbf{0} & \textbf{0} & .42 & \textbf{.02} & 1 & \textbf{0} & \textbf{0} &  & .47 & 1 & 1 & \textbf{.05} & 1 \\
            & Adam & stat & 2.44 & 6.39 & 6.73 & 4.81 & .78 & 3.06 & 0 & -7.10 & 5.15 &  & -2.13 & .66 & .05 & -6.04 & -0.84 \\
            &  & pval & .02 & 0 & 0 & 0 & .44 & 0 & 1 & 0 & 0 &  & .04 & .51 & .96 & 0 & .40 \\
            &  & post & .18 & \textbf{0} & \textbf{0} & \textbf{0} & 1 & \textbf{.03} & 1 & \textbf{0} & \textbf{0} &  & .34 & 1 & 1 & \textbf{0} & 1 \\
            %%%& $\mathcal{M}_{0.4}$NAG & stat & 2.86 & 4.77 & 3.46 & 5.81 & 2.64 & 3.26 & 0 & -6.96 & 6.96 &  & -1.23 & 4.02 & 1.14 & 2.09 & 2.36\\
            %%%& & pval& 0.01 & 0 & 0 & 0 & .01 & 0 & 1 & 0 & 0 &  & .22 & 0 & .26 & .04 & .02\\
            %%%& & post & 0.08 & 0 & .01 & 0 & .14 & .02 & 1 & 0 & 0 &  & 1 & 0 & 1 & .49 & .26\\
            %
            &$\mathcal{M}_{0.4}$Adam & stat & 2.88 & 4.65 & 5.20 & 5.23 & 2.12 & 4.75 & .30 & -6.24 & 6.90 &  & -1.69 & 2.30 & .31 & .56 & .55  \\ 
            & & pval & .01 & 0 & 0 & 0 & .04 & 0 & .76 & 0 & 0 &  & .10 & .03 & .76 & .58 & .59  \\ 
            & & post & .07 & \textbf{0} & \textbf{0} & \textbf{0} & .50 & \textbf{0} & 1 & \textbf{0} & \textbf{0} &  & 1 & .30 & 1 & 1 & 1  \\ 
            \cmidrule{1-2}
            \parbox[t]{3mm}{\multirow{12}{*}{\rotatebox[origin=c]{90}{Trees}}}
            &HFNT$^{\text{S}}$ & stat & 5.31 & 5.26 & 9.07 & 6.75 & 3.72 & 3.84 & 5.84 & 11.80 & 8.14 &  & 2.51 & 2.67 &  & 1.41 & 2.79 \\
            & & pval & 0 & 0 & 0 & 0 & 0 & 0 & 0 & 0 & 0 &  & .01 & .01 &  & .16 & .01 \\
            & & post & \textbf{0} & \textbf{0} & \textbf{0} & \textbf{0} & \textbf{.01} & \textbf{0} & \textbf{0} & \textbf{0} & \textbf{0} &  & \textbf{0} & \textbf{0} & \textbf{0} & \textbf{.04} & \textbf{0} \\            
            &HFNT$^{\text{M}}$ & stat & 4.69 & 8.97 & 7.44 & 7.97 & 5.04 & 4.39 & 5.46 & 15.45 & 9.42 &  & 2.09 & 2.59 &  & 2.03 & 2.70 \\
            & & pval & 0 & 0 & 0 & 0 & 0 & 0 & 0 & 0 & 0 &  & .04 & .01 &  & .05 & .01 \\
            & & post & \textbf{0} & \textbf{0} & \textbf{0} & \textbf{0} & \textbf{0} & \textbf{0} & \textbf{0} & \textbf{0} & \textbf{0} &  & .59 & 1 & 1 & 1 & .97 \\
            & DT & stat & 10.88 & 10.20 & 7.25 & 12.41 & 10.59 & 6.40 & 9.79 & 3.87 & 2.04 &  & 6.36 & 11.20 & 14.34 & 3.10 & 7.24 \\
            &  & pval & 0 & 0 & 0 & 0 & 0 & 0 & 0 & 0 & .05 &  & 0 & 0 & 0 & 0 & 0 \\
            &  & post & \textbf{0} & \textbf{0} & \textbf{0} & \textbf{0} & \textbf{0} & \textbf{0} & 0 & \textbf{0} & .46 &  & \textbf{0} & \textbf{0} & \textbf{0} & \textbf{.03} & \textbf{0} \\
            & RF & stat & 3.43 & 7.32 & 2.69 & 9.75 & 6.73 & 5.50 & 1.55 & .44 & -2.21 &  & .07 & .65 & 1.92 & -3.43 & -0.40 \\
            & & pval & 0 & 0 & .01 & 0 & 0 & 0 & .13 & .66 & .03 &  & .94 & .52 & .06 & 0 & .69 \\
            & & post & \textbf{.01} & \textbf{0} & .12 & \textbf{0} & \textbf{0} & \textbf{0} & 1 & 1 & .40 &  & 1 & 1 & \textbf{.02} & \textbf{0} & 1 \\
            \cmidrule{1-2}
            \parbox[t]{3mm}{\multirow{9}{*}{\rotatebox[origin=c]{90}{Others}}}
            & GP & stat & 4.12 & 7.03 & 4.52 & 7.35 & 4.19 & 5.03 & 1.14 & -8.37 & 2.17 &  & .65 & 1.61 & .32 & 1.86 & 7.32 \\
            &  & pval & 0 & 0 & 0 & 0 & 0 & 0 & .26 & 0 & .03 &  & .52 & .11 & .75 & .07 & 0 \\
            &  & post & \textbf{0} & \textbf{0} & \textbf{0} & \textbf{0} & \textbf{0} & \textbf{0} & 1 & \textbf{0} & .34 &  & 1 & 1 & 1 & .62 & \textbf{0} \\
            & NBC & stat & 12.03 & 5.53 & 8.72 & 8.75 & 11.12 & 6.22 & 2.60 & 28.51 & 14.55 &  &  &  &  &  &  \\
            &  & pval & 0 & 0 & 0 & 0 & 0 & 0 & .01 & 0 & 0 &  &  &  &  &  &  \\
            &  & post & \textbf{0} & \textbf{0} & \textbf{0} & \textbf{0} & \textbf{0} & \textbf{0} & .12 & \textbf{0} & \textbf{0} &  &  &  &  &  &  \\
            & SVM & stat & 4.43 & 5.50 & 7.99 & 7.72 & 2.83 & 7.57 & 1.88 & -1.15 & 8.85 &  & .66 & -0.08 & 3.27 & -4.91 & .63 \\
            &  & pval & 0 & 0 & 0 & 0 & .01 & 0 & .06 & .26 & 0 &  & .51 & .93 & 0 & 0 & .53 \\
            &  & post & \textbf{0} & \textbf{0} & \textbf{0} & \textbf{0} & .06 & \textbf{0} & .65 & 1 & \textbf{0} &  & 1 & 1 & \textbf{.02} & \textbf{0} & 1 \\
            \bottomrule
        \end{tabular}
    \end{table}
    
\end{document}